%% file: main.tex
\documentclass[11pt]{article}

\usepackage[margin=1in]{geometry}

\usepackage[utf8]{inputenc}
\usepackage[T1]{fontenc}
\usepackage{lmodern}
\usepackage{graphicx}
\graphicspath{{figures/}}
\makeatletter
\@ifundefined{Gread@transgroupfalse}{\let\includegraphicsnogroup\includegraphics}{\newcommand\includegraphicsnogroup[2][]{{\def\Gread@pdf##1{\let\Gin@decode\@empty
  \Gin@interpolatefalse\Gread@transgroupfalse\Gread@pdftex{##1}}\includegraphics[#1]{#2}}}}
\makeatother
\usepackage{longtable}
\usepackage{hyperref}
\usepackage{url}
\usepackage{booktabs}
\usepackage{amsfonts}
\usepackage{amsmath}
\usepackage{amssymb}
\usepackage{mathtools}
\usepackage{microtype}
\usepackage[capitalize,noabbrev]{cleveref}
\usepackage{enumitem}
\usepackage{authblk}
\usepackage{natbib}
\usepackage[only,llbracket,rrbracket]{stmaryrd}
\newcommand{\R}{\mathbb{R}}
\newcommand{\E}{\mathbb{E}}
\newcommand{\diag}{\operatorname{diag}}
\newcommand{\sgn}{\operatorname{sign}}
\newcommand{\wbar}{\bar{w}}
\newcommand{\mbar}{\bar{m}}
\newcommand{\nubar}{\bar{\nu}}
\newcommand{\zbar}{\bar{z}}
\newcommand{\gbar}{\bar{g}}
\newcommand{\thetabar}{\bar{\theta}}
\newcommand{\dhat}{\hat{\delta}}
\newcommand{\jump}[1]{\llbracket #1 \rrbracket}
\newcommand{\gjump}{\jump{g}}
\newcommand{\Ac}{\mathcal{A}}
\newcommand{\Psic}{\Psi}
\newcommand{\Psit}{\widetilde{\Psi}}
\newcommand{\Kmat}{K}
\newcommand{\Tt}{\mathcal{T}}
\newcommand{\feas}{\Omega}
\newcommand{\Fval}{\mathcal{F}}
\newcommand{\thr}{S^{\star}}
\newcommand{\ind}{\iota}
\newcommand{\inner}[2]{\langle #1, #2\rangle}
\newcommand{\norm}[1]{\lVert #1\rVert}
\newcommand{\ot}{\otimes}
\newcommand{\dd}{\mathrm{d}}
\newcommand{\ddt}[1]{\frac{\mathrm{d}#1}{\mathrm{d}t}}
\newcommand{\lamone}{\lambda_{1}}
\newcommand{\lammax}{\lambda_{\max}}

\title{Unifying Central Flow, Rod Flow, and the Edge Coupling}

\author[1]{Eric Regis}
\author[1]{Sinho Chewi}
\affil[1]{Yale University}
\affil[ ]{\texttt{\{eric.regis, sinho.chewi\}@yale.edu}}

\date{}

\begin{document}

\maketitle

\begin{abstract}
At the edge of stability, the iterates of gradient descent oscillate along the sharp directions of the Hessian. This poses a significant challenge for continuous-time models as the usual continuous-time approximation---the gradient flow---admits no oscillations. Several continuous-time models have been proposed to capture the behavior at the edge of stability, among them central flow and rod flow. Although both models can be interpreted as performing a coarse-graining in time, the connection between them remains poorly understood. Here we show that both flows can be derived from a third object, the edge coupling: a functional on consecutive iterates whose critical points are the fixed points and period-two orbits of gradient descent. It is revealed that the two flows correspond to coarse-grainings \emph{at different time scales}. Rod flow coarse-grains over only the fastest time scale, while central flow coarse-grains over both the fast and intermediate time scales. Furthermore, using the edge coupling, we derive rod flows and central flows for momentum methods such as heavy ball and Adam. We verify the derived flows experimentally on representative neural network architectures, where they track the discrete iterates throughout all phases of training---including the transition from the stable regime to the edge of stability.
\end{abstract}

\tableofcontents

\input{sections/01-intro}
\input{sections/02-related}
\input{sections/03-background}
\input{sections/04-rodflow-gda}
\input{sections/05-rod-flows-family}
\input{sections/06-central-from-coupling}
\input{sections/07-central-flows-momentum}
\input{sections/08-timescales}
\input{sections/09-experiments}
\input{sections/10-conclusion}

\clearpage
\bibliography{references}

\clearpage
\appendix
\crefalias{section}{appendix}
\crefalias{subsection}{subappendix}
\crefalias{subsubsection}{subsubappendix}
\input{appendix/A-notation}
\input{appendix/B-structured}
\input{appendix/C-nesterov}
\input{appendix/D-adamw}
\input{appendix/E-rodflow}
\input{appendix/F-centralflow}
\input{appendix/G-experimental-details}
\input{appendix/H-more-experiments}

\end{document}

%% file: sections/01-intro.tex
\section{Introduction}
\label{sec:introduction}

\citet{cohen2021gradient} observed that gradient descent operates at the edge of stability. During training, the sharpness (the largest eigenvalue of the Hessian) rises in a phase called \textit{progressive sharpening}. The sharpness then hovers at $2/\eta$, where $\eta$ is the learning rate. Once there, the iterates bounce across the valley of the loss landscape rather than rolling downhill smoothly. It's been found that momentum methods such as Adam \textit{also} operate at the edge of stability \citep{cohen2022adaptive}.

Continuous-time models are a standard tool for analyzing discrete-time dynamics. However, the edge of stability poses a significant challenge for such models. At the edge of stability, the iterates oscillate along the sharp directions of the Hessian. The usual continuous-time limit of gradient descent---the gradient flow---cannot capture these oscillations.

There are several continuous-time models of gradient descent that aim to capture the dynamics at the edge of stability \citep{Cohen+25CentralFlow, regis2026rod, chodron2026free, marion2026edge}. \citet{Cohen+25CentralFlow} introduced central flow: a continuous-time model that aims to capture not the individual iterates, but their time-averaged position. It acts as a projected gradient flow---gradient flow that is constrained to the feasible region where the sharpness is at most $2/\eta$. The oscillations are represented by a covariance matrix that is found by solving a semidefinite complementarity problem (SDCP). In the same paper, they also developed central flows for scalar RMSProp and RMSProp.

Rod flow~\citep{regis2026rod} is a continuous-time model that represents consecutive iterates as an extended one-dimensional object---a ``rod''---whose midpoint and extent evolve smoothly. In contrast with the covariance matrix of central flow, the extent matrix of rod flow evolves according to an ODE. \citet{regis2026adamrod} then lifted the rod from weight space to phase space, yielding rod flows for momentum methods such as heavy ball and Adam.

Central flow and rod flow both aim to model the dynamics of gradient descent when coarse-grained in time. As a result, one would expect that the two flows must capture the same underlying structure. Yet the two flows were derived under very different modeling assumptions. Central flow was derived by time-averaging the iterates and imposing the empirically-observed constraint that the sharpness remains at $2/\eta$. Meanwhile, rod flow was derived by assuming that the midpoint of consecutive iterates and the outer product of their half-displacement vary smoothly. Whether a deeper link exists between them has remained unclear.

Recently, \citet{litman2026origin} introduced the \textit{edge coupling}: a functional on consecutive iterates whose critical points encode the fixed points and period-two orbits of gradient descent. With it, he sheds light on why the dynamics drive the sharpness to $2/\eta$. \citet{regis2026variational} then extended the edge coupling to heavy ball and Nesterov momentum.

\begin{figure}[t]
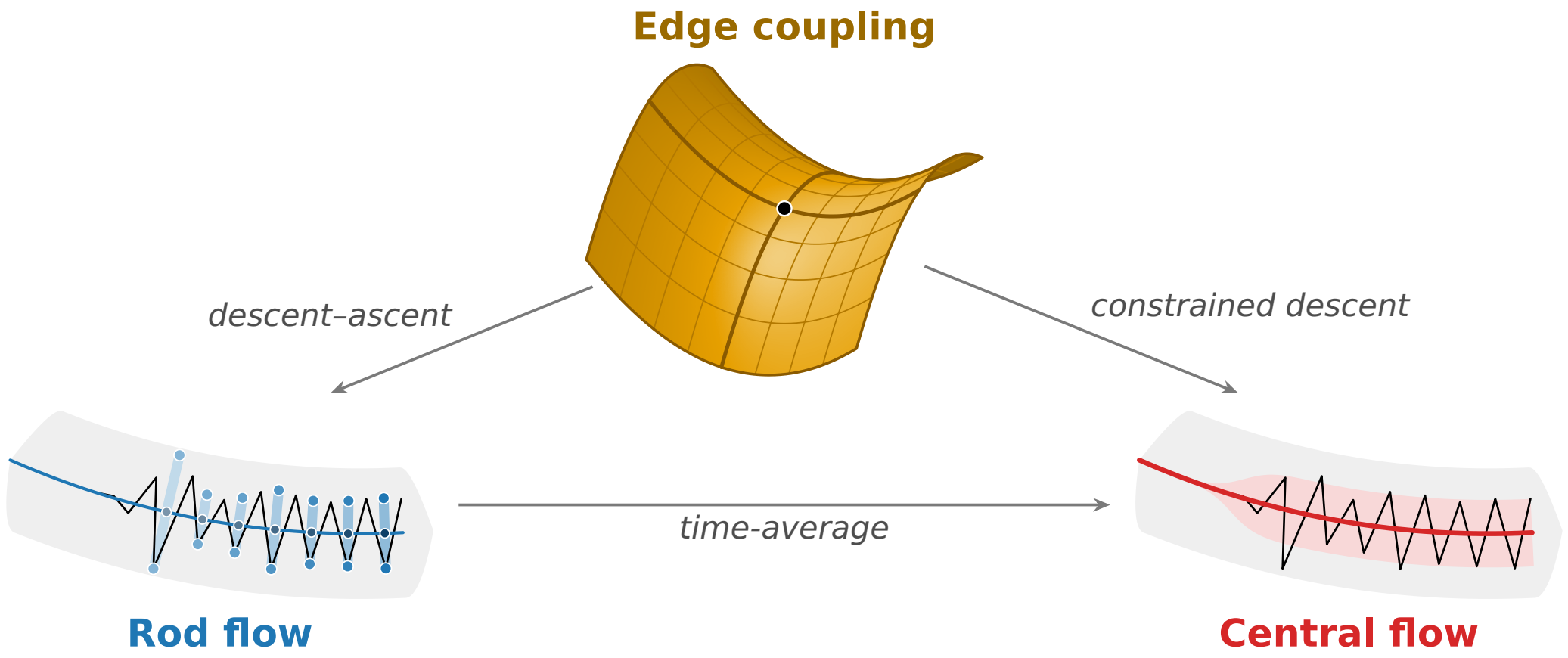

	\centering
	\includegraphicsnogroup{schematic/figure_1_triangle_wide.pdf}
	\caption{\textbf{The Edge Coupling Generates Both Rod Flow and Central Flow.} Rod flow is gradient descent--ascent on the centered edge coupling, with the oscillations tracked by an explicit state variable. Central flow is constrained descent on the same objective, with the covariance matrix acting as a Lagrange multiplier. Central flow is rod flow averaged over the intermediate time scale of self-stabilization.}
	\label{fig:overview}
\end{figure}

Here we establish connections between all three objects: central flow, rod flow, and the edge coupling (\cref{fig:overview}).

\paragraph{Rod flow and edge coupling.} Analogous to the relationship between gradient flow and the loss function, one can derive a rod flow by taking the centered edge coupling as the generating function. However, the resulting dynamics are not gradient descent, but \textit{gradient descent--ascent}: descent on the midpoint $\bar{w}$ and ascent on the extent $\delta$. This alternative formulation of rod flow fixes a subtle error in the original, which allows it to track the discrete iterates through the transition to the edge of stability. We then generalize the construction to the other optimizers of the Adam family---deriving rod flows for RMSProp, heavy ball momentum, and Adam.

\begin{figure}[t]
	\centering
	\includegraphics[width=0.92\textwidth]{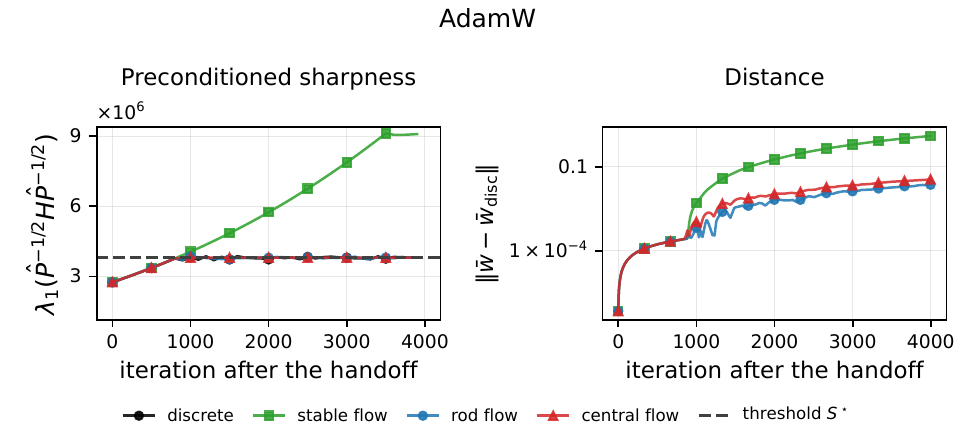}
	\caption{\textbf{Rod Flow and Central Flow Track AdamW.} AdamW on an MLP ($\eta = 10^{-5}$, $\beta_1 = 0.9$, $\beta_2 = 0.999$, $\zeta = 10$). \textbf{Left:} With both rod flow and central flow, the preconditioned sharpness self-stabilizes at the correct value. \textbf{Right:} Both flows track the discrete iterates more accurately than the stable flow does.}
	\label{fig:overview-adamw}
\end{figure}

\paragraph{Central flow and edge coupling.} One can also re-derive the central flow for gradient descent from the centered edge coupling. To do so, we replace the loss in the edge coupling with its local quadratic approximation, which makes the edge coupling linear in the covariance matrix $\Sigma$. Then, instead of being governed by an ODE, $\Sigma$ is determined as the solution of a maximization problem parametrized by the midpoint $\bar{w}$. By imposing a persistence condition on $\Sigma$, we recover precisely the SDCP of central flow. Using a similar approach, we derive novel central flows for momentum methods such as heavy ball and Adam (\cref{fig:overview-adamw}).

Because rod flow is also generated by the edge coupling, the two flows now share a common derivation. Mathematically, they differ only in whether the oscillation is treated as a state variable or as a Lagrange multiplier.

\paragraph{Rod flow and central flow.} \citet{berthier2026multiple} derived the central flow for gradient descent using the method of multiple scales, identifying three time scales. There's the fast time scale on which the components of the iterates along sharp directions alternate in sign. There's the intermediate time scale on which the oscillation amplitude self-stabilizes. And there's the slow time scale that governs the drift of the center. Building off of Berthier's observation, we show that rod flow and central flow correspond to coarse-grainings \textit{at different time scales}. Rod flow coarse-grains over the fast time scale only. It models the intermediate time scale dynamics of the oscillation envelope, exhibiting fluctuations at the onset of self-stabilization. In contrast, central flow coarse-grains over both the fast \textit{and} intermediate time scales.

We thus unify the three objects.

\subsection{Contributions}
\begin{itemize}[leftmargin=*,itemsep=0pt,topsep=2pt]
    \item A rod flow for gradient descent, from the centered edge coupling (\cref{sec:rodflow-gda}).
    \item An extension of rod flow to heavy ball and Adam, from the phase-space edge coupling (\cref{sec:rod-flows-family}).
    \item A derivation of the central flow for gradient descent from the centered edge coupling (\cref{sec:central-from-coupling}).
    \item A construction of novel central flows for heavy ball and Adam (\cref{sec:central-flows-momentum}).
    \item A demonstration that rod flow and central flow coarse-grain at different time scales (\cref{sec:timescales}).
    \item An experimental verification of rod flow and central flow (\cref{sec:experiments}).
\end{itemize}

%% file: sections/02-related.tex
\section{Related Work}
\label{sec:related}

\textbf{Edge of stability.} \citet{jastrzebski2020breakeven} documented a break-even point on the optimization trajectories of SGD. \citet{cohen2021gradient} observed that full-batch gradient descent operates at the edge of stability: the sharpness rises to $2/\eta$ and then hovers there. \citet{cohen2022adaptive} extended the picture to momentum and adaptive methods, and \citet{islamov2026noneuclidean} to non-Euclidean gradient descent. Theoretical accounts include self-stabilization \citep{DamNicLee23SelfStab}, the two-step analysis of period-two orbits \citep{ahn2023learning, chen2023beyond}, and convergence in the unstable regime \citep{ahn2022understanding, arora2022understanding, zhu2023understanding}. Others study progressive sharpening in quadratic regression \citep{AgaPedPen23SecondOrder}, the catapult mechanism \citep{lewkowycz2020large}, and its amplification by momentum \citep{Phu+24MomentumCatapults}. Further lines include loss-landscape and bifurcation analyses \citep{MaKunYin22BeyondQuadratic, song2023trajectory}, the edge of stochastic stability of minibatch SGD \citep{andreyev2024edge} and of its momentum variants \citep{andreyev2026momentum}, chaotic convergence to a statistical distribution \citep{KonTao20StochasticityGD}, and a fixed-point analysis with a period-doubling route to chaos \citep{kalra2025universal}. \citet{litman2026origin} introduced the edge coupling, a functional on consecutive iterates whose critical points are the fixed points and period-two orbits of gradient descent. \citet{regis2026variational} extended the edge coupling to heavy ball and Nesterov momentum.

\textbf{Continuous-time models.} Gradient flow is the classical continuous limit of gradient descent but cannot capture the oscillations of the edge of stability. \citet{rosca2023continuous} study its instabilities. Modified equations from backward error analysis \citep{barrett2021implicit, smith2021origin, hairer2006geometric} do not reach the oscillatory regime either. Central flow \citep{Cohen+25CentralFlow} was the first continuous-time model of gradient descent, scalar RMSProp and RMSProp at the edge of stability. \citet{berthier2026multiple} derived central flow for gradient descent by the method of multiple scales, identifying three time scales. Rod flow \citep{regis2026rod} models consecutive iterates as an extended one-dimensional object. \citet{regis2026adamrod} lifted the rod to phase space, deriving rod flows for momentum and preconditioned methods such as Adam. Other continuous-time models at the edge of stability include edge flow \citep{marion2026edge} and the mean--fluctuation model \citep{chodron2026free}.

\textbf{Discrete mechanics.} The edge coupling can be identified with the two-period Störmer--Verlet action \citep{verlet1967computer, hairer2003acta}, connecting the edge of stability to discrete mechanics \citep{marsden2001discrete}. The linear stability of periodic orbits through the Hessian of a discrete action goes back to \citet{mackay1983linear}. Symplectic integrators and their shadow Hamiltonians are treated by \citet{hairer2006geometric} and \citet{benettin1994shadow}. A separate line connects momentum methods to variational and conformally symplectic dynamics \citep{wibisono2016variational, franca2020conformal, muehlebach2021optimization}.

%% file: sections/03-background.tex
\section{Background}
\label{sec:background}

We briefly recap central flow, rod flow, and the edge coupling.

\subsection{Setup}

Consider gradient descent with step size $\eta$ on a loss function $L:\R^d\to\R$:
\begin{equation}
	w_{t+1} = w_t - \eta\,\nabla L(w_t).
	\label{eq:gd}
\end{equation}
Let $H = \nabla^2 L$ denote the Hessian of the loss, and let $\lamone(H)$ denote the sharpness---the largest eigenvalue of the Hessian. For quadratic objectives, the GD iterates diverge when the sharpness exceeds $2/\eta$.

Gradient descent operates at the \emph{edge of stability} (EoS) when the sharpness hovers at the threshold $2/\eta$. At EoS, the iterates oscillate along the top eigenvectors of the Hessian, with the sharp components of the gradient alternating in sign from one iteration to the next. These oscillatory dynamics pose a significant challenge for continuous-time models that aim to describe gradient descent at the edge of stability.

\subsection{Central Flow}
\label{sec:background-central}

Central flow~\citep{Cohen+25CentralFlow} is a continuous-time model of gradient descent at the edge of stability. Instead of tracking the individual oscillating iterates, it models their time-averaged position $\bar{w}$ and a covariance matrix $\Sigma$ which represents the oscillations. The central flow is given by the ODE:
\begin{equation}
	\ddt{\bar{w}} = -\eta \Bigl[ \nabla L(\bar{w}) + \tfrac{1}{2} \nabla_{\bar{w}} \inner{H(\bar{w})}{\Sigma} \Bigr].
	\label{eq:central-flow}
\end{equation}
The first term is the familiar gradient flow. The second term acts as a \emph{sharpness-reduction force}, pushing the flow toward flatter regions of the loss landscape. Indeed, when $H$ has only one sharp eigenvalue and the oscillations are confined to the sharp direction, we have:
\begin{equation}
	\tfrac{1}{2} \nabla_{\bar{w}} \inner{H(\bar{w})}{\Sigma} = \tfrac{\sigma^2}{2} \nabla_{\bar{w}} \lamone(H(\bar{w})),
\end{equation}
where $\sigma^2$ is the squared magnitude of the oscillations.

Central flow requires that the following three static conditions must hold along the trajectory:
\begin{enumerate}
\item The covariance matrix must be positive semidefinite ($\Sigma \succeq 0$).
\item The slack $\Lambda = \tfrac{2}{\eta} I_d - H$ must also be positive semidefinite ($\Lambda \succeq 0$).
\item The oscillations must be confined to the sharp directions ($\langle \Lambda, \Sigma \rangle = 0$).
\end{enumerate}
However, these static conditions alone are not sufficient to determine the covariance matrix. To do so, one must invoke persistence: we require that the dynamics also preserve the constraints.

Let $U$ be a matrix whose columns form an orthonormal basis for the critical directions---directions whose sharpness equals $2/\eta$. Let $k$ denote the number of critical directions. We can then express the covariance matrix as $\Sigma = U X U^\top$ where $X$ is a $k \times k$ positive semidefinite matrix. Differentiating in time yields the following dynamic constraints:
\begin{enumerate}
	\item The slack cannot decrease along critical directions ($U^\top \dot{\Lambda} U \succeq 0$).
	\item There can be no oscillation in directions whose slack is increasing ($\langle U^\top \dot{\Lambda} U, X \rangle = 0$).
\end{enumerate}
Together with the condition that $X \succeq 0$, these dynamic constraints define a \textit{semidefinite complementarity problem}---which determines $\Sigma$.

Central flow can also be interpreted as a \textit{projected gradient flow}. Let $\Omega$ denote the feasible region, the set of points whose sharpness is at most $2/\eta$:
\begin{equation}
	\Omega = \left\{ w \in \mathbb{R}^d : \lamone(H) \le 2/\eta \right\}.
\end{equation}
Central flow is gradient flow constrained to $\Omega$, with $\Sigma$ acting as the Lagrange multiplier that enforces the constraint.

\subsection{Rod Flow}
\label{sec:background-rod}

Rod flow~\citep{regis2026rod} is a continuous-time model of gradient descent where consecutive iterates are modeled as an extended one-dimensional object---a ``rod''. Define the midpoint $\bar{w}$ and half-displacement $\hat{\delta}$ of consecutive iterates:
\begin{align}
\bar{w}_{t} &= \tfrac{1}{2}(w_t + w_{t+1}) \\
\hat{\delta}_t &= \tfrac{1}{2}(w_t - w_{t+1})
\end{align}
The core modeling assumption of rod flow is that, although the individual iterates oscillate at the edge of stability, the midpoint $\wbar_t$ and outer product $\dhat_t \ot \dhat_t$ vary smoothly---and therefore can be tracked by ODEs.

Let $\Sigma$ denote the extent matrix, the continuous-time analog of $\hat{\delta} \otimes \hat{\delta}$. And let $\delta$ denote the principal eigenvector of $\Sigma$ scaled by the square root of its principal eigenvalue. Although the extent matrix is not constrained to have rank 1, it has been found empirically that it can safely be approximated as rank 1.

Let $\bar{g}$ denote the average of the gradients evaluated at the rod endpoints, and let $\overline{g^{\otimes 2}}$ denote the average of their outer products:
\begin{equation}
	\bar{g} = \tfrac{1}{2} \bigl[ \nabla L(\bar{w} + \delta) + \nabla L(\bar{w} - \delta) \bigr],
	\qquad
	\overline{g^{\otimes 2}} = \tfrac{1}{2} \bigl[ \nabla L(\bar{w} + \delta)^{\otimes 2} + \nabla L(\bar{w} - \delta)^{\otimes 2} \bigr].
\end{equation}
The rod flow ODEs for $\bar{w}$ and $\Sigma$ are given as:
\begin{align}
  \ddt{\wbar} &= -\eta\,\gbar, \label{eq:rod-center}\\
  \ddt{\Sigma} &= \frac{\eta^2}{2} \, \overline{g^{\otimes 2}} - 2\Sigma. \label{eq:rod-sigma}
\end{align}
The midpoint $\bar{w}$ moves with the mean gradient, while the extent $\Sigma$ is driven by the second moment of the endpoint gradients and damped by the $-2\Sigma$ term. Along directions flatter than the $2/\eta$ threshold, damping dominates and $\Sigma$ decays quickly. As a result, the extent matrix remains confined to the sharp directions. However, along critical directions, the source and damping terms nearly cancel. Along these directions, $\Sigma$ grows when the sharpness exceeds the threshold and shrinks when the sharpness falls below it. Thus, unlike with central flow, $\Sigma$ is not fixed by $\wbar$---instead, it is set dynamically by self-stabilization.

\subsection{Edge Coupling}
\label{sec:coupling}

The edge coupling \citep{litman2026origin} is a functional that takes two points in weight space as arguments:
\begin{equation}
	\Ac_\eta(w, w') = L(w) + L(w') - \tfrac{1}{2\eta}\,\norm{w - w'}^2 .
	\label{eq:A-def}
\end{equation}
The critical points of the edge coupling encode the fixed points and period-two orbits of gradient descent.

The edge coupling can be thought of as a generalization of the loss function. The loss has the property that its critical points are precisely the fixed points of gradient descent:
\begin{equation}
\nabla L(w_t) = 0 \iff w_{t+1} = w_t.
\end{equation}
The edge coupling characterizes not only the fixed points but \textit{also} the period-two orbits. Taking the partial derivative of $\mathcal{A}_\eta$ with respect to $w$ and setting it equal to zero recovers the gradient descent update equation from $w$ to $w'$:
\begin{equation}
	\nabla_w \Ac_\eta = 0
	\iff
	w' = w - \eta\nabla L(w).
\end{equation}
And because $\Ac_\eta$ is symmetric under swapping $w\leftrightarrow w'$, we also have that:
\begin{equation}
	\nabla_{w'} \Ac_\eta = 0
	\iff
	w = w' - \eta\nabla L(w').
\end{equation}
There are only two circumstances in which both criticality conditions hold: $w = w'$ is a fixed point of gradient descent, or $w\neq w'$ and $(w, w')$ form a period-two orbit.
\begin{equation}
	\nabla\Ac_\eta(w_t, w_{t+1}) = 0 \iff w_{t+2} = w_t .
	\label{eq:litman-property}
\end{equation}

For any smooth, invertible change of coordinates, critical points correspond bijectively between the old and new coordinates. We may therefore express the edge coupling in whatever coordinates are most convenient. Consider the coordinate system:
\begin{equation}
	\wbar = \frac{w + w'}{2}, \qquad \delta = \frac{w - w'}{2}.
	\label{eq:centered-coords}
\end{equation}
We can then define the \emph{centered edge coupling}:
\begin{align}
	\Psic(\wbar, \delta) &\equiv \tfrac{1}{2}\Ac_\eta(\wbar + \delta, \wbar - \delta) \\
	&= \tfrac12\bigl[L(\wbar + \delta) + L(\wbar - \delta)\bigr] - \tfrac{1}{\eta}\,\norm{\delta}^2 .
\end{align}
The factor of $\tfrac{1}{2}$, while not strictly necessary, will prove convenient later.

The edge coupling encodes not only the fixed points and period-two orbits of gradient descent, but also their linear stability---through its Hessian. In the special case where the Hessians at the two arguments agree--- $H(\wbar+\delta) = H(\wbar-\delta)$---the cross block $\nabla^2_{\wbar\delta}\Psic = \tfrac12[H(\wbar+\delta) - H(\wbar-\delta)]$ vanishes, so the Hessian of the centered edge coupling is block-diagonal. We then have that a period-two orbit is linearly stable if and only if:
\begin{equation}
	\nabla^2_{\wbar\wbar}\Psic \succ 0 \qquad\text{and}\qquad \nabla^2_{\delta\delta}\Psic \prec 0 .
	\label{eq:minmax}
\end{equation}

In this case, linearly-stable period-two orbits correspond to min--max critical points of the centered edge coupling.

%% file: sections/04-rodflow-gda.tex
\section{Rod Flow and Edge Coupling}
\label{sec:rodflow-gda}

We show that one can derive a rod flow for gradient descent as continuous-time gradient descent--ascent on the centered edge coupling. We also derive a rod flow for RMSProp.

\subsection{Gradient Descent--Ascent}
\label{sec:criticality}

Let $\gjump$ denote the half-difference of the gradients of the endpoints of the rod.
\begin{equation}
	\gjump = \tfrac{1}{2}[\nabla L(\bar{w} + \delta) - \nabla L(\bar{w} - \delta)]
\end{equation}
Taking derivatives of the centered edge coupling with respect to $\bar{w}$ and $\delta$, we obtain:
\begin{align}
	\nabla_{\wbar}\Psic &= \gbar \\
	\nabla_{\delta}\Psic &= \gjump - \frac{2}{\eta}\,\delta .
	\label{eq:Psi-grad}
\end{align}
Examining the right-hand side of our equation for $\nabla_{\bar{w}} \Psic$, we find a familiar quantity: the average gradient $\bar{g}$. Recall that the average gradient is precisely what drives the midpoint $\bar{w}$ in rod flow. The expression for $\nabla_{\delta} \Psic$ is also reminiscent of the right-hand side of the $\Sigma$ ODE in rod flow:
\begin{equation*}
	\ddt{\Sigma} = \frac{\eta^2}{2}\,\overline{g^{\otimes 2}} - 2\Sigma.
\end{equation*}
Even though $\delta$ is a vector while $\Sigma$ is a matrix, they nonetheless share a similar form. In rod flow, $\Sigma$ relaxes toward a function of the endpoint gradients, whereas our criticality condition for $\delta$ is a self-consistency condition: $\delta$ must equal the half-difference of the endpoint gradients multiplied by $\eta/2$.

Can one derive a rod flow for gradient descent by letting $\bar{w}$ and $\delta$ evolve under gradient flow on the centered edge coupling?

Yes---though there is one subtlety. For gradient flow on the loss, the stable fixed points correspond to critical points of the loss that are local minima. However, recall from~\eqref{eq:minmax} that for the centered edge coupling---when the endpoint Hessians agree---stable period-two orbits correspond to critical points that are minima in $\bar{w}$ and maxima in $\delta$.

So instead of gradient descent, we must perform \textit{gradient descent--ascent}: descending in $\bar{w}$ and ascending in $\delta$. This yields the following rod flow:
\begin{equation}
	\boxed{
		\ddt{}\begin{pmatrix} \bar{w} \\ \delta \end{pmatrix}
		= \eta \begin{pmatrix} -I_d & 0 \\ 0 & I_d \end{pmatrix} \nabla_{(\bar{w}, \delta)} \Psic
	}
\end{equation}
Expressing the ODEs for $\bar{w}$ and $\delta$ separately, we have:
\begin{align}
\ddt{\bar{w}} &= - \eta \bar{g} \\
\ddt{\delta} &= \eta \gjump - 2 \delta
\label{eq:gda}
\end{align}

\subsection{The Zero Fixed Point}
\label{sec:seed}

Before the edge of stability, $\delta = 0$ is a strong attractor of rod flow. This can be seen by Taylor expanding the gradient half-difference to first order in $\delta$:
\begin{equation}
\ddt{\delta} = \eta \gjump - 2 \delta \approx [\eta H - 2I_d]\delta
\label{eq:gda-linear}
\end{equation}
Thus, in the stable regime---when all eigenvalues of $H$ lie below $2/\eta$---each component of $\delta$ decays exponentially, at a rate proportional to the gap between the corresponding Hessian eigenvalue and $2/\eta$.

This attraction to the zero fixed point has a practical consequence: even after the sharpness crosses the $2/\eta$ threshold, it takes time for $\delta$ to recover from the small value it decayed to during the stable phase in order to grow large enough to trigger self-stabilization.

This phenomenon is not specific to rod flow. Gradient descent exhibits the same phenomenon. Let $\dhat_t = \tfrac{1}{2}(w_t - w_{t+1}) = \tfrac{\eta}{2}\nabla L(w_t)$ be the half-displacement between consecutive iterates, and let $\widetilde{H}_t$ be the Hessian averaged along the gradient step:
\begin{equation}
\widetilde{H}_t = \int_{0}^{1} \nabla^2 L\bigl(w_t - s\,\eta\nabla L(w_t)\bigr)\,\dd s .
\end{equation}
By the fundamental theorem of calculus:
\begin{equation}
\nabla L(w_{t+1}) - \nabla L(w_t) = -\eta\,\widetilde{H}_t\nabla L(w_t).
\end{equation}
We can then deduce:
\begin{equation}
\dhat_{t+1} = \bigl(I - \eta\widetilde{H}_t\bigr)\,\dhat_t .
\label{eq:gd-halfstep}
\end{equation}
Let $u$ denote an eigenvector of $\widetilde{H}_t$ with eigenvalue $\lambda$. The $u$-component of the half-displacement is multiplied by $1 - \eta\lambda$ at every step, so it contracts exactly when $\lambda \in (0, 2/\eta)$.

We have that both gradient descent and rod flow are predicted to exhibit exponential decay in the stable regime for $\hat{\delta}_t$ and $\delta$, respectively. However, it's been empirically observed that the sharp component of gradient descent hovers at a small value instead of decaying further (\cref{fig:floor}). So why does the sharp component of GD's half-step $\hat{\delta}_t$ remain small but nonzero, whereas rod flow's $\delta$ plummets toward zero?

The difference turns out to have nothing to do with the dynamics. Rather, it's due to the imprecision of floating-point arithmetic on computers. Each time the weights are stored during gradient descent, rounding due to finite precision injects a small but nonzero noise into the sharp component of $\hat{\delta}_t$. In our setting, it's on the order of $10^{-8}$ for float32. This small rounding noise is what causes gradient descent to self-stabilize earlier than rod flow.

To replicate the effect of finite floating-point precision, we impose a floor on $\delta$. With this floor in place, rod flow tracks the discrete iterates accurately, including the timing of the onset of self-stabilization.

There is precedent for adding a noise floor. Edge flow \citep{marion2026edge}, an alternative continuous-time model of gradient descent at the edge of stability, implements its own version of a noise floor. More details are provided in \cref{app:floor}.

\subsection{Covariance Versus Second Moment}
\label{sec:chain-rule}

\begin{figure}[t]
	\centering
	\includegraphics[width=\textwidth]{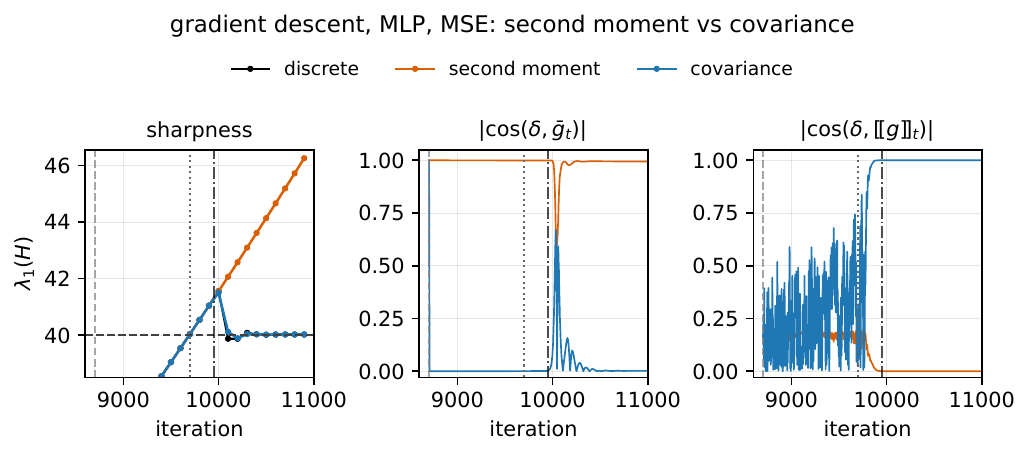}
	\caption{\textbf{Second Moment Versus Covariance.} Second-moment and covariance rod flows on an MLP ($\eta = 0.05$). \textbf{Left:} Second-moment rod flow fails to self-stabilize at the transition, whereas covariance rod flow succeeds. \textbf{Middle and right:} Alignment of each flow's extent with the mean gradient $\gbar_t$ and with the gradient half-difference $\gjump_t$ of the discrete iterates. The second-moment rod stays aligned with the mean gradient, while the covariance rod correctly aligns with the gradient half-difference at the onset.}
	\label{fig:covsm-flows}
\end{figure}

As observed by \citet{marion2026edge}, the original formulation of rod flow \citep{regis2026rod} was unable to track the onset of self-stabilization even though it could track the GD iterates if it was initialized in the steady-state edge of stability regime. But the new formulation of rod flow is able to track the iterates through all phases of training (\cref{fig:covsm-flows}). What's the key difference between the two formulations?

It helps to compare their respective fixed points for the extent. For the edge-coupling-derived formulation of rod flow, we have that the fixed point for $\delta$ is given by the half-difference of the gradient, scaled by $\eta/2$. We can then take the outer product to get the fixed point for $\Sigma_{\text{EC}} = \delta \otimes \delta$.
\begin{equation}
\Sigma^\ast_{\text{EC}} = \tfrac{\eta^2}{4}\gjump^{\otimes 2}
\end{equation}
For the original formulation of rod flow, we have that the fixed point of $\Sigma_{\text{SM}}$ is given by the second moment of the gradient.
\begin{equation}
\Sigma^\ast_{\text{SM}} = \tfrac{\eta^2}{4} \overline{g^{\otimes 2}}
\end{equation}
We can express the second moment of the gradient in terms of the average gradient and the gradient half-difference.
\begin{equation}
\Sigma^\ast_{\text{SM}} =\tfrac{\eta^2}{4} [\bar{g}^{\otimes 2} + \gjump^{\otimes 2}]
\end{equation}
This is the key difference between the two formulations: in the original rod flow, $\Sigma$ tracks the \textit{second moment} of the gradient, whereas in the formulation derived from the edge coupling, it tracks the gradient's \textit{covariance}.

\begin{figure}[t]
	\centering
	\includegraphics[width=0.7\textwidth]{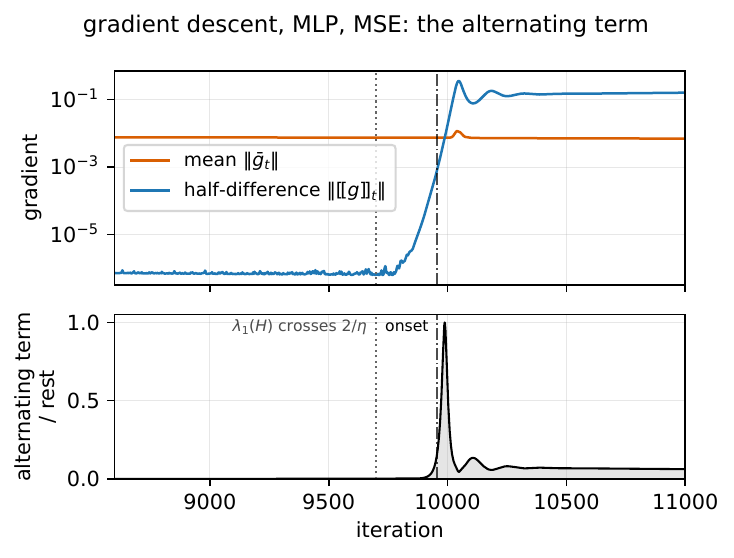}
	\caption{\textbf{The Alternating Term at the Onset.} Gradient descent on an MLP ($\eta = 0.05$). \textbf{Top:} Norms of the mean $\gbar_t$ and the half-difference $\gjump_t$ of consecutive gradients. Before the edge of stability, $\|\gbar_t\| \gg \|\gjump_t\|$. In the steady-state edge-of-stability regime, $\|\gbar_t\| \ll \|\gjump_t\|$. \textbf{Bottom:} Norm of the alternating term $\gbar_t\otimes\gjump_t + \gjump_t\otimes\gbar_t$ in $\dhat_t\otimes\dhat_t$, relative to that of the non-alternating terms $\gbar_t^{\otimes 2} + \gjump_t^{\otimes 2}$. The alternating term is negligible before the crossing but peaks just after the onset, when $\gbar_t$ and $\gjump_t$ are similar in magnitude.}
	\label{fig:covsm-premise}
\end{figure}

Why might it be better for $\Sigma$ to track the covariance of the gradient rather than its second moment? Recall that rod flow was motivated by the observation that the outer product $\hat{\delta} \otimes \hat{\delta}$ varies smoothly. But this is not strictly true. We can decompose the half-displacement into the average gradient $\bar{g}$ and half-difference gradient $\gjump$, which alternates each iteration:
\begin{equation}
	\hat{\delta}_t \approx \tfrac{\eta}{2} \left(\bar{g} + (-1)^t \gjump\right)
	\label{eq:dhat-decomp}
\end{equation}
Taking the outer product, we have:
\begin{equation}
	\hat{\delta}_t \otimes \hat{\delta}_t \approx \tfrac{\eta^2}{4} \left( \bar{g}^{\otimes 2} + \gjump^{\otimes 2} + (-1)^t [\bar{g} \otimes \gjump + \gjump \otimes \bar{g}] \right)
\end{equation}
We have that $\hat{\delta}_t \otimes \hat{\delta}_t$ varies smoothly when the alternating terms are small relative to the non-alternating terms. That's true when $\|\bar{g}\| \gg \|\gjump\|$ or $\|\bar{g}\| \ll \|\gjump\|$. But when $\bar{g}$ and $\gjump$ are of similar magnitude, as is the case at the onset of self-stabilization (\cref{fig:covsm-premise}), the alternating terms are comparable in magnitude to the non-alternating terms, violating the assumption that $\hat{\delta}_t \otimes \hat{\delta}_t$ varies smoothly. The edge-coupling-derived rod flow does not have this issue, since its $\delta$ tracks only $\gjump$.

Removing the slow-drift component also means that the edge-coupling-derived rod flow \textit{only} encodes the oscillations---which makes a certain amount of physical sense.

\subsection{Endpoint Rod Flow}

Recall the uncentered edge coupling:
\begin{equation*}
	\Ac_\eta(w, w') = L(w) + L(w') - \tfrac{1}{2\eta}\,\norm{w - w'}^2 . \tag{\ref{eq:A-def}}
\end{equation*}
One can actually construct a rod flow using the uncentered edge coupling:
\begin{equation}
	\boxed{
		\ddt{}\begin{pmatrix} w \\ w' \end{pmatrix} = - \eta \begin{pmatrix} 0 & I_d \\ I_d & 0 \end{pmatrix} \nabla_{(w, w')} \mathcal{A}
	}
\end{equation}
Expressing the ODEs for $w$ and $w'$ separately:
\begin{align}
	\ddt{w} &= -(w - [w' -\eta \nabla L(w')]) \\
	\ddt{w'} &= -(w' - [w -\eta \nabla L(w)])
\end{align}
Rather than tracking the midpoint $\bar{w}$ and the extent $\delta$, this formulation tracks the two endpoints of the rod directly. Therefore, an apt name for this formulation is \textit{endpoint rod flow}. We can then call the $(\bar{w}, \delta)$ formulation \textit{midpoint rod flow}.

Each endpoint regresses toward a ``target'' created by the other endpoint. This target simultaneously ensures that the two endpoints are the correct distance apart \textit{and} that both are rolling downhill in accordance with the average gradient.

Although endpoint rod flow appears superficially different from the midpoint formulation of rod flow, the two are equivalent---as follows from a straightforward application of the chain rule. Consequently, they perform identically in experiments (\cref{sec:exp-results}).

\subsection{RMSProp}
\label{sec:rmsprop-rod}

RMSProp \citep{tieleman2012rmsprop} can be considered Adam without momentum. It is an adaptive modification of gradient descent: instead of updating the parameters using the raw gradient, RMSProp maintains an exponential moving average $\nu$ of the componentwise squared gradients and scales each component of the update by the inverse square root of $\nu$. The preconditioner $P$ is defined as
\begin{equation}
	P(\nu) = \diag(\sqrt{\nu}) + \epsilon I_d,
\end{equation}
where $\epsilon > 0$ is a small numerical constant.

Let $v^{\odot 2}$ denote the componentwise square of a vector $v$. The update equations for RMSProp are
\begin{equation}
	\begin{aligned}
		\nu_{t+1} &= \beta_2\,\nu_t + (1-\beta_2)\,\nabla L(w_t)^{\odot2}, \\
		w_{t+1}   &= w_t - \eta\,P_{t+1}^{-1}\,\nabla L(w_t),
	\end{aligned}
	\label{eq:rmsprop}
\end{equation}
where $\beta_2 \in [0,1)$ and $P_t = P(\nu_t)$.

We can define an edge coupling for RMSProp by slightly modifying the edge coupling for gradient descent. The main difference is that the preconditioner $P(\nu)$ now acts as a metric tensor, changing the norm used to measure the displacement between consecutive iterates. Because $P$ depends on the second-moment estimate $\nu$, the edge coupling must also take $\nu$ as an extra argument:
\begin{equation}
	\mathcal{A}_{\eta, \nu}(w, w')
	=
	L(w) + L(w') - \tfrac{1}{2\eta}\,(w - w')^\top P(\nu)\,(w - w').
\end{equation}
Similarly, we can also derive a centered edge coupling for RMSProp:
\begin{equation}
	\Psi_{\eta, \nu}(\bar{w}, \delta)
	=
	\tfrac{1}{2}\bigl[L(\bar{w} + \delta) + L(\bar{w} - \delta)\bigr]
	- \tfrac{1}{\eta}\,\delta^\top P(\nu)\,\delta.
\end{equation}
As with gradient descent, we can derive a rod flow for RMSProp from the centered edge coupling. However, with RMSProp, only the position variable has rod structure. Because squaring kills the sign flip, the second moment $\nu$ doesn't oscillate at the edge of stability. So $\bar{\nu}$ evolves as a smooth auxiliary variable.
Recall that
\begin{equation}
	\overline{g^{\odot 2}} = \gbar^{\odot 2} + \gjump^{\odot 2}.
\end{equation}
We can then derive the following rod flow for RMSProp:
\begin{equation}
	\boxed{
	\begin{aligned}
		\ddt{}\begin{pmatrix} \bar{w} \\ \delta \end{pmatrix}
		&= \eta \begin{pmatrix} -P(\bar{\nu})^{-1} & 0 \\ 0 & P(\bar{\nu})^{-1} \end{pmatrix} \nabla_{(\bar{w}, \delta)} \Psic_{\eta,\nu} \\
		\ddt{\nubar}  &= (1-\beta_2)\Bigl[\gbar^{\odot2} + \gjump^{\odot2} - \nubar\Bigr].
	\end{aligned}}
\end{equation}
Notice how simple the interpretation of $P$ as a metric tensor is. We have that the identity matrix in the rod flow ODEs for GD is replaced by $P^{-1}$. Indeed, it \textit{must} be the inverse that appears here because it acts on the gradient---which is a covector.

For the individual ODEs, we have:
\begin{equation}
		\begin{aligned}
			\ddt{\wbar}   &= -\eta P^{-1}\gbar, \\
			\ddt{\delta}  &= \eta P^{-1}\gjump - 2\delta, \\
			\ddt{\nubar}  &= (1-\beta_2)\Bigl[\gbar^{\odot2} + \gjump^{\odot2} - \nubar\Bigr].
		\end{aligned}
	\label{eq:rmsprop-rodflow}
\end{equation}
For the endpoint rod flow, we have:
\begin{equation}
	\boxed{
	\begin{aligned}
		\ddt{}\begin{pmatrix} w \\ w' \end{pmatrix}
		&= -\eta \begin{pmatrix} 0 & P(\nubar)^{-1} \\ P(\nubar)^{-1} & 0 \end{pmatrix} \nabla_{(w, w')} \mathcal{A}_{\eta,\nubar}, \\
		\ddt{\nubar}  &= (1-\beta_2)\bigl[\tfrac12\bigl(\nabla L(w)^{\odot 2} + \nabla L(w')^{\odot 2}\bigr) - \nubar\bigr].
	\end{aligned}}
\end{equation}
Writing out the individual ODEs:
\begin{equation}
	\begin{aligned}
		\ddt{w}  &= -\bigl(w - \bigl[w' - \eta P^{-1}\nabla L(w')\bigr]\bigr), \\
		\ddt{w'} &= -\bigl(w' - \bigl[w - \eta P^{-1}\nabla L(w)\bigr]\bigr), \\
		\ddt{\nubar}  &= (1-\beta_2)\bigl[\tfrac12\bigl(\nabla L(w)^{\odot 2} + \nabla L(w')^{\odot 2}\bigr) - \nubar\bigr].
	\end{aligned}
	\label{eq:rmsprop-endpoint}
\end{equation}

%% file: sections/05-rod-flows-family.tex
\section{Rod Flow for Momentum Methods}
\label{sec:rod-flows-family}

We derive the phase-space edge coupling of heavy ball. We then use the edge coupling to obtain rod flows for heavy ball and Adam.

\subsection{The Phase-Space Edge Coupling}
\label{sec:phase-space-edge-coupling}

Let $w$ and $m$ denote the position and momentum, respectively. The update equations for heavy ball momentum \citep{polyak1964some} are given as:
\begin{align}
m_{t+1} &= \beta m_t + (1-\beta) \nabla L(w_t) \\
w_{t+1} &= w_t - \eta m_{t+1}
\end{align}
where $\beta \in [0,1)$ is the momentum coefficient.

Let $z = (w, m)$ denote a point in phase space. Heavy ball admits a phase-space edge coupling: a functional $\mathcal{B}_{\eta, \beta}(z, z')$ of two phase-space points whose critical points encode the fixed points and period-two orbits of heavy ball \citep{regis2026variational}. It's given as:
\begin{equation}
	\mathcal{B}_{\eta, \beta}(z, z') = L(w) + L(w') - \tfrac{1}{2\eta} \|w - w'\|^2 + \tfrac{\beta}{1-\beta} (m-m') \cdot (w - w') - \tfrac{\eta \beta}{1-\beta} m \cdot m'.
	\label{eq:hb-B}
\end{equation}
While this property can be verified by direct computation, we will show how one can \emph{derive} the phase-space edge coupling. Furthermore, our approach applies to all ``structured gradient methods''---a family of optimization methods that includes every optimizer in the Adam family.

\paragraph{Structured gradient methods.} Consider a discrete-time optimizer whose update rule is of the form
\begin{equation}
	z_{t+1} = z_t + \Kmat\nabla\mathcal{L}(z_t),
	\label{eq:structured-step}
\end{equation}
where $\mathcal{L}$ is the \emph{generating function} and $\Kmat$ is the \emph{structure matrix}. We require that the structure matrix $\Kmat$ be both constant and symmetric. We will call such an optimizer a \emph{structured gradient method}. The edge coupling is given as:
\begin{equation}
	\boxed{
		\mathcal{B}(z, z') = \mathcal{L}(z) + \mathcal{L}(z') + \tfrac{1}{2}(z - z')^\top\Kmat^{-1}(z - z').
	}
	\label{eq:structured-coupling}
\end{equation}
Gradient descent corresponds to $\mathcal{L} = L$ and $\Kmat = -\eta I_d$. RMSProp, with the second moment $\nu$ frozen, corresponds to $\mathcal{L} = L$ and $\Kmat = -\eta P(\nu)^{-1}$.

Let $\Phi$ denote the update map:
\begin{equation}
	\Phi(z) = z + \Kmat\nabla\mathcal{L}(z).
\end{equation}
The partial gradients of the edge coupling are
\begin{equation}
	\nabla_z\mathcal{B}(z, z') = \Kmat^{-1}\bigl(\Phi(z) - z'\bigr),
	\qquad
	\nabla_{z'}\mathcal{B}(z, z') = \Kmat^{-1}\bigl(\Phi(z') - z\bigr).
	\label{eq:residual-identity}
\end{equation}
By inspection, one can see that $\nabla_{(z,z')}\mathcal{B} = 0$ \textit{precisely} when $z' = \Phi(z)$ and $z = \Phi(z')$. As desired, the critical points of $\mathcal{B}$ encode the fixed points and period-two orbits of the optimizer.

Now consider heavy ball momentum. Heavy ball can be expressed as a structured gradient method with generating function and structure matrix:
\begin{equation}
	\mathcal{L}(w, m) = L(w) - \tfrac{\eta\beta}{2(1-\beta)}\norm{m}^2,
	\qquad
	\Kmat =
	\begin{pmatrix}
		-\eta(1-\beta)I_d & (1-\beta)I_d \\
		(1-\beta)I_d & \tfrac{(1-\beta)^2}{\eta\beta}I_d
	\end{pmatrix}.
	\label{eq:hb-generating}
\end{equation}
By substituting this $\mathcal{L}$ and $\Kmat$ into our formula for the edge coupling, we obtain the phase-space edge coupling for heavy ball.

Note that while heavy ball can be expressed as a damped Hamiltonian flow, the structure matrix in that formulation is \emph{not} symmetric, so it doesn't yield a valid edge coupling. In fact, for the case of heavy ball, the generating function and symmetric structure matrix above are unique up to a trivial simultaneous rescaling. More details on edge couplings for structured gradient methods are given in \cref{app:structured}.

\subsection{Heavy Ball Rod Flow}
\label{sec:hb-rod}

Define the centered coordinates:
\begin{equation}
	\bar{z} = \begin{pmatrix} \bar{w} \\ \bar{m} \end{pmatrix} = \frac{z + z'}{2}, \qquad \Delta = \begin{pmatrix} \delta \\ \gamma \end{pmatrix} = \frac{z - z'}{2}
\end{equation}
and the centered phase-space edge coupling:
\begin{align}
	\Upsilon_{\eta, \beta} (\bar{z}, \Delta) &:= \tfrac{1}{2}\mathcal{B}_{\eta, \beta} (\bar{z} + \Delta, \bar{z} - \Delta) \\ &= \tfrac{1}{2}[L(\bar{w} + \delta) + L(\bar{w} - \delta)] - \tfrac{1}{\eta} \|\delta\|^2 + \tfrac{2\beta}{1-\beta} \gamma \cdot \delta - \tfrac{\eta \beta}{2(1-\beta)}(\|\bar{m}\|^2 - \|\gamma\|^2)
\end{align}
For gradient descent, we obtained the rod flow by performing gradient descent--ascent on the centered edge coupling. Using a similar approach, we can derive the heavy ball rod flow---though we must now incorporate the structure matrix $K$. By applying $\Kmat$ to the gradient in the midpoint $\zbar$ and $-\Kmat$ to the gradient in the extent $\Delta$, we obtain the following rod flow:
\begin{equation}
	\boxed{
		\ddt{}\begin{pmatrix} \bar{z} \\ \Delta \end{pmatrix}
		=
		\begin{pmatrix} K & 0 \\ 0 & -K \end{pmatrix}
		\nabla_{(\bar{z}, \Delta)} \Upsilon
	}
\end{equation}
Writing out the individual ODEs:
\begin{equation}
	\begin{aligned}
		\ddt{\wbar} &= -\eta\bigl[\beta\mbar+(1-\beta)\gbar\bigr], \\
		\ddt{\mbar} &= (1-\beta)\bigl[\gbar-\mbar\bigr], \\
		\ddt{\delta} &= -2\delta+\eta\bigl[\beta\gamma+(1-\beta)\gjump\bigr], \\
		\ddt{\gamma} &= -(1+\beta)\gamma-(1-\beta)\gjump.
	\end{aligned}
	\label{eq:hb-rodflow}
\end{equation}
Note that at $\beta = 0$ the heavy ball rod flow reduces to the gradient descent rod flow.

For the endpoint rod flow, we have:
\begin{equation}
	\boxed{
		\ddt{}\begin{pmatrix} z \\ z' \end{pmatrix}
		=
		\begin{pmatrix} 0 & \Kmat \\ \Kmat & 0 \end{pmatrix}
		\nabla_{(z, z')} \mathcal{B}_{\eta, \beta}
	}
\end{equation}
Writing out the individual ODEs:
\begin{equation}
	\begin{aligned}
		\ddt{w} &= -\bigl(w - \bigl[w' - \eta\bigl(\beta m' + (1-\beta)\nabla L(w')\bigr)\bigr]\bigr), \\
		\ddt{m} &= -\bigl(m - \bigl[\beta m' + (1-\beta)\nabla L(w')\bigr]\bigr), \\
		\ddt{w'} &= -\bigl(w' - \bigl[w - \eta\bigl(\beta m + (1-\beta)\nabla L(w)\bigr)\bigr]\bigr), \\
		\ddt{m'} &= -\bigl(m' - \bigl[\beta m + (1-\beta)\nabla L(w)\bigr]\bigr).
	\end{aligned}
	\label{eq:hb-endpoint}
\end{equation}
Each endpoint regresses toward the heavy-ball image of the other endpoint.

An analogous rod flow can be derived for Nesterov momentum using the
Nesterov phase-space edge coupling. More details are provided in \cref{app:nesterov}.

\subsection{Adam}
\label{sec:adam-rod}

Adam \citep{kingma2015adam} combines momentum with the adaptive preconditioner of RMSProp. Adam's update equations are:
\begin{equation}
	\begin{aligned}
		m_{t+1}
		&= \beta_1 m_t + (1-\beta_1)\nabla L(w_t), \\
		\nu_{t+1}
		&= \beta_2 \nu_t + (1-\beta_2)\nabla L(w_t)^{\odot 2}, \\
		w_{t+1}
		&= w_t - \eta P_{t+1}^{-1}m_{t+1},
	\end{aligned}
\end{equation}
where $P(\nu) = \diag(\sqrt{\nu}) + \epsilon I_d$---the same as in RMSProp. For simplicity, we omitted the bias correction, though it can be incorporated into the formalism without much difficulty. Details are provided in \cref{app:bias-correction}.

Treating $\nu$ as fixed, Adam is a structured gradient method on phase space. Its generating function and structure matrix are those of heavy ball except that the preconditioner $P$ replaces the identity matrix as the acting metric tensor.
\begin{equation}
	\mathcal{L}_\nu(w, m) = L(w) - \tfrac{\eta\beta_1}{2(1-\beta_1)}m^\top P^{-1}m,
	\qquad
	\Kmat_{\nu}
	=
	\begin{pmatrix}
		-\eta(1-\beta_1)P(\nu)^{-1} & (1-\beta_1)I_d \\
		(1-\beta_1)I_d &
		\tfrac{(1-\beta_1)^2}{\eta\beta_1}P(\nu)
	\end{pmatrix}.
	\label{eq:adam-K}
\end{equation}
For the edge coupling, we have:
\begin{equation}
	\mathcal{B}_{\eta,\beta_1,\nu}(z,z')
	= L(w) + L(w')
	- \tfrac{1}{2\eta}(w-w')^\top P(w-w')
	+ \tfrac{\beta_1}{1-\beta_1}(m-m')\cdot(w-w')
	- \tfrac{\eta\beta_1}{1-\beta_1}m^\top P^{-1}m'.
\end{equation}
And for the centered edge coupling, we have:
\begin{equation}
		\Upsilon_{\eta,\beta_1,\nu}(\bar{z},\Delta)
= \tfrac{1}{2}\bigl[L(\bar{w}+\delta)+L(\bar{w}-\delta)\bigr]
		- \tfrac{1}{\eta}\delta^\top P\delta
		+ \tfrac{2\beta_1}{1-\beta_1}\gamma\cdot\delta
		- \tfrac{\eta\beta_1}{2(1-\beta_1)}
		\bigl[\bar{m}^\top P^{-1}\bar{m}
		- \gamma^\top P^{-1}\gamma\bigr].
\end{equation}
As with RMSProp, we treat the second-moment $\bar{\nu}$ as a smooth auxiliary variable. We then obtain the following rod flow for Adam:
\begin{equation}
	\boxed{
		\begin{aligned}
		\ddt{}
		\begin{pmatrix}
			\bar{z} \\ \Delta
		\end{pmatrix}
		&=
		\begin{pmatrix}
			\Kmat_{\bar{\nu}} & 0 \\
			0 & -\Kmat_{\bar{\nu}}
		\end{pmatrix}
		\nabla_{(\bar{z},\Delta)}
		\Upsilon_{\eta,\beta_1,\bar{\nu}}, \\
	\ddt{\nubar} &= (1-\beta_2)
	\bigl[\gbar^{\odot 2}+\gjump^{\odot 2}-\nubar\bigr].
	\end{aligned}}
\end{equation}
Writing out the individual ODEs:
\begin{equation}
	\begin{aligned}
		\ddt{\wbar} &= -\eta P^{-1}\bigl[\beta_1\mbar+(1-\beta_1)\gbar\bigr], \\
		\ddt{\mbar} &= (1-\beta_1)\bigl[\gbar-\mbar\bigr], \\
		\ddt{\nubar} &= (1-\beta_2)
		\bigl[\gbar^{\odot 2}+\gjump^{\odot 2}-\nubar\bigr], \\
		\ddt{\delta} &= -2\delta+\eta P^{-1}\bigl[\beta_1\gamma+(1-\beta_1)\gjump\bigr], \\
		\ddt{\gamma} &= -(1+\beta_1)\gamma-(1-\beta_1)\gjump.
	\end{aligned}
	\label{eq:adam-rodflow}
\end{equation}
For the endpoint rod flow, we have:
\begin{equation}
	\boxed{
		\begin{aligned}
		\ddt{}\begin{pmatrix} z \\ z' \end{pmatrix}
		&=
		\begin{pmatrix} 0 & \Kmat_{\nubar} \\ \Kmat_{\nubar} & 0 \end{pmatrix}
		\nabla_{(z, z')} \mathcal{B}_{\eta, \beta_1, \nubar}, \\
		\ddt{\nubar} &= (1-\beta_2)\bigl[\tfrac12\bigl(\nabla L(w)^{\odot 2} + \nabla L(w')^{\odot 2}\bigr) - \nubar\bigr].
		\end{aligned}}
\end{equation}
Writing out the individual ODEs:
\begin{equation}
	\begin{aligned}
		\ddt{w} &= -\bigl(w - \bigl[w' - \eta P^{-1}\bigl(\beta_1 m' + (1-\beta_1)\nabla L(w')\bigr)\bigr]\bigr), \\
		\ddt{m} &= -\bigl(m - \bigl[\beta_1 m' + (1-\beta_1)\nabla L(w')\bigr]\bigr), \\
		\ddt{\nubar} &= (1-\beta_2)\bigl[\tfrac12\bigl(\nabla L(w)^{\odot 2} + \nabla L(w')^{\odot 2}\bigr) - \nubar\bigr], \\
		\ddt{w'} &= -\bigl(w' - \bigl[w - \eta P^{-1}\bigl(\beta_1 m + (1-\beta_1)\nabla L(w)\bigr)\bigr]\bigr), \\
		\ddt{m'} &= -\bigl(m' - \bigl[\beta_1 m + (1-\beta_1)\nabla L(w)\bigr]\bigr).
	\end{aligned}
	\label{eq:adam-endpoint}
\end{equation}
One can also derive an edge coupling and rod flow for AdamW, a modification of Adam whose update rule decays the weights towards zero. The details are provided in \cref{app:adamw}.

%% file: sections/06-central-from-coupling.tex
\section{Central Flow and Edge Coupling}
\label{sec:central-from-coupling}

We derive central flow from the centered edge coupling. Whereas the extent of rod flow is governed by an ODE, the covariance matrix of the central flow acts as a Lagrange multiplier, constraining the dynamics to the feasible region.

\subsection{The Truncated Edge Coupling}
\label{sec:lift-truncate}

Losses in machine learning have been found to be well described by their local quadratic approximation \citep{meterez2026defense}. Applying this approximation about the midpoint $\bar{w}$, we have that:
\begin{equation}
	\tfrac{1}{2}[L(\bar{w} + \delta) + L(\bar{w} - \delta)] \approx L(\bar{w}) + \tfrac{1}{2} \delta^\top H(\bar{w})\delta.
\end{equation}
Substituting this approximation into the centered edge coupling, we obtain the \emph{truncated edge coupling}:
\begin{equation}
	\widetilde{\Psi}(\bar{w}, \delta) = L(\bar{w}) + \tfrac{1}{2} \delta^\top H(\bar{w})\delta - \tfrac{1}{\eta} \|\delta\|^2 .
\end{equation}
Notice that the quadratic terms of the truncated edge coupling depend on $\delta$ only through $\delta \otimes \delta$:
\begin{equation}
	\tfrac12 \delta^\top H(\wbar)\delta - \tfrac1\eta\|\delta\|^2
	= \tfrac12\inner{H(\wbar) - \tfrac{2}{\eta}I_d}{\delta \otimes \delta}.
\end{equation}
We will now replace the positive semidefinite rank-one matrix $\delta \otimes \delta$ with a general positive semidefinite matrix $\Sigma$, which we call the \emph{covariance matrix}. The truncated edge coupling can then be written as a function of $\Sigma$:
\begin{equation}
	\Psit(\wbar, \Sigma) = L(\wbar) + \tfrac12\inner{H(\wbar) - \tfrac{2}{\eta}I_d}{\Sigma}.
	\label{eq:Psit}
\end{equation}
Why may we consider general positive semidefinite matrices $\Sigma$ rather than only positive semidefinite rank-one matrices? Recall that every positive semidefinite matrix can be written as a convex combination of positive semidefinite rank-one matrices. Since $\Psit$ is linear in $\Sigma$, its supremum over all positive semidefinite matrices equals its supremum over positive semidefinite rank-one matrices, so the maximum value is unchanged.

There is also a physical reason to allow arbitrary positive semidefinite $\Sigma$. As we will see in \cref{sec:ranks}, $\Sigma$ arises as the time average $\langle \delta \otimes \delta \rangle$ over intermediate time scales---and a time average of rank-one positive semidefinite matrices can be arbitrary rank.

\subsection{Covariance Matrix as Lagrange Multiplier}
\label{sec:multiplier}

Recall that, when the endpoint Hessians agree, stable period-two orbits correspond to min--max critical points of the centered edge coupling in $\bar{w}$ and $\delta$. When we replace $\delta \otimes \delta$ by $\Sigma$, we retain the maximization criterion. However, unlike with rod flow, central flow does not determine the dynamics of $\Sigma$ by an ODE. Instead, $\Sigma$ is determined at each instant by a maximization problem parametrized by $\bar{w}$.

Notice that the truncated edge coupling is linear in $\Sigma$. Let $\Lambda = \tfrac{2}{\eta} I_d - H$ denote the slack. Since $L(\wbar)$ does not depend on $\Sigma$, the maximization over $\Sigma$ is governed entirely by the linear term
\begin{equation*}
	-\tfrac{1}{2}\langle \Lambda, \Sigma \rangle .
\end{equation*}
The argmax is determined by the sharpness $\lambda_1(H)$. When the sharpness is less than $2/\eta$, $\Lambda$ is positive definite and the argmax is the zero matrix. When the sharpness is greater than $2/\eta$, no argmax exists: the maximization problem is unbounded as any component of $\Sigma$ along the top eigenspace of $H$ can be increased without bound. When the sharpness is exactly $2/\eta$, the solution is not unique: any $\Sigma$ supported on the kernel of $\Lambda$ is a valid maximizer, since it contributes nothing to $\inner{\Lambda}{\Sigma}$. We then have that:
\begin{equation}
	\operatorname*{argmax}_{\Sigma \succeq 0} \widetilde{\Psi}(\bar{w}, \Sigma) =
	\begin{cases}
		\{0\}, & \lambda_1(H) < \tfrac{2}{\eta}, \\
		\mathcal{K}(\bar{w}), & \lambda_1(H) = \tfrac{2}{\eta}, \\
		\emptyset, & \lambda_1(H) > \tfrac{2}{\eta},
	\end{cases}
	\label{eq:sup}
\end{equation}
where
\begin{equation}
	\mathcal{K}(\bar{w}) := \{U X U^\top : X \succeq 0\}, \qquad X \in \mathbb{R}^{k \times k},
\end{equation}
is the \emph{critical face}. $U \in \mathbb{R}^{d \times k}$ is an orthonormal basis for the $k$-dimensional eigenspace of $H(\bar{w})$ at $2/\eta$. $\mathcal{K}$ represents the set of covariance matrices whose oscillations are confined to the critical subspace.

There are three conditions that \citet{Cohen+25CentralFlow} impose on the central flow:
\begin{equation}
	\Sigma \succeq 0, \qquad \Lambda \succeq 0, \qquad \langle \Lambda, \Sigma \rangle = 0 .
	\label{eq:dcp}
\end{equation}
The first condition requires the covariance matrix to be positive semidefinite. The second requires the slack to be positive semidefinite, which is equivalent to confining the flow to the feasible region $\Omega$. The third condition requires the covariance matrix $\Sigma$ to be orthogonal to the slack $\Lambda$. This encodes the fact that oscillations are confined to the critical subspace of $H$. We have that these three conditions are obtained by maximizing the truncated edge coupling over $\Sigma$.

Whereas rod flow encodes the oscillations in the state variable $\delta$, the covariance matrix $\Sigma$ of central flow acts as a \textit{Lagrange multiplier} that constrains the dynamics to the feasible region. Thus, the dynamics of central flow correspond to a constrained minimization problem rather than a min--max problem.

\subsection{Persistence}
\label{sec:persistence}

Central flow can be expressed as gradient flow on the truncated edge coupling:
\begin{align}
	\ddt{\bar{w}}
	&= -\eta \, \partial_{\bar{w}} \widetilde{\Psi}(\bar{w}, \Sigma) \\
	&= -\eta \nabla L(\bar{w})
	- \tfrac{\eta}{2} \nabla_{\bar{w}}
	\langle H(\bar{w}), \Sigma \rangle,
\end{align}
where $\Sigma$ is a maximizer of $\widetilde{\Psi}(\bar{w}, \cdot\,)$.
The resultant system defines a \emph{differential complementarity problem}:
an ordinary differential equation coupled to complementarity conditions.

When the sharpness is below $2/\eta$, the maximizer is $\Sigma = 0$---central flow reduces to gradient flow. However, when the sharpness equals $2/\eta$, the maximizer is no longer unique. How do we select $\Sigma$?

We select $\Sigma$ by invoking \emph{persistence}: feasibility and complementarity must hold not only at a single instant but along the entire trajectory. It's these dynamic constraints that determine $\Sigma$.

Define the rate of change of the slack:
\begin{align}
\dot{\Lambda}&= \tfrac{\dd}{\dd t}\Lambda(\bar{w}(t)) \\
&= - \nabla_{\bar{w}} H(\bar{w}) [\dot{\bar{w}}].
\end{align}
For the trajectory to remain feasible, we require that the slack not decrease along the sharp directions (where the slack is zero). These directions are given by the columns of $U$---which yields the condition:
\begin{equation}
	U^\top \dot{\Lambda} U \succeq 0 .
\end{equation}
This condition allows a critical eigenvalue to move below $2/\eta$, but prevents it from moving above $2/\eta$.

Complementarity requires that oscillations occur only in directions where the slack is zero. Persistence of complementarity then dictates that if a
direction develops a positive slack, it can no longer carry oscillations.
Any direction that carries oscillations must therefore not only have zero slack, but also stay at zero slack to
first order---$\langle \dot{\Lambda}, \Sigma \rangle = 0$. Rearranging, we obtain the following condition:
\begin{equation}
	\langle U^\top \dot{\Lambda} U, X \rangle = 0.
\end{equation}

Altogether, we have that the following conditions define the central flow for gradient descent:
\begin{equation}
	\boxed{
		\begin{aligned}
			\ddt{\bar{w}} &= -\eta \, \partial_{\bar{w}} \widetilde{\Psi}(\bar{w}, \Sigma), \qquad \Sigma = U X U^\top, \\[4pt]
			X &\succeq 0, \qquad U^\top \dot{\Lambda} U \succeq 0, \qquad \langle U^\top \dot{\Lambda} U, X \rangle = 0.
		\end{aligned}
	}
\end{equation}
One can show that the SDCP is equivalent to a convex quadratic program for $X$. Further details are provided in \cref{app:qp}.

\subsection{RMSProp}
\label{sec:rmsprop-central}

We can also derive a central flow for RMSProp. The truncated edge coupling is given as:
\begin{equation}
	\widetilde{\Psi}_{\eta,\nubar}(\wbar,\Sigma)
	= L(\wbar)
	+ \tfrac12\inner{H(\wbar)-\tfrac{2}{\eta}P(\nubar)}{\Sigma}.
\end{equation}
Define the slack:
\begin{equation}
	\Lambda_{\mathrm{RMSProp}}
	:= \tfrac{2}{\eta}P(\nubar)-H(\wbar).
\end{equation}
Maximization over $\Sigma\succeq0$ gives the static conditions:
\begin{equation}
	\Sigma\succeq0,
	\qquad
	\Lambda_{\mathrm{RMSProp}}\succeq0,
	\qquad
	\inner{\Lambda_{\mathrm{RMSProp}}}{\Sigma}=0.
\end{equation}
Let $U$ be an orthonormal basis for the kernel of
$\Lambda_{\mathrm{RMSProp}}$. Then $\Sigma=UXU^\top$ for some
$X\succeq0$.

The midpoint and second-moment evolve according to:
\begin{equation}
	\begin{aligned}
		\ddt{\wbar}
		&= -\eta P^{-1}
		\Bigl[\nabla L+\tfrac12\nabla_{\wbar}\inner{H}{\Sigma}\Bigr],\\
		\ddt{\nubar}
		&= (1-\beta_2)
		\Bigl[(\nabla L)^{\odot2}
		+\operatorname{diag}(H\Sigma H)-\nubar\Bigr],
	\end{aligned}
	\label{eq:rmsprop-central}
\end{equation}
where we make the approximation
\begin{equation}
	\gbar^{\odot2}+\gjump^{\odot2}
	\approx (\nabla L)^{\odot2}+\operatorname{diag}(H\Sigma H),
\end{equation}
which follows from the local quadratic approximation of the loss about $\wbar$ and by replacing $\delta\otimes\delta$ with $\Sigma$.
For the time derivative of the slack, we have:
\begin{equation}
	\dot{\Lambda}_{\mathrm{RMSProp}}
	= \tfrac{2}{\eta}\nabla_{\bar{\nu}}P(\nubar)[\dot{\nubar}]
	- \nabla_{\bar{w}}H(\wbar)[\dot{\wbar}]
\end{equation}
At fixed $(\wbar,\nubar)$, both terms are affine in $\Sigma$.
Persistence of feasibility and complementarity gives:
\begin{equation}
	U^\top\dot{\Lambda}_{\mathrm{RMSProp}}U\succeq0,
	\qquad
	\langle U^\top\dot{\Lambda}_{\mathrm{RMSProp}}U, X \rangle=0.
\end{equation}
These conditions define a semidefinite
complementarity problem that determines the covariance $\Sigma$. Altogether, we have that the following conditions define the central flow for RMSProp:
\begin{equation}
	\boxed{
		\begin{aligned}
			\ddt{\bar{w}} &= -\eta P(\nubar)^{-1} \partial_{\bar{w}} \widetilde{\Psi}_{\eta,\nubar}(\bar{w}, \Sigma), \qquad \Sigma = U X U^\top, \\[4pt]
			\ddt{\nubar} &= (1-\beta_2)\Bigl[(\nabla L)^{\odot2} + \operatorname{diag}(H\Sigma H) - \nubar\Bigr], \\[4pt]
			X &\succeq 0, \qquad U^\top \dot{\Lambda}_{\mathrm{RMSProp}} U \succeq 0, \qquad \langle U^\top \dot{\Lambda}_{\mathrm{RMSProp}} U, X \rangle = 0.
		\end{aligned}
	}
\end{equation}
This is precisely the RMSProp central flow of \citet{Cohen+25CentralFlow}.

%% file: sections/07-central-flows-momentum.tex
\section{Central Flow for Momentum Methods}
\label{sec:central-flows-momentum}

Using the phase-space edge coupling, we derive novel central flows for heavy ball and Adam.

\subsection{Eliminating the Momentum Oscillation}
\label{sec:hb-central}

We would like to derive a heavy-ball central flow. To do so, it helps to first eliminate the momentum oscillations $\gamma$, so that the edge coupling is expressed only in terms of $(\bar{z}, \delta)$.

Recall the centered phase-space edge coupling $\Upsilon_{\eta,\beta}$ of heavy ball. The momentum half-difference $\gamma$ enters it through only two terms:
\begin{equation}
	\frac{2\beta}{1-\beta}\gamma\cdot\delta
	+\frac{\eta\beta}{2(1-\beta)}\|\gamma\|^2
	=
	\frac{\eta\beta}{2(1-\beta)}
	\left\|\gamma+\frac{2}{\eta}\delta\right\|^2
	-\frac{2\beta}{\eta(1-\beta)}\|\delta\|^2.
\end{equation}
As period-two orbits are critical points of $\Upsilon_{\eta,\beta}$, they are stationary in $\gamma$. Since $\Upsilon_{\eta,\beta}$ is strictly convex in $\gamma$, stationarity has the unique solution
\begin{equation}
	\gamma = -\frac{2}{\eta}\delta ,
	\label{eq:hb-gamma}
\end{equation}
which minimizes $\Upsilon_{\eta,\beta}$ over $\gamma$. For period-two orbits, the momentum oscillation is not an independent degree of freedom: it is fixed by the position oscillation.

Define the \emph{reduced edge coupling}:
\begin{equation}
	\begin{aligned}
		\Upsilon^{\mathrm{red}}_{\eta,\beta}(\bar{z},\delta)
		&:= \min_{\gamma}\,\Upsilon_{\eta,\beta}\bigl(\bar{z},(\delta,\gamma)\bigr) \\
		&= \tfrac12\bigl[L(\bar{w}+\delta)+L(\bar{w}-\delta)\bigr]
		-\tfrac{1+\beta}{\eta(1-\beta)}\|\delta\|^2
		-\tfrac{\eta\beta}{2(1-\beta)}\|\bar{m}\|^2 .
	\end{aligned}
	\label{eq:hb-red}
\end{equation}
Note that the critical points of the reduced edge coupling exactly correspond to those of the centered phase-space edge coupling.

\subsection{The Truncated Reduced Edge Coupling}

As with gradient descent, we approximate the loss about the midpoint $\bar{w}$ by its local quadratic approximation:
\begin{equation}
	\tfrac{1}{2}[L(\bar{w} + \delta) + L(\bar{w} - \delta)] \approx L(\bar{w}) + \tfrac{1}{2} \delta^\top H(\bar{w}) \delta.
\end{equation}
Substituting into the reduced edge coupling and again replacing $\delta\otimes\delta$ with a general positive semidefinite matrix $\Sigma\succeq0$, we obtain the \textit{truncated reduced edge coupling}:
\begin{equation}
	\widetilde{\Upsilon}^{\mathrm{red}}_{\eta,\beta}
	(\bar{z},\Sigma)
	=
	L(\bar{w})
	-\tfrac{\eta\beta}{2(1-\beta)}\|\bar{m}\|^2
	-\tfrac12
	\langle \Lambda_{\mathrm{HB}}(\bar{w}),
	\Sigma\rangle,
\end{equation}
where
\begin{equation}
	\Lambda_{\mathrm{HB}}(\bar{w})
	:=
	\tfrac{2(1+\beta)}{\eta(1-\beta)}I_d-H(\bar{w}).
\end{equation}
The truncated edge coupling of heavy ball is quite similar to that of gradient descent---though there are two differences. First, the slack differs between heavy ball and gradient descent, reflecting the differing sharpness thresholds of the two methods. Second, the heavy-ball reduced edge coupling has an extra term proportional to the squared midpoint momentum.

Maximizing over $\Sigma\succeq0$, we obtain the following conditions:
\begin{equation}
	\Sigma\succeq0,
	\qquad
	\Lambda_{\mathrm{HB}}\succeq0,
	\qquad
	\langle \Lambda_{\mathrm{HB}},\Sigma\rangle=0.
\end{equation}
As was the case with gradient descent, the covariance matrix $\Sigma$ acts as a Lagrange multiplier---now constraining the dynamics to the feasible region of heavy ball.

Following a similar approach, one can also derive a central flow for Nesterov momentum. Details are provided in \cref{app:nesterov}.

\subsection{Persistence}
\label{sec:hb-persistence}

Taking derivatives of the truncated reduced edge coupling with respect to the midpoint variables gives:
\begin{align}
	\partial_{\bar{w}}
	\widetilde{\Upsilon}^{\mathrm{red}}_{\eta,\beta}
	&=
	\nabla L(\bar{w})
	+\tfrac{1}{2}\nabla_{\bar{w}}
	\langle H(\bar{w}),\Sigma\rangle,
	\\
	\partial_{\bar{m}}
	\widetilde{\Upsilon}^{\mathrm{red}}_{\eta,\beta}
	&=
	-\tfrac{\eta\beta}{1-\beta}\bar{m}.
\end{align}
As with the heavy ball rod flow, we apply the structure matrix $K$ to the gradient of the truncated reduced edge coupling to obtain the central flow ODE for heavy ball.
\begin{equation}
	\ddt{\bar{z}}
	=
	\Kmat\,
	\partial_{\bar{z}}
	\widetilde{\Upsilon}^{\mathrm{red}}_{\eta,\beta}
	(\bar{z},\Sigma).
\end{equation}
Writing the ODEs for $\bar{w}$ and $\bar{m}$ individually:
\begin{equation}
	\begin{aligned}
		\ddt{\bar{w}}
		&=-\eta
		\left[
		\beta\bar{m}
		+(1-\beta)\Bigl(
		\nabla L(\bar{w})
		+\tfrac{1}{2}\nabla_{\bar{w}}
		\langle H(\bar{w}),\Sigma\rangle
		\Bigr)
		\right],
		\\
		\ddt{\bar{m}}
		&=(1-\beta)
		\left[
		\nabla L(\bar{w})
		+\tfrac{1}{2}\nabla_{\bar{w}}
		\langle H(\bar{w}),\Sigma\rangle
		-\bar{m}
		\right].
	\end{aligned}
	\label{eq:hb-central}
\end{equation}
Note that the central flow ODEs for the midpoint variables are quite similar to the ODEs of continuous-time heavy-ball flow---with the only difference being that every gradient is now accompanied by an additional sharpness-reducing force.

As with the central flow for gradient descent, to determine $\Sigma$ we invoke persistence: feasibility and complementarity must hold along the trajectory.

Let $U$ be an orthonormal basis for the kernel of
$\Lambda_{\mathrm{HB}}$. Static complementarity requires:
\begin{equation}
	\Sigma=UXU^\top,
	\qquad X\succeq0.
\end{equation}
Computing the rate of change of the slack:
\begin{equation}
	\dot{\Lambda}_{\mathrm{HB}}
	= -\nabla_{\bar{w}}H(\bar{w})[\dot{\bar{w}}]
\end{equation}
Persistence of feasibility requires that the slack not decrease along sharp directions:
\begin{equation}
	U^\top\dot{\Lambda}_{\mathrm{HB}}U\succeq0.
\end{equation}
Persistence of complementarity requires that there be no oscillations in directions whose slack is increasing:
\begin{equation}
	\langle U^\top\dot{\Lambda}_{\mathrm{HB}}U,X\rangle=0.
\end{equation}
As with gradient descent, the persistence conditions define a semidefinite complementarity problem for $X$. Altogether, we have that the following conditions define the central flow for heavy ball:
\begin{equation}
	\boxed{
		\begin{aligned}
			\ddt{\bar{z}} &= \Kmat\, \partial_{\bar{z}} \widetilde{\Upsilon}^{\mathrm{red}}_{\eta,\beta}(\bar{z}, \Sigma), \qquad \Sigma = U X U^\top, \\[4pt]
			X &\succeq 0, \qquad U^\top \dot{\Lambda}_{\mathrm{HB}} U \succeq 0, \qquad \langle U^\top \dot{\Lambda}_{\mathrm{HB}} U, X \rangle = 0.
		\end{aligned}
	}
\end{equation}

\begin{figure}[t]
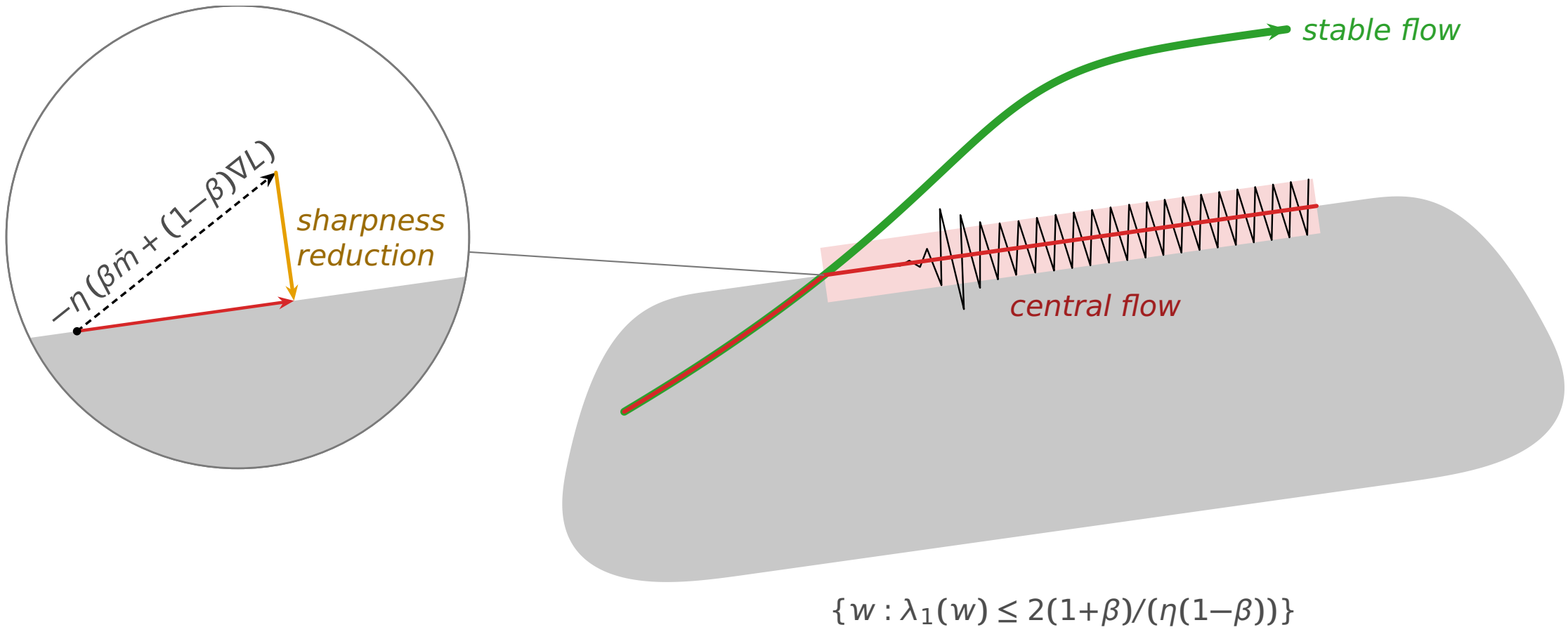

	\centering
	\includegraphicsnogroup{schematic/hb_central_flow.pdf}
	\caption{\textbf{Heavy-Ball Central Flow.} The gray region is the feasible region for heavy ball. While the stable flow leaves the feasible region, the central flow turns a corner at the boundary and follows it to a constrained minimum. \textbf{Inset:} The heavy-ball update points out of the feasible region. The sharpness-reduction force cancels its outward component, and what remains is the velocity of the central flow.}
	\label{fig:hb-central}
\end{figure}

\Cref{fig:hb-central} illustrates the central flow for heavy ball. As with gradient descent, it can be interpreted as a projected flow: at the boundary of the feasible region, the sharpness-reduction force cancels the outward component of the motion. What is projected, however, is the heavy-ball step rather than the negative gradient.

\subsection{Adam}
\label{sec:adam-central}

We can also obtain a central flow for Adam. With the second moment $\bar{\nu}$ held fixed, the centered phase-space edge coupling of Adam depends on $\gamma$ as:
\begin{equation}
		\frac{2\beta_1}{1-\beta_1}\gamma\cdot\delta
		+\frac{\eta\beta_1}{2(1-\beta_1)}\gamma^\top P^{-1}\gamma
		=
		\frac{\eta\beta_1}{2(1-\beta_1)}
		\left(\gamma+\frac{2}{\eta}P\delta\right)^\top
		P^{-1}
		\left(\gamma+\frac{2}{\eta}P\delta\right)
			-\frac{2\beta_1}{\eta(1-\beta_1)}\,\delta^\top P\delta .
\end{equation}
Stationarity in $\gamma$ gives $\gamma=-\tfrac{2}{\eta}P\delta$. Eliminating $\gamma$ from the centered phase-space edge coupling yields the reduced edge coupling:
\begin{equation}
	\Upsilon^{\mathrm{red}}_{\eta,\beta_1,\bar{\nu}}(\bar{z},\delta)
	=
	\tfrac12\bigl[L(\bar{w}+\delta)+L(\bar{w}-\delta)\bigr]
	-\tfrac{1+\beta_1}{\eta(1-\beta_1)}\,\delta^\top P\delta
	-\tfrac{\eta\beta_1}{2(1-\beta_1)}
	\bar{m}^\top P^{-1}\bar{m}.
\end{equation}
This closely parallels the reduced edge coupling for heavy ball, except that the preconditioner $P$ now acts as the metric tensor.
Replacing $\delta\otimes\delta$ with $\Sigma\succeq0$, the truncated edge coupling becomes:
\begin{equation}
	\widetilde{\Upsilon}^{\mathrm{red}}_{\eta,\beta_1,\bar{\nu}}
	(\bar{z},\Sigma)
	=
	L(\bar{w})
	-\tfrac{\eta\beta_1}{2(1-\beta_1)}
	\bar{m}^\top P^{-1}\bar{m}
	-\tfrac12
	\langle\Lambda_{\mathrm{Adam}},\Sigma\rangle,
\end{equation}
where the Adam slack is given as:
\begin{equation}
	\Lambda_{\mathrm{Adam}}(\bar{w},\bar{\nu})
	:=
	\tfrac{2(1+\beta_1)}{\eta(1-\beta_1)}P(\bar{\nu})
	-H(\bar{w}).
\end{equation}
Maximizing over $\Sigma\succeq0$ gives the following static conditions:
\begin{equation}
	\Sigma\succeq0,
	\qquad
	\Lambda_{\mathrm{Adam}}\succeq0,
	\qquad
	\langle\Lambda_{\mathrm{Adam}},\Sigma\rangle=0.
\end{equation}
Let $U$ be an orthonormal
basis for $\ker\Lambda_{\mathrm{Adam}}$. As before, we write:
\begin{equation}
	\Sigma=UXU^\top,
	\qquad X\succeq0.
\end{equation}
We can then obtain the central flow for Adam:
\begin{equation}
	\begin{aligned}
		\ddt{\bar{w}}
		&=
		-\eta P^{-1}
		\left[
		\beta_1\bar{m}
		+(1-\beta_1)\Bigl(
		\nabla L(\bar{w})
		+\tfrac12\nabla_{\bar{w}}
		\langle H(\bar{w}),\Sigma\rangle
		\Bigr)
		\right],
		\\
		\ddt{\bar{m}}
		&=
		(1-\beta_1)
		\left[
		\nabla L(\bar{w})
		+\frac12\nabla_{\bar{w}}
		\langle H(\bar{w}),\Sigma\rangle
		-\bar{m}
		\right],
		\\
		\ddt{\bar{\nu}}
		&=
		(1-\beta_2)
		\left[
		(\nabla L(\bar{w}))^{\odot2}
		+\operatorname{diag}
		\left(
		H(\bar{w})\Sigma H(\bar{w})
		\right)
		-\bar{\nu}
		\right].
	\end{aligned}
	\label{eq:adam-central}
\end{equation}
As with RMSProp, the second-moment equation uses the approximation
\begin{equation}
	\bar{g}^{\odot2}+\gjump^{\odot2}
	\approx
	(\nabla L(\bar{w}))^{\odot2}
	+\operatorname{diag}(H\Sigma H),
\end{equation}
which follows from the local quadratic approximation of the loss about $\wbar$ and by replacing $\delta\otimes\delta$ with $\Sigma$.

We determine the covariance by imposing persistence of feasibility
and complementarity. The time derivative of the slack is:
\begin{equation}
	\dot{\Lambda}_{\mathrm{Adam}}
	= \tfrac{2(1+\beta_1)}{\eta(1-\beta_1)}\nabla_{\bar{\nu}}P(\nubar)[\dot{\nubar}]
	- \nabla_{\bar{w}}H(\wbar)[\dot{\wbar}].
\end{equation}
At fixed $(\bar{w},\bar{\nu})$, both the $\dot{\bar{w}}$ and $\dot{\bar{\nu}}$ terms are affine in $\Sigma$.
Persistence then requires:
\begin{equation}
	U^\top\dot{\Lambda}_{\mathrm{Adam}}U\succeq0,
	\qquad
	\left\langle
	U^\top\dot{\Lambda}_{\mathrm{Adam}}U,X
	\right\rangle=0.
\end{equation}
These persistence conditions define a semidefinite complementarity problem for $X$. Altogether, we have that the following conditions define the central flow for Adam:
\begin{equation}
	\boxed{
		\begin{aligned}
			\ddt{\bar{z}} &= \Kmat_{\bar{\nu}}\, \partial_{\bar{z}} \widetilde{\Upsilon}^{\mathrm{red}}_{\eta,\beta_1,\bar{\nu}}(\bar{z}, \Sigma), \qquad \Sigma = U X U^\top, \\[4pt]
			\ddt{\bar{\nu}} &= (1-\beta_2)\Bigl[(\nabla L(\bar{w}))^{\odot2} + \operatorname{diag}\bigl[H(\bar{w})\Sigma H(\bar{w})\bigr] - \bar{\nu}\Bigr], \\[4pt]
			X &\succeq 0, \qquad U^\top \dot{\Lambda}_{\mathrm{Adam}} U \succeq 0, \qquad \langle U^\top \dot{\Lambda}_{\mathrm{Adam}} U, X \rangle = 0.
		\end{aligned}
	}
\end{equation}

One can also derive a central flow for AdamW. Details are provided in \cref{app:adamw}.

%% file: sections/08-timescales.tex
\section{Rod Flow and Central Flow}
\label{sec:timescales}

We show that the key difference between rod flow and central flow is the time scale at which they coarse-grain. Rod flow coarse-grains over only the fast time scale, whereas central flow coarse-grains over both the fast and the intermediate time scales.

\subsection{Fast, Intermediate, and Slow Time Scales}
\label{sec:three-timescales}

\citet{berthier2026multiple} derived central flow by the method of multiple scales, identifying three time scales. There's the \textit{fast} time scale, given by the iteration counter itself. On this time scale, the sharp components of the iterates alternate in sign. There's the \textit{intermediate} time scale of self-stabilization. When the oscillation amplitude departs from its steady-state value, this time scale governs how long it takes for the amplitude to re-equilibrate. Finally, there's the \textit{slow} time scale. It's the time scale that governs the drift of the center of the oscillating iterates.

Under this framework, rod flow is an effective theory for gradient descent that coarse-grains over the fast time scale only. Recall that the half-displacement of the iterates decomposes as
\begin{equation*}
	\dhat_t \approx \tfrac{\eta}{2}\bigl(\gbar + (-1)^t\gjump\bigr). \tag{\ref{eq:dhat-decomp}}
\end{equation*}
The alternating component flips sign at every step of the iteration counter. By tracking the unsigned envelope of the oscillation, rod flow coarse-grains over this sign flip.

What are the dynamics of the envelope? It helps to consider the rod flow ODEs when expanded to leading order in $\delta$:
\begin{equation}
	\ddt{\delta} = \bigl[\eta H(\wbar) - 2I_d\bigr]\delta,
	\qquad
	\ddt{\wbar} = -\eta\Bigl[\nabla L(\wbar) + \tfrac{1}{2}\nabla_{\wbar}\inner{H(\wbar)}{\delta\ot\delta}\Bigr].
\end{equation}
The ODE for $\delta$ says that oscillations grow along components sharper than the $2/\eta$ threshold and decay along components that are less sharp than the threshold. The ODE for $\bar{w}$ is reminiscent of the one for central flow: it is a gradient flow on the loss, plus a sharpness-reduction force whose strength is proportional to the squared magnitude of the oscillations.

Together, the two equations form a feedback loop. Excess sharpness causes the amplitude to grow along sharp directions. The larger oscillations strengthen the sharpness-reduction force, which pushes the sharpness below the threshold. The oscillations then decay, weakening the sharpness-reduction force. The sharpness then grows until it is back above the threshold. The result is self-stabilization: the iterates hover at the $2/\eta$ sharpness threshold without diverging.

We can express the self-stabilization loop as a Lotka--Volterra-type system. Consider the case of a single sharp direction. Let $\delta = s\,\hat{u}$ where $\hat{u}$ is the unit vector in the sharp direction. And let $x = \eta\lamone - 2$ denote the excess sharpness. Then we can obtain the following ODEs for the amplitude $s$ and the excess sharpness $x$:
\begin{equation}
	\ddt{s} = x\,s,
	\qquad
	\ddt{x} = \eta\Bigl(r - \tfrac{\eta}{2}\,c\,s^2\Bigr),
	\label{eq:lv}
\end{equation}
where $r = -\eta\inner{\nabla\lamone}{\nabla L}$ and $c = \norm{\nabla\lamone}^2$. Here $r$ is the \textit{progressive sharpening constant}: it measures how much descending the loss tends to increase the sharpness. We have that $r > 0$ corresponds to progressive sharpening. Likewise, $c$ is the \textit{sharpness reduction coefficient}: it measures the strength of the sharpness-reduction force per unit of squared amplitude.

Where does central flow fit in? The solutions to the Lotka--Volterra-type system above trace out closed orbits. Averaging the excess sharpness and the squared amplitude over one orbit gives:
\begin{equation}
	\langle x\rangle = 0,
	\qquad
	\langle s^2\rangle = \frac{2r}{\eta c} .
	\label{eq:lv-avg}
\end{equation}
Over one orbit, the average excess sharpness is zero. And the average squared amplitude is proportional to the progressive sharpening constant and inversely proportional to the sharpness reduction coefficient. We have that oscillations are larger when progressive sharpening is strong or when the third derivative of the loss is small.

Central flow corresponds to the approximation in which we replace the instantaneous value of the amplitude-squared with its average over one orbit. Instead of coarse-graining over just the fast time scale of alternation as with rod flow, central flow also coarse-grains over the intermediate time scale of self-stabilization.

One subtlety is that the orbits are closed only when $r$ and $c$ evolve on the slow time scale and can therefore be treated as constant. For $r$, this assumption holds only while the oscillation is small. Once the oscillation becomes large, the orbits are no longer closed: the period lengthens and the orbits decay. Importantly, however, the orbit averages remain valid even in this regime---which is what justifies central flow's approximation.

\subsection{Central Flow as Time-Averaged Rod Flow}
\label{sec:ranks}

That the two flows coarse-grain over different time scales also helps explain another key difference between them---the rank of the object each uses to encode the oscillations.

\begin{figure}[t]
	\centering
	\includegraphics{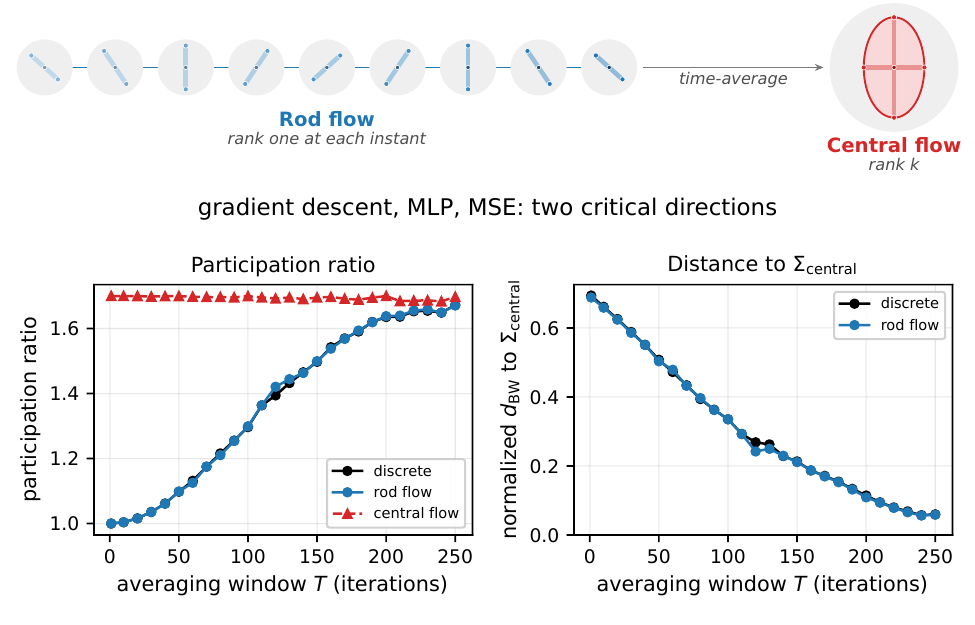}
	\caption{\textbf{Central Flow as Time-Averaged Rod Flow.} \textbf{Top:} The extent matrix of rod flow has rank one at any given instant. In the presence of multiple sharp eigenvalues, rod flow self-stabilizes by turning in the critical subspace. The covariance matrix of central flow can be interpreted as the time average of the extent matrix. It has rank $k$, where $k$ is the number of critical eigenvalues. \textbf{Bottom:} Gradient descent on an MLP ($\eta = 0.05$). There are two sharp eigenvalues. The extent matrix of rod flow and the discrete iterates are averaged over windows of $T$ iterations. \textbf{Left:} As the width of the averaging window increases, their participation ratio rises from one to that of $\Sigma_{\mathrm{central}}$. \textbf{Right:} Their normalized Bures--Wasserstein distance to $\Sigma_{\mathrm{central}}$ falls toward zero.}
	\label{fig:time-average}
\end{figure}

Rod flow is rank one: it tracks a single vector $\delta$. This modeling assumption is natural, as a single step of gradient descent displaces the iterate along a single vector. But it raises a question: how does rod flow continue to track the iterates when there are multiple sharp directions?

Central flow is rank $k$, where $k$ is the number of critical eigenvalues. This makes sense too. Each oscillation direction self-stabilizes a different sharp direction, and all $k$ directions are needed for the SDCP of central flow to have a solution. But it raises a question of its own: what observable of the discrete iterates does a rank-$k$ covariance correspond to?

Let $\Sigma_{\mathrm{rod}} = \delta \otimes \delta$ denote the extent matrix of rod flow, and let $\Sigma_{\mathrm{central}}$ denote the covariance matrix of central flow. We claim that the extent matrix of rod flow, averaged over the intermediate time scale, equals the covariance matrix of central flow:
\begin{equation}
	\Sigma_{\mathrm{central}}(t) = \frac{1}{T}\int_{t-T/2}^{t+T/2} \Sigma_{\mathrm{rod}}(t')\,\dd t',
\end{equation}
where the averaging window $T$ spans one full swing of the rod through the critical subspace. The window is long on the intermediate time scale but short on the slow one, so $\wbar$ barely moves across it. Empirically, we find that $T$ corresponds to roughly two orbits of the self-stabilization loop.

Rod flow maintains self-stabilization by \textit{turning}. Although the extent is rank one at any given instant, its axis swings back and forth within the critical subspace. Central flow, whose covariance matrix encodes the time-averaged oscillations, carries all of the sharp directions at once.
That the covariance matrix has rank $k$ is necessary. As \citet{berthier2026multiple} notes, when the self-stabilization loop is generalized to more than one sharp direction, it generically has no rank-one fixed point.

We tested this hypothesis experimentally in a setting with two sharp eigenvalues (\cref{fig:time-average}). We averaged the extent matrix of rod flow over windows of increasing length and compared each average with the covariance matrix of central flow. Rather than comparing the ranks of these matrices, we compared their participation ratios (PR) as the participation ratio is less sensitive to small eigenvalues.
\begin{equation}
	\mathrm{PR}(\Sigma) = \frac{(\operatorname{tr}\Sigma)^2}{\operatorname{tr}(\Sigma^2)} = \frac{\bigl(\sum_i \sigma_i\bigr)^2}{\sum_i \sigma_i^2},
\end{equation}
where $\sigma_i$ are the eigenvalues of $\Sigma$.

As expected, the extent matrix of rod flow averaged over a window of one step has rank one. As the window grows, the participation ratio rises. It completes $90\%$ of its rise at a window length of about $185$ iterations and levels off by $250$, reaching $1.67$ compared with $1.70$ for central flow. The participation ratio stays below two because the two critical directions are unequally excited.

While the participation ratio tells us that the oscillations are spread out across directions to a similar degree, it doesn't tell us whether those directions coincide. To probe this, we also computed the Bures--Wasserstein distance between the time-averaged extent matrix and the covariance matrix of central flow. We normalized the distance by $\sqrt{\operatorname{tr}\Sigma_{\mathrm{central}}}$, so that it does not depend on the overall size of the oscillations:
\begin{equation}
		\hat{d}_{\mathrm{BW}}(T) = \frac{d_{\mathrm{BW}}\bigl(\langle\Sigma_{\mathrm{rod}}\rangle_T,\, \Sigma_{\mathrm{central}}\bigr)}{\sqrt{\operatorname{tr}\Sigma_{\mathrm{central}}}}
\end{equation}
where $\langle \Sigma_{\mathrm{rod}} \rangle_T$ denotes the time-averaged extent matrix over a window of length $T$ iterations.

Roughly, a normalized distance near one indicates very dissimilar matrices, whereas a distance near zero indicates nearly identical ones. When comparing $\langle \Sigma_{\mathrm{rod}} \rangle_T$ and $\Sigma_{\mathrm{central}}$ across averaging windows of increasing width, the normalized distance falls from $0.69$ for a single step to $0.06$ at $T = 250$. This shows that the two matrices agree not only on the magnitude of the oscillations but also in their direction.

\subsection{Turning Rate of the Rod}
\label{sec:turning}

\begin{figure}[t]
	\centering
	\includegraphics[width=\textwidth]{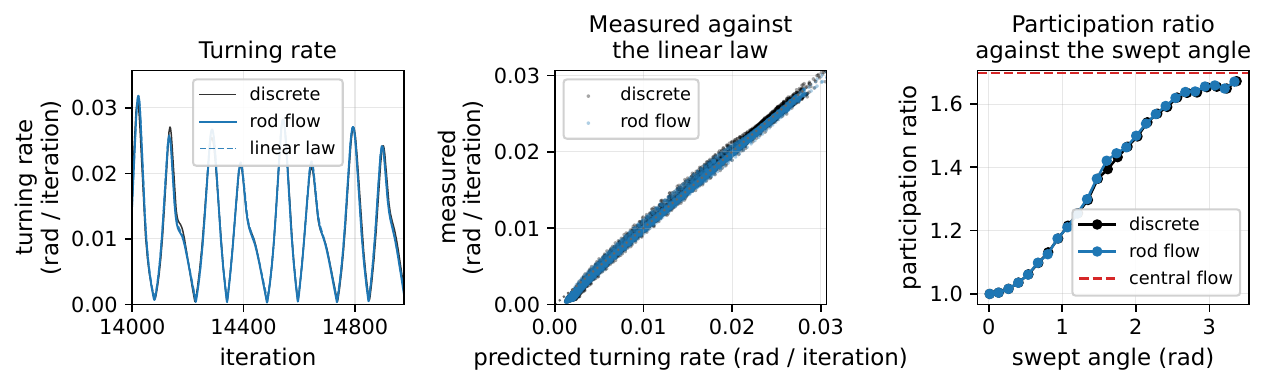}
	\caption{\textbf{Turning Rate of the Rod.} Gradient descent on an MLP ($\eta = 0.05$). \textbf{Left:} Time series of the turning rate for rod flow and for the discrete oscillation, compared with the prediction of the linear turning law. \textbf{Middle:} Scatter plot of measured versus predicted turning rates for the discrete oscillation and for rod flow, showing strong agreement. \textbf{Right:} Participation ratio of the time-averaged extent matrix as a function of the swept angle.}
	\label{fig:turning-gd}
\end{figure}

What determines how fast the rod turns? Write $\delta = s\,\hat{v}$ with $\norm{\hat{v}} = 1$. The rod flow ODE for $\delta$ then splits into separate ODEs for the amplitude $s$ and the axis $\hat{v}$:
\begin{equation}
	\ddt{s} = \eta\inner{\gjump}{\hat{v}} - 2s,
	\qquad
	\ddt{\hat{v}} = \frac{\eta}{s}\,\gjump_\perp,
	\label{eq:axis}
\end{equation}
where $\gjump_\perp := \gjump - \inner{\gjump}{\hat{v}}\,\hat{v}$ is the component of the gradient half-difference orthogonal to the current axis of the rod. All of the turning comes from this transverse component. Taylor-expanding the gradient half-difference and retaining only the Hessian term of $\gjump$, we obtain:
\begin{equation}
	\ddt{\hat{v}} \approx \eta\bigl(H\hat{v} - \inner{H\hat{v}}{\hat{v}}\,\hat{v}\bigr) = \eta\,(H\hat{v})_\perp .
	\label{eq:turning-law}
\end{equation}
As the right-hand side is orthogonal to $\hat{v}$, the flow preserves $\norm{\hat{v}} = 1$. By the Pythagorean theorem, $\norm{(H\hat{v})_\perp}^2 = \norm{H\hat{v}}^2 - \inner{H\hat{v}}{\hat{v}}^2$. And since $H$ is symmetric, $\norm{H\hat{v}}^2 = \inner{H^2\hat{v}}{\hat{v}}$. Hence
\begin{equation}
	\norm{\dot{\hat{v}}} \approx \eta\sqrt{\inner{H^2\hat{v}}{\hat{v}} - \inner{H\hat{v}}{\hat{v}}^2} .
\end{equation}
To help interpret this expression, write
\begin{equation}
	\hat{v} = \sum_{i=1}^d c_i\,\hat{u}_i
\end{equation}
where the $\hat{u}_i$ are orthonormal eigenvectors of the Hessian with corresponding eigenvalues $\lambda_i$. Note that $\sum_i c_i^2 = 1$. We then have that:
\begin{equation}
	\norm{\dot{\hat{v}}} \approx \eta\sqrt{\sum_{i} c_i^2\,(\lambda_i - \bar{\lambda})^2} ,
\end{equation}
where $\bar{\lambda} = \sum_i c_i^2\,\lambda_i$ is the Rayleigh quotient of $H$ at $\hat{v}$. The rod therefore turns faster when $\hat{v}$ has substantial components along several eigenvectors with different eigenvalues.

To make this concrete, consider the case where we have two sharp eigenvalues $\lamone > \lambda_2$ with respective eigenvectors $\hat{u}_1$ and $\hat{u}_2$. Write $\hat{v} = \cos\theta\,\hat{u}_1 + \sin\theta\,\hat{u}_2$. Then
\begin{equation}
\dot{\hat{v}} = \dot{\theta}\,\hat{e}_\theta
\end{equation}
where $\hat{e}_\theta = -\sin\theta\,\hat{u}_1 + \cos\theta\,\hat{u}_2$ is the unit vector perpendicular to $\hat{v}$ in the two-dimensional critical plane. Using $H\hat{v} = \lamone\cos\theta\,\hat{u}_1 + \lambda_2\sin\theta\,\hat{u}_2$ and $\inner{H\hat{v}}{\hat{v}} = \lamone\cos^2\theta + \lambda_2\sin^2\theta$, a direct computation gives:
\begin{equation*}
	H\hat{v} - \inner{H\hat{v}}{\hat{v}}\,\hat{v} = (\lambda_2 - \lamone)\sin\theta\cos\theta\,\hat{e}_\theta .
\end{equation*}
Matching the coefficients of $\hat{e}_\theta$ yields:
\begin{equation}
	\ddt{\theta} = -\frac{\eta}{2}\,(\lamone - \lambda_2)\,\sin 2\theta .
	\label{eq:two-mode}
\end{equation}
The turning rate is thus proportional to the gap between the two critical eigenvalues. In particular, since $\lamone > \lambda_2$, the rod turns toward $\theta = 0$---the direction of the sharper eigenvector $\hat{u}_1$.

We tested the turning-rate law by comparing the measured turning rate with its predicted value and found strong agreement between the two (\cref{fig:turning-gd}). Furthermore, in the case of two sharp eigenvalues, we found that the participation ratio of $\langle \Sigma_{\mathrm{rod}} \rangle_T$ converged to that of $\Sigma_{\mathrm{central}}$ after roughly $\pi$ radians of turning---in line with expectations.

\subsection{Turning Rate for Momentum Methods}
\label{sec:timescales-momentum}

\begin{figure}[t]
	\centering
	\includegraphics[width=\textwidth]{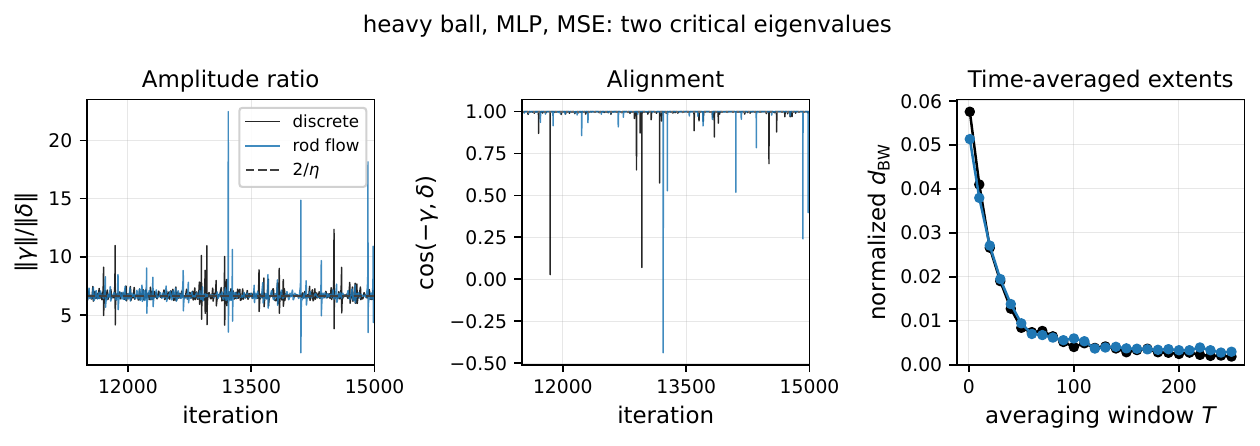}
	\caption{\textbf{Position and Momentum Extents of Heavy Ball.} Heavy ball on an MLP ($\eta = 0.3$, $\beta = 0.9$). There are two critical eigenvalues. \textbf{Left:} The amplitude ratio $\norm{\gamma}/\norm{\delta}$ over time. For period-two orbits, the amplitude ratio is $2/\eta$ (dashed line). The ratio hovers around this value but shows non-trivial fluctuations. \textbf{Middle:} The alignment $\cos(-\gamma, \delta)$ between the position and momentum extents. \textbf{Right:} The normalized Bures--Wasserstein distance between the time averages $\langle \tfrac{\eta^2}{4}\gamma\gamma^\top \rangle_T$ and $\langle \delta\delta^\top \rangle_T$. As the averaging window $T$ grows, the time-averaged position and momentum extent matrices increasingly agree with the relation they satisfy on a period-two cycle.}
	\label{fig:two-cycle-hb}
\end{figure}

We will now consider the turning rate for heavy ball. Let $\delta = s \hat{v}$. As before, the rate of change of the direction $\hat{v}$ is given by the orthogonal component of the update for $\delta$, divided by the amplitude $s$:
\begin{equation}
	\ddt{\hat{v}} = \frac{\eta}{s}\bigl[(1-\beta)\gjump + \beta\gamma\bigr]_\perp .
\end{equation}
In contrast with gradient descent, there is now a term proportional to the momentum extent $\gamma$---which must be handled with care. For a period-two orbit, the position and momentum half-differences are in fixed proportion. However, this is \textit{not} generically true of the position and momentum extents over the course of the dynamics (\cref{fig:two-cycle-hb}).

To see why, consider heavy ball on the one-dimensional quadratic $L(w) = \tfrac12\lambda w^2$. The update is linear in phase space:
\begin{equation*}
	\begin{pmatrix} w_{t+1} \\ m_{t+1} \end{pmatrix}
	= A \begin{pmatrix} w_t \\ m_t \end{pmatrix},
	\qquad
	A = \begin{pmatrix} 1 - (1-\beta)\eta\lambda & -\eta\beta \\ (1-\beta)\lambda & \beta \end{pmatrix}.
\end{equation*}
It helps to examine the eigenvalues and eigenvectors of $A$. Since $\operatorname{tr} A = 1 + \beta - (1-\beta)\eta\lambda$ and $\det A = \beta$, the characteristic equation is
\begin{equation}
	\alpha^2 - \tau\alpha + \beta = 0,
	\qquad
	\tau = 1 + \beta - (1-\beta)\eta\lambda .
\end{equation}
The eigenvalues and corresponding eigenvectors are
\begin{equation}
	\alpha_\pm = \frac{\tau \pm \sqrt{\tau^2 - 4\beta}}{2},
	\qquad
	v_\pm = \begin{pmatrix} \alpha_\pm - \beta \\ (1-\beta)\lambda \end{pmatrix}.
\end{equation}
One can see that the eigenvalues are complex when $\tau^2 < 4\beta$. This is satisfied when:
\begin{equation*}
	\frac{1-\sqrt{\beta}}{1+\sqrt{\beta}} < \eta\lambda < \frac{1+\sqrt{\beta}}{1-\sqrt{\beta}} .
\end{equation*}
At $\beta = 0.9$, the window is $0.026 < \eta\lambda < 37.97$---against a threshold value of $38$. Because sharp eigenvalues hover around the sharpness threshold rather than staying exactly at it, for large momentum coefficients they spend a significant fraction of the time in the complex window. This causes the sharp components to rotate in the $(w, m)$ plane. As a result, their amplitude ratio fluctuates instead of staying at exactly $2/\eta$.

We must therefore treat $\gamma$ as a separate degree of freedom. Define $\tilde{\gamma} = \gamma/s$, the momentum extent divided by the position amplitude. Linearizing $\gjump \approx sH\hat{v}$ as before, the $\gamma$ ODE gives
\begin{equation*}
	\ddt{\tilde{\gamma}} = \frac{\dot{\gamma}}{s} - \frac{\dot{s}}{s}\,\tilde{\gamma}
	\approx -(1+\beta)\tilde{\gamma} - (1-\beta)H\hat{v} - \frac{\dot{s}}{s}\,\tilde{\gamma} .
\end{equation*}
The amplitude enters through the ratio $\dot{s}/s$, which we can express in terms of $\hat{v}$ and $\tilde{\gamma}$ using the ODE for $s$:
\begin{equation}
	\frac{\dot{s}}{s} \approx -2 + \eta\, \hat{v}^\top \bigl[(1-\beta)H\hat{v} + \beta\tilde{\gamma}\bigr].
\end{equation}
Substituting this expression, we obtain a closed system of ODEs for the pair $(\hat{v}, \tilde{\gamma})$:
\begin{equation}
	\begin{aligned}
		\ddt{\hat{v}} &\approx \eta\bigl[(1-\beta)H\hat{v} + \beta\tilde{\gamma}\bigr]_\perp, \\[4pt]
		\ddt{\tilde{\gamma}} &\approx (1-\beta)\bigl(\tilde{\gamma} - H\hat{v}\bigr) - \eta\, \hat{v}^\top \bigl[(1-\beta)H\hat{v} + \beta\tilde{\gamma}\bigr]\,\tilde{\gamma} .
	\end{aligned}
	\label{eq:hb-turning-law}
\end{equation}
We tested this turning law for heavy ball and again found good agreement with the measured turning rate. More details are provided in \cref{app:rod-turning,app:two-cycle}.

\subsection{Summary}

There are four main modeling differences between central flow and rod flow.
\begin{enumerate}
	\item Central flow treats the covariance matrix $\Sigma$ as a Lagrange multiplier, whereas in rod flow the extent $\delta$ evolves according to its own ODE.
	\item Central flow replaces the loss with its local quadratic approximation so that $\dot{\bar{w}}$ is affine in $\Sigma$, whereas rod flow evaluates the exact loss at the endpoints of the rod.
	\item The covariance matrix of central flow is rank $k$, where $k$ is the number of critical eigenvalues, whereas the extent matrix of rod flow is rank one.
	\item For momentum methods, central flow represents only the position oscillations, whereas rod flow carries the position and momentum extents as distinct degrees of freedom.
\end{enumerate}
This time-scale analysis reveals that these modeling choices are not independent: all four differences follow from the fact that central flow coarse-grains over the intermediate time scale.

Averaging over the self-stabilization loop removes the amplitude as a dynamical variable. What survives is its orbit average, which is the Lagrange multiplier that holds the sharpness at $2/\eta$. Because the covariance matrix of central flow is a Lagrange multiplier rather than a state variable, it must be determined by solving an optimization problem. To make this optimization problem tractable, it helps to replace the loss with its local quadratic approximation---which makes the dynamics of $\bar{w}$ affine in $\Sigma$ and yields an SDCP. Once $\Sigma$ is fixed by the SDCP, persistence dictates that its rank must equal the number of critical directions. And while the momentum and position oscillations may be independent at any given instant, their time averages converge to the amplitude ratio of a period-two orbit---rendering them interdependent over intermediate time scales.

%% file: sections/09-experiments.tex
\section{Experiments}
\label{sec:experiments}

We evaluate the rod flows and central flows on three representative neural network architectures: an MLP~\citep{RumHinWil86Backprop}, a CNN~\citep{LeC+98CNN}, and a ViT~\citep{Dos+21ViT}. We compare them with both the discrete iterates and the stable flow, the na\"{\i}ve continuous-time limit of the discrete optimizer.

\subsection{Setup}
\label{sec:exp-setup}

\begin{figure}[t]
	\centering
	\includegraphics[width=\textwidth]{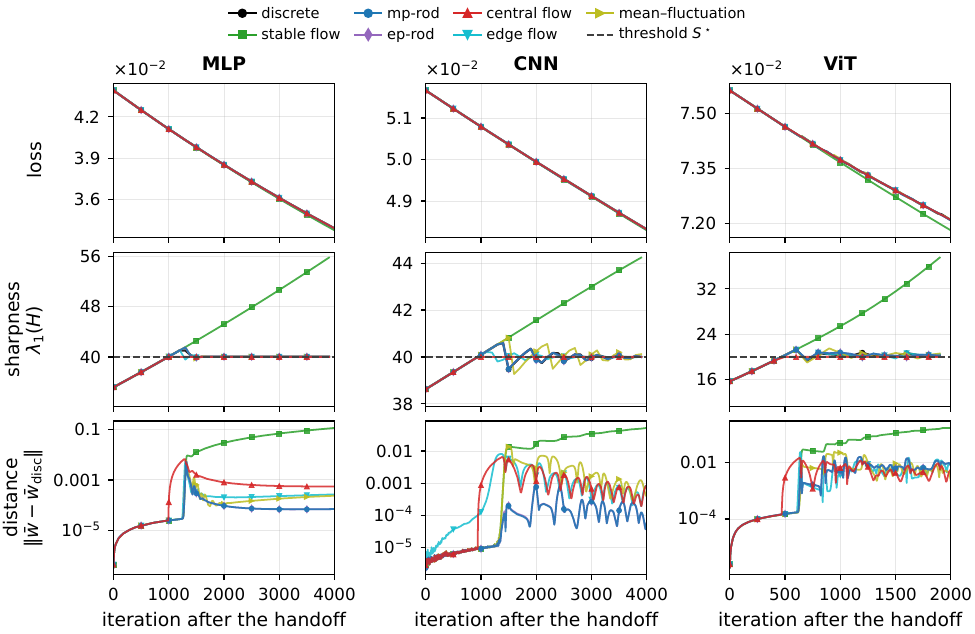}
	\caption{\textbf{Experimental Results for Gradient Descent.} \textbf{Left:} MLP ($\eta = 0.05$). \textbf{Center:} CNN ($\eta = 0.05$). \textbf{Right:} ViT ($\eta = 0.1$). The flows are handed off from the discrete iterates $1\,000$ iterations before the crossing ($500$ for the ViT). All flows except the stable flow self-stabilize at the sharpness threshold, and they track the discrete iterates orders of magnitude more closely than the stable flow does.}
	\label{fig:exp-gd}
\end{figure}

All three models were trained full-batch on the same subset of $5\,000$ examples from CIFAR-10~\citep{Kri09CIFAR}. The loss function was mean squared error (MSE) on one-hot targets. The MLP had two hidden layers of width $50$ ($156\,710$ parameters). The CNN had two $3\times3$ convolutional layers with $32$ channels ($30\,634$ parameters). The ViT was a SimpleViT~\citep{beyer2022better} with width $64$, depth $3$, $8$ attention heads of width $8$, and MLP blocks of width $256$ ($153\,322$ parameters). The step sizes were chosen using probe runs, so that the edge of stability was reached after a few thousand iterations.

We ran experiments on the following optimizers: gradient descent, RMSProp, heavy-ball momentum, Nesterov momentum, Adam, and AdamW. In this section, we will focus on the results for gradient descent and Adam. The experimental results for all the optimizers can be found in \cref{app:more-experiments}.

For each optimizer, we ran stable flow, midpoint and endpoint rod flow (mp-rod and ep-rod in the figures), and central flow. Both rod flows were run with noise floors. For gradient descent, we also ran two other continuous-time models: edge flow~\citep{marion2026edge} and the mean--fluctuation model~\citep{chodron2026free}. We implemented the mean--fluctuation model from the first version of \citet{chodron2026free}, whose equations are unchanged in the second.

Each flow was initialized from the discrete iterates $1\,000$ iterations before the transition to the edge of stability and run until $3\,000$ iterations after it ($500$ before and $1\,500$ after for the ViT). The flows were integrated with four forward Euler substeps per iteration.

\subsection{Results}
\label{sec:exp-results}
\begin{figure}[t]
	\centering
	\includegraphics[width=0.8\textwidth]{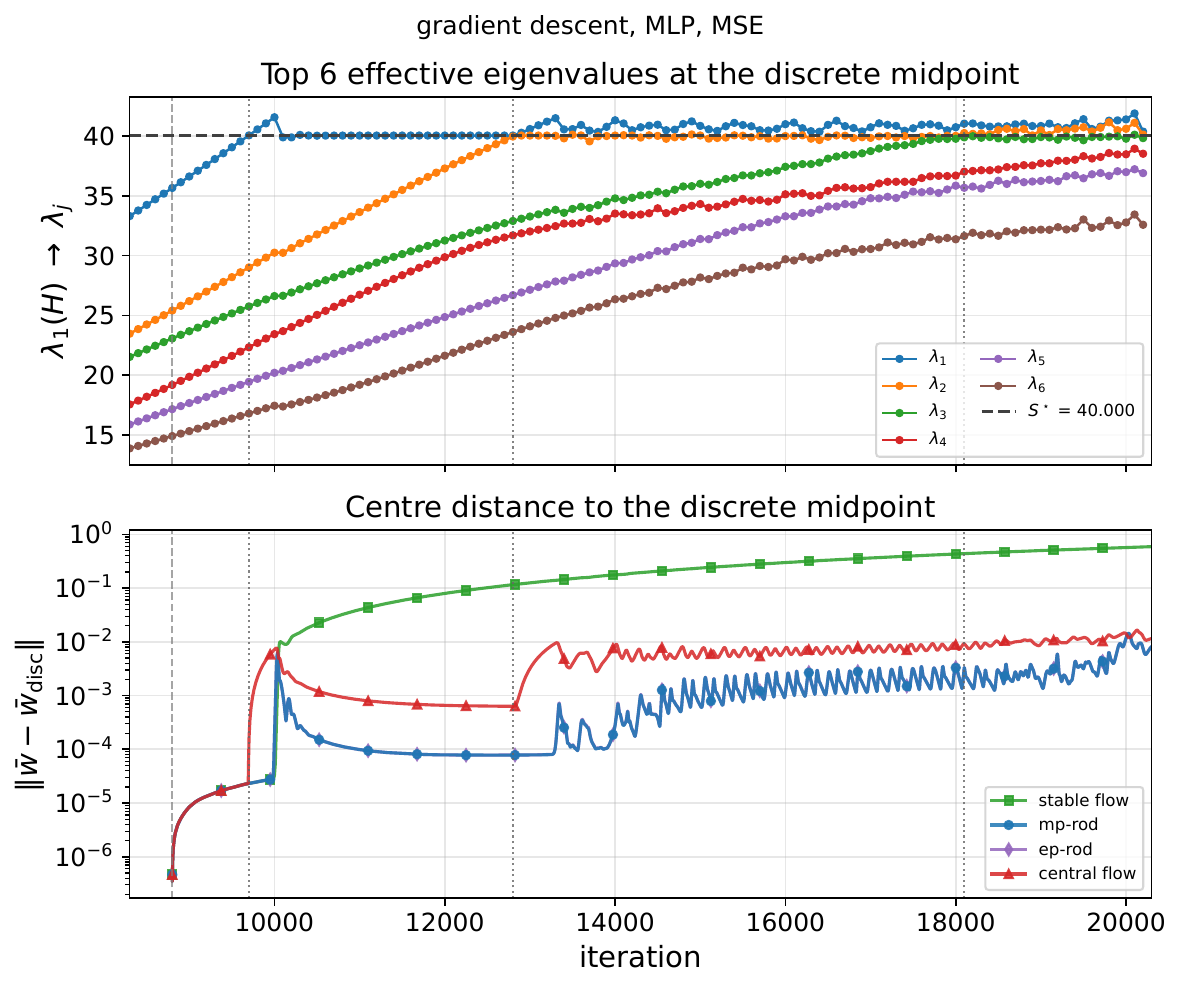}
	\caption{\textbf{Tracking Through Three Critical Eigenvalues.} Gradient descent on an MLP ($\eta = 0.05$). \textbf{Top:} The top six eigenvalues of the Hessian at the discrete midpoint. $\lamone$, $\lambda_2$, and $\lambda_3$ reach the threshold one after another. \textbf{Bottom:} The distance from each flow's center to the discrete midpoint. Rod flow and central flow stay close to the iterates through all three transitions, while the stable flow drifts away.}
	\label{fig:exp-longrun}
\end{figure}

\paragraph{Gradient descent.} \Cref{fig:exp-gd} shows the results for gradient descent on the three architectures. Every flow except the stable flow self-stabilizes at the $2/\eta$ sharpness threshold. The sharpness of central flow sits exactly at the threshold, while the sharpnesses of the other flows fluctuate about the threshold. Importantly, rod flow now successfully tracks the discrete iterates \textit{through} the transition to the edge of stability---fixing a shortcoming of the previous formulation.

Every flow tracks the center of the iterates more than an order of magnitude more closely than the stable flow does. Rod flow tracks best on the MLP and the CNN, staying three to four times closer to the iterates than the next-best flow. On the ViT, the five flows are within a factor of two of one another, with central flow tracking slightly better than the other flows. For all three architectures, the midpoint and endpoint rod flows coincide.

\paragraph{Several critical eigenvalues.} It is of interest to test the performance of rod flow and central flow in the presence of multiple sharp eigenvalues. To do so, we conducted an extended run on the MLP---until three eigenvalues had reached the threshold (\cref{fig:exp-longrun}). Each new critical eigenvalue raises the tracking error: about tenfold when the second largest eigenvalue becomes critical, and about twofold when the third largest one does. Furthermore, once a second eigenvalue becomes critical, the distance between the flows and the discrete iterates starts to wobble. Still, even in the presence of multiple sharp eigenvalues, both rod flow and central flow successfully track the discrete iterates, staying more than an order of magnitude closer to them than the stable flow does.

\begin{figure}[t]
	\centering
	\includegraphics[width=\textwidth]{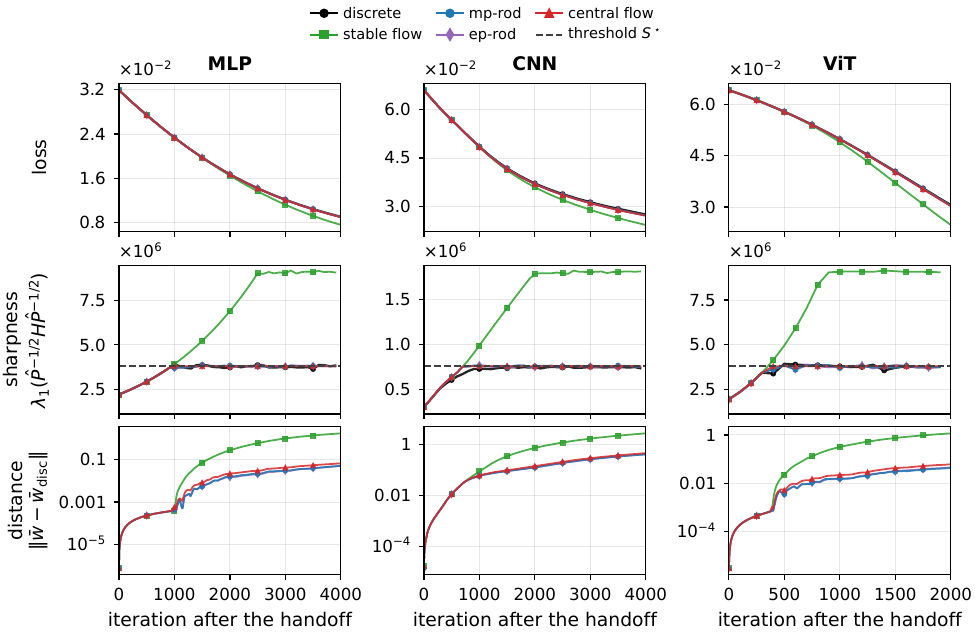}
	\caption{\textbf{Experimental Results for Adam} (with bias correction). \textbf{Left:} MLP ($\eta = 10^{-5}$). \textbf{Center:} CNN ($\eta = 5\times10^{-5}$). \textbf{Right:} ViT ($\eta = 10^{-5}$). All three with $\beta_1 = 0.9$, $\beta_2 = 0.999$. All flows except the stable flow self-stabilize the preconditioned sharpness at its threshold $38/\eta$, and they track the discrete iterates more closely than the stable flow does.}
	\label{fig:exp-adam}
\end{figure}

\paragraph{Adam.}

The previous formulation of the rod flow for Adam~\citep{regis2026adamrod} had several shortcomings. Like the original rod flow for gradient descent, it could not track the transition to the edge of stability and had to be initialized in the steady-state edge of stability regime. But even there, it failed to track the iterates for large values of the momentum coefficient---the largest value demonstrated was $\beta_1 = 0.8$ for the MLP.

In our experimental runs, all three architectures use $\beta_1 = 0.9$, $\beta_2 = 0.999$, and bias correction---hyperparameter settings that are actually used in practice. And as was the case with gradient descent, the flows are initialized before the transition.

\Cref{fig:exp-adam} shows the results for Adam on the three architectures. Let $\hat{P}_t$ denote the bias-corrected preconditioner:
\begin{equation}
	\hat{P}_t = P\!\left(\frac{\nu_t}{1 - \beta_2^{t}}\right).
	\label{eq:adam-precond}
\end{equation}
Every flow except the stable flow self-stabilizes the preconditioned sharpness $\lamone(\hat P^{-1/2}H\hat P^{-1/2})$ at the $38/\eta$ threshold---that of heavy ball at $\beta_1 = 0.9$. The preconditioned sharpness of the central flow sits at the threshold, while those of the rod flow and the discrete iterates fluctuate within a few percent of it.

Every other flow also tracks the center of the iterates more closely than the stable flow does: twenty to thirty times more closely on the MLP and the ViT, and about six times more closely on the CNN. However, on the MLP and the CNN, this is a far smaller margin than was the case for gradient descent. Also, the rod flow and central flow for Adam perform much more alike than they did for gradient descent. Rod flow still tracks better, but central flow is only ten to forty percent farther from the discrete iterates than rod flow.

Why does the edge-coupling-derived formulation of the Adam rod flow perform well while the original formulation did not? We conjecture that the discrepancy comes from how the two formulations handle the slow drift. As $\beta_1$ grows, the drift makes up a larger and larger share of the momentum. Because the extent of the original formulation tracks the second moment of the momentum---drift included---it can no longer self-stabilize once the drift becomes comparable in size to the oscillations.

\citet{regis2026adamrod} had suggested that period-doubling bifurcations could be a possible cause for the failure of self-stabilization of rod flow at large values of $\beta_1$. However, we found that the oscillations of Adam are in fact mostly period-two---though with some bursty behavior. More details on the validity of the near-period-two assumption can be found in \cref{app:two-periodic}.

%% file: sections/10-conclusion.tex
\section{Conclusion}
\label{sec:conclusion}

We have established connections between central flow, rod flow, and the edge coupling. Using the edge coupling, we derived an alternative formulation of rod flow---which now accurately tracks the iterates throughout all phases of training. In addition, we used the edge coupling to re-derive the central flows for gradient descent and RMSProp. This allowed us to generalize our derivation to obtain central flows for heavy ball and Adam. We have also shown that central flow and rod flow represent different coarse-grained descriptions of the dynamics---which explains how they can differ while both remaining accurate.
\paragraph{Limitations.}
Rod flow and central flow currently apply only to the full-batch setting. Both are also more expensive than the optimizers they model, so they are tools for understanding optimization dynamics rather than substitutes for the optimizers. Finally, while we have provided extensive experimental verification, we have not formally proven that rod flow or central flow precisely tracks the discrete iterates.

\paragraph{Future work.}
Natural next steps include extending rod flow and central flow to the minibatch setting. One would also like to prove formal tracking bounds between the flows and the discrete iterates. Finally, extending this framework to optimizers outside the Adam family remains future work.

\section*{Acknowledgements}

We would like to thank Rapha{\"e}l Berthier for discussions. We used various LLMs---Opus 5.5, Fable 5.1, and GPT-6 Astra---in the brainstorming, drafting, and editing of this paper. We thank the authors of the Central Flow paper for making their code publicly available. Our implementation builds on code from \url{https://github.com/alex-damian/EOS} and \url{https://github.com/centralflows/centralflows}.

%% file: appendix/A-notation.tex
\section{Notation}
\label{app:notation}

{\small
\begin{longtable}{cl}
\caption{Notation used throughout the main body.}
\label{tab:notation} \\
\toprule
Symbol & Meaning \\
\midrule
\endfirsthead
\multicolumn{2}{c}{\tablename~\thetable, continued.} \\
\toprule
Symbol & Meaning \\
\midrule
\endhead
\bottomrule
\endlastfoot
$\eta$ & Step size \\
$\beta$ & Momentum coefficient \\
$\beta_1$ & First-moment decay rate \\
$\beta_2$ & Second-moment decay rate \\
$\epsilon$ & Numerical stability constant \\
\midrule
$d$ & Number of parameters \\
$L$ & Loss \\
$H$ & Hessian of the loss \\
$\lamone(H)$ & Sharpness: the largest eigenvalue of $H$ \\
$\lambda_i$, $\hat{u}_i$ & Eigenvalues and eigenvectors of $H$ \\
$\thr$ & Sharpness threshold \\
$t_c$ & Crossing time: first iteration with sharpness above $\thr$ \\
$\widetilde{H}_t$ & Hessian averaged along a gradient step \\
$w_t$ & Position at iteration $t$ \\
$m_t$ & Momentum at iteration $t$ \\
$\nu_t$ & Second moment at iteration $t$ \\
$z = (w, m)$ & Phase-space point \\
$\Phi$ & Update map of the optimizer \\
$P(\nu)$ & Preconditioner $\diag(\sqrt{\nu}) + \epsilon I_d$  \\
$\hat P$ & Preconditioner at the bias-corrected second moment \\
$v^{\odot 2}$ & Elementwise square of a vector $v$ \\
\midrule
$\wbar$ & Midpoint of consecutive positions \\
$\mbar$ & Midpoint of consecutive momenta \\
$\nubar$ & Second moment of the flows \\
$\zbar = (\wbar, \mbar)$ & Phase-space midpoint \\
$\dhat_t$ & Half-displacement of consecutive iterates \\
$a_t$ & Oscillation envelope of the discrete iterates  \\
$\delta$ & Position half-difference (extent of the rod) \\
$\gamma$ & Momentum half-difference \\
$\Delta = (\delta, \gamma)$ & Phase-space half-difference \\
$\Sigma$ & Extent matrix (rod flow); covariance matrix (central flow) \\
$\gbar$ & Average of the gradients at the rod's endpoints \\
$\gjump$ & Half-difference of the gradients at the rod's endpoints \\
$\overline{g^{\otimes 2}}$ & Average outer product of the endpoint gradients \\
$\Sigma^\ast_{\mathrm{EC}}$, $\Sigma^\ast_{\mathrm{SM}}$ & Fixed points of the extent: covariance and second moment \\
$\Kmat$ & Structure matrix \\
$\mathcal{L}$  & Generating function of a structured gradient method  \\
\midrule
$\mathcal{B}$ & Edge coupling of a structured gradient method \\
$\Ac_\eta$, $\Psic$ & Edge coupling; centered edge coupling \\
$\Ac_{\eta,\nu}$, $\Psi_{\eta,\nu}$ & Edge couplings of RMSProp \\
$\mathcal{B}_{\eta,\beta}$, $\Upsilon_{\eta,\beta}$ & Phase-space edge coupling; centered phase-space edge coupling \\
$\mathcal{B}_{\eta,\beta_1,\nu}$, $\Upsilon_{\eta,\beta_1,\nu}$ & Phase-space edge couplings of Adam \\
$\widetilde{\Psi}$, $\widetilde{\Upsilon}$ & Truncated edge couplings\\
$\Upsilon^{\mathrm{red}}$ & Reduced edge coupling \\
$\widetilde{\Upsilon}^{\mathrm{red}}$ & Truncated reduced edge coupling \\
\midrule
$\Lambda$ & Slack $\tfrac{2}{\eta} I_d - H$ \\
$\dot{\Lambda}$ & Rate of change of the slack \\
$\Lambda_{\mathrm{RMSProp}}$, $\Lambda_{\mathrm{HB}}$, $\Lambda_{\mathrm{Adam}}$ & Slacks of RMSProp, heavy ball, and Adam \\
$U$ & Orthonormal basis of the critical directions \\
$k$ & Number of critical directions \\
$X$ & Covariance expressed in the basis $U$ \\
$\sigma^2$ & Squared magnitude of the oscillation \\
$\mathcal{K}(\wbar)$ & Critical face \\
$\Omega$ & Feasible region \\
\midrule
$s$ & Oscillation amplitude \\
$x$ & Excess sharpness\\
$r$ & Progressive-sharpening constant \\
$c$ & Sharpness-reduction coefficient \\
$T$ & Averaging window \\
$\hat{v}$ & Axis of the rod \\
$\Sigma_{\mathrm{rod}}$, $\Sigma_{\mathrm{central}}$ & Extent matrix of rod flow; covariance matrix of central flow \\
$\mathrm{PR}$ & Participation ratio $(\operatorname{tr}\Sigma)^2/\operatorname{tr}(\Sigma^2)$ \\
\end{longtable}
}

%% file: appendix/B-structured.tex
\section{Edge Couplings of Structured Gradient Methods}
\label{app:structured}

This appendix provides further information about the edge coupling for arbitrary structured gradient methods.

\subsection{Structured Gradient Methods}
\label{app:structured-linear}

Consider a linear optimizer of the form
\begin{equation}
	z_{t+1} = z_t + J\nabla\mathcal{G}(z_t),
	\label{eq:app-linear}
\end{equation}
where $J$ is a constant matrix and $\mathcal{G}$ is a generating function. We will call such an optimizer a \emph{linear gradient method}, since its update is linear in the gradient of $\mathcal{G}$. Importantly, we are \textit{not} assuming that $J$ is symmetric.

We will show that if an edge coupling generates the residuals as its derivatives, then the structure matrix must be symmetric. Let $\mathcal{B}(z, z')$ be a function that is symmetric under the exchange $z \leftrightarrow z'$. Suppose that $\mathcal{B}$ satisfies
\begin{equation}
	\nabla_z\mathcal{B}(z, z') = \Kmat^{-1}\bigl(z + J\nabla\mathcal{G}(z) - z'\bigr)
	\label{eq:app-converse-hyp}
\end{equation}
for some constant invertible matrix $\Kmat$. By the symmetry of $\mathcal{B}$, we also have that
\begin{equation}
	\nabla_{z'}\mathcal{B}(z, z') = \Kmat^{-1}\bigl(z' + J\nabla\mathcal{G}(z') - z\bigr).
\end{equation}
By differentiating the first equation with respect to $z'$ and the second equation with respect to $z$, we can obtain the mixed second derivatives of $\mathcal{B}$:
\begin{equation}
	\nabla_{z'}\nabla_z\mathcal{B} = -\Kmat^{-1},
	\qquad
	\nabla_z\nabla_{z'}\mathcal{B} = -\Kmat^{-1}.
\end{equation}
Because mixed derivatives commute, we have that
\begin{equation}
\nabla_{z'}\nabla_z\mathcal{B} = (\nabla_z\nabla_{z'}\mathcal{B})^T \implies K = K^T.
\end{equation}
We have that $K$ is symmetric.

Next, we can subtract from $\mathcal{B}$ the quadratic term of a structured edge coupling and take the derivative with respect to $z$:
\begin{align}
	\nabla_z\Bigl[\mathcal{B}(z, z') - \tfrac{1}{2}(z - z')^\top\Kmat^{-1}(z - z')\Bigr]
	&= \Kmat^{-1}\bigl(z + J\nabla\mathcal{G}(z) - z'\bigr) - \Kmat^{-1}(z - z') \nonumber \\
	&= \Kmat^{-1}J\nabla\mathcal{G}(z).
\end{align}
The right-hand side doesn't depend on $z'$. That means there exists a function $\mathcal{L}$ such that
\begin{equation}
\Kmat^{-1}J\nabla\mathcal{G}(z) = \nabla \mathcal{L}(z)
\end{equation}
By taking the anti-derivative of both sides with respect to $z$, it then follows:
\begin{equation}
	\mathcal{B}(z, z') - \tfrac{1}{2}(z - z')^\top\Kmat^{-1}(z - z') = \mathcal{L}(z) + f(z')
\end{equation}
for some function $f$. By the symmetry of $\mathcal{B}$, $f$ equals $\mathcal{L}$ up to an additive constant $C$. Setting $C=0$, it then follows that:
\begin{equation}
	\mathcal{B}(z, z') = \mathcal{L}(z) + \mathcal{L}(z') + \tfrac{1}{2}(z - z')^\top\Kmat^{-1}(z - z').
\end{equation}
The above shows that \textit{if} an edge coupling exists for a linear gradient method, then the associated structure matrix must be symmetric. A reasonable follow-up question is: For a given linear gradient method, can we find a mathematically equivalent structured gradient method? Specifically, for a matrix $J$ and generating function $\mathcal{G}$, do there exist a symmetric structure matrix $\Kmat$ and a generating function $\mathcal{L}$ such that
\begin{equation}
	J\nabla\mathcal{G} = \Kmat\nabla\mathcal{L}?
\end{equation}
Let $z = (w, m) \in \mathbb{R}^{2d}$, where $w, m \in \mathbb{R}^d$. Consider generating functions of the form
\begin{equation}
	\mathcal{G}(z) = L(w) + \tfrac{1}{2} z^\top Q z,
\end{equation}
where $Q$ is a constant symmetric matrix. Many common optimizers are generated by functions of this form, including heavy-ball momentum---whose generating function is its Hamiltonian. We will show that every linear gradient method generated by such a function can be put in structured gradient form.

Recall that, assuming $\mathcal{L}$ and $\Kmat$ exist, we have
\begin{equation}
	\nabla\mathcal{L}(z) = \Kmat^{-1} J \nabla\mathcal{G}(z).
	\label{eq:app-gradient}
\end{equation}
Differentiating both sides with respect to $z$ gives
\begin{equation}
	\nabla^2 \mathcal{L}(z) = \Kmat^{-1} J \nabla^2 \mathcal{G}(z).
\end{equation}
Since mixed partial derivatives commute, $\nabla^2 \mathcal{L}(z)$ is symmetric. It follows that
\begin{equation}
	\Kmat^{-1} J \nabla^2 \mathcal{G}(z) \text{ is symmetric for all } z.
	\label{eq:app-key}
\end{equation}
We can use this constraint to solve for $K$.

For the Hessian of $\mathcal{G}$, we have
\begin{equation}
	\nabla^2\mathcal{G}(z) = \begin{pmatrix} H(w) & 0 \\ 0 & 0 \end{pmatrix} + Q.
\end{equation}
As the structure matrix should not depend on the specific loss function, \eqref{eq:app-key} must hold for every symmetric $H$---including $H = 0$. It must therefore hold for the loss term and the quadratic term separately:
\begin{equation}
	\Kmat^{-1} J \begin{pmatrix} H & 0 \\ 0 & 0 \end{pmatrix} \text{ is symmetric}
	\qquad \text{and} \qquad
	\Kmat^{-1} J Q \text{ is symmetric.}
\end{equation}
For the loss term, we have that:
\begin{equation}
	\Kmat^{-1}J\begin{pmatrix} H & 0 \\ 0 & 0 \end{pmatrix} = \begin{pmatrix} (\Kmat^{-1}J)_{ww}H & 0 \\ (\Kmat^{-1}J)_{mw}H & 0 \end{pmatrix}.
\end{equation}
The upper-right block of the matrix on the right-hand side is zero. For this matrix to be symmetric, the lower-left block must vanish as well. Since this must hold for every symmetric $H$, we can deduce that
\begin{equation}
	(\Kmat^{-1} J)_{mw} = 0.
\end{equation}
Now consider the upper-left block $(\Kmat^{-1} J)_{ww} H$, which must be symmetric for \textit{every} symmetric $H$. The only matrices with this property are multiples of the identity, so
\begin{equation}
	(\Kmat^{-1} J)_{ww} = \alpha I_d
\end{equation}
for some scalar $\alpha$. Combining these two results, the first block column of $\Kmat^{-1} J$ is
\begin{equation}
	\bigl(\Kmat^{-1} J\bigr) \begin{pmatrix} I_d \\ 0 \end{pmatrix} = \alpha \begin{pmatrix} I_d \\ 0 \end{pmatrix}.
\end{equation}
By associativity, this is equivalent to
\begin{equation}
	\Kmat^{-1} \left( J \begin{pmatrix} I_d \\ 0 \end{pmatrix} \right) = \alpha \begin{pmatrix} I_d \\ 0 \end{pmatrix}.
\end{equation}
Assuming $\alpha \neq 0$ and multiplying both sides on the left by $\Kmat / \alpha$, we obtain
\begin{equation}
	\Kmat \begin{pmatrix} I_d \\ 0 \end{pmatrix} = \frac{1}{\alpha} J \begin{pmatrix} I_d \\ 0 \end{pmatrix}.
	\label{eq:app-first-column}
\end{equation}
This fixes the first block column of $\Kmat$. And because $\Kmat$ is symmetric, this also fixes the first block row. Note that for $K$ to be symmetric we require that $J_{ww}$ is symmetric. The only remaining unknown of $\Kmat$ is the $(m, m)$ block:
\begin{equation}
	\Kmat =
	\begin{pmatrix}
		\tfrac{1}{\alpha}J_{ww} & \tfrac{1}{\alpha}J_{mw}^\top \\
		\tfrac{1}{\alpha}J_{mw} & \Kmat_{mm}
	\end{pmatrix}.
\end{equation}
To solve for $\Kmat_{mm}$, we use the condition that $\Kmat^{-1} J Q$ must be symmetric. We assume that the sub-blocks of $J$ and $Q$ are symmetric and commute with one another. Without this assumption, the symmetry of $\Kmat^{-1} J Q$ would impose more scalar equations than $\Kmat_{mm}$ has entries, and in general there would be no solution. We likewise seek $\Kmat_{mm}$ among matrices that commute with these blocks.

Under these assumptions, we can invert $\Kmat$ as we would a $2 \times 2$ matrix:
\begin{equation}
	\Kmat^{-1} = \bigl(\Kmat_{ww}\Kmat_{mm} - \Kmat_{wm}^2\bigr)^{-1}
	\begin{pmatrix} \Kmat_{mm} & -\Kmat_{wm} \\ -\Kmat_{wm} & \Kmat_{ww} \end{pmatrix}.
\end{equation}
Every sub-block of $\Kmat^{-1} J Q$ is then built from commuting symmetric matrices, and is therefore itself symmetric. Hence $\Kmat^{-1} J Q$ is symmetric exactly when its two off-diagonal sub-blocks are equal. Setting them equal and cancelling the common factor $\bigl(\Kmat_{ww}\Kmat_{mm} - \Kmat_{wm}^2\bigr)^{-1}$, we obtain
\begin{equation}
	\Kmat_{mm} (JQ)_{wm} - \Kmat_{wm} (JQ)_{mm} = \Kmat_{ww} (JQ)_{mw} - \Kmat_{wm} (JQ)_{ww}.
\end{equation}
Recall that the first block column of $\Kmat$ is $1/\alpha$ times the first block column of $J$, so $\Kmat_{ww} = J_{ww}/\alpha$ and $\Kmat_{mw} = J_{mw}/\alpha$. Substituting these expressions and using $\Kmat_{wm} = \Kmat_{mw}$, we obtain a linear equation for $\Kmat_{mm}$:
\begin{equation}
	(JQ)_{wm} \Kmat_{mm} = \frac{1}{\alpha} \Bigl[ J_{ww} (JQ)_{mw} + J_{mw} \bigl( (JQ)_{mm} - (JQ)_{ww} \bigr) \Bigr].
\end{equation}
If $(JQ)_{wm}$ is invertible, this equation has exactly one solution:
\begin{equation}
	\Kmat_{mm} = \frac{1}{\alpha} (JQ)_{wm}^{-1} \Bigl[ J_{ww} (JQ)_{mw} + J_{mw} \bigl( (JQ)_{mm} - (JQ)_{ww} \bigr) \Bigr].
\end{equation}
The structure matrix is therefore
\begin{equation}
	\boxed{
		\Kmat = \frac{1}{\alpha}
		\begin{pmatrix}
			J_{ww} & J_{mw} \\
			J_{mw} & (JQ)_{wm}^{-1} \bigl[ J_{ww} (JQ)_{mw} + J_{mw} \bigl( (JQ)_{mm} - (JQ)_{ww} \bigr) \bigr]
		\end{pmatrix}
	}.
	\label{eq:app-K}
\end{equation}
As we have found $\Kmat$, we can integrate $K^{-1} J \nabla \mathcal{G}$ to obtain $\mathcal{L}$. Substituting the gradient of $\mathcal{G}$, and using \eqref{eq:app-first-column}, we have that:
\begin{equation}
\begin{aligned}
	\nabla\mathcal{L}(z) &= K^{-1} J \nabla \mathcal{G}\\
	&= \Kmat^{-1}J\begin{pmatrix} \nabla L(w) \\ 0 \end{pmatrix} + \Kmat^{-1}JQ\,z \\
	&= \alpha\begin{pmatrix} \nabla L(w) \\ 0 \end{pmatrix} + \Kmat^{-1}JQ\,z.
\end{aligned}
\end{equation}
The first term is the gradient of $\alpha L(w)$. Since $\Kmat^{-1}JQ$ is symmetric, the second term is the gradient of $\tfrac{1}{2}z^\top\Kmat^{-1}JQ\,z$. Taking the anti-derivative, we obtain the generating function:
\begin{equation}
	\boxed{
		\mathcal{L}(z) = \alpha L(w) + \tfrac{1}{2}z^\top\Kmat^{-1}JQ\,z.
	}
	\label{eq:app-L}
\end{equation}
Since the structure matrix $\Kmat$ is proportional to $1/\alpha$, the product $\Kmat^{-1} J Q$ is proportional to $\alpha$. It follows that $\mathcal{L}$ is also proportional to $\alpha$, while $\Kmat \nabla\mathcal{L}$ is independent of $\alpha$. We therefore obtain a one-parameter family of valid structured forms---and as long as $(JQ)_{wm}$ is invertible, there are no others. In the rest of the paper we take $\alpha = 1$, so that the loss appears in $\mathcal{L}$ with coefficient one.

\subsection{Heavy Ball}
\label{app:structured-hb}

Heavy ball can be expressed as a linear gradient method by using the Hamiltonian as the generating function \citep{franca2020conformal}. Let $z = (w,m)$ denote a point in phase space. For the Hamiltonian, we have:
\begin{equation}
	\mathcal{H}(w, m) = L(w) + \tfrac{\eta}{2(1-\beta)}\norm{m}^2 .
\end{equation}
We have that the loss plays the role of the potential energy, and the momentum-squared plays the role of the kinetic energy. Note that we have that
\begin{equation}
	Q = \begin{pmatrix} 0 & 0 \\ 0 & \tfrac{\eta}{1-\beta}I_d \end{pmatrix}.
\end{equation}
The heavy ball update is given as $z_{t+1} = z_t + J\nabla\mathcal{H}(z_t)$, where
\begin{align}
	J &= -(1-\beta)\begin{pmatrix} \eta I_d & \beta I_d \\ -I_d & \frac{1-\beta}{\eta}I_d \end{pmatrix} \nonumber \\
	&= -\frac{(1-\beta)(1+\beta)}{2}\begin{pmatrix} 0 & I_d \\ -I_d & 0 \end{pmatrix}
	- (1-\beta)\begin{pmatrix} \eta I_d & -\frac{1-\beta}{2}I_d \\ -\frac{1-\beta}{2}I_d & \frac{1-\beta}{\eta}I_d \end{pmatrix}.
\end{align}
The first part of $J$ is antisymmetric. Up to an overall factor, it is the symplectic matrix of Hamiltonian mechanics, which encodes how force drives changes in momentum and momentum drives changes in position. The second part is symmetric and negative definite---it supplies the damping. Since $J$ is not symmetric, the damped formulation is not in structured gradient form.

As $Q$ only involves $Q_{mm}$, we can actually derive a simplified expression for $\Kmat$. The product $JQ$ only has a second block column:
\begin{equation}
	JQ = \begin{pmatrix} 0 & J_{wm}Q_{mm} \\ 0 & J_{mm}Q_{mm} \end{pmatrix}.
\end{equation}
Substituting these blocks into \eqref{eq:app-K}, only the term $J_{mw}(JQ)_{mm}$ survives in the lower-right block. We then have that:
\begin{equation}
\begin{aligned}
	\Kmat_{mm} &= \bigl(J_{wm}Q_{mm}\bigr)^{-1}J_{mw}J_{mm}Q_{mm} \\ &= J_{wm}^{-1}J_{mw}J_{mm}.
\end{aligned}
\end{equation}
Recall that the first block column of $\Kmat$ coincides with that of $J$. We also have that
\begin{equation}
	\begin{aligned}
		\Kmat_{wm} &= J_{mw} \\
		&= J_{wm}\bigl(J_{wm}^{-1}J_{mw}\bigr).
	\end{aligned}
\end{equation}
Thus, the second block column of $\Kmat$ is that of $J$, right-multiplied by $J_{wm}^{-1}J_{mw}$. Altogether,
\begin{equation}
	\Kmat = J\begin{pmatrix} I_d & 0 \\ 0 & J_{wm}^{-1}J_{mw} \end{pmatrix}.
	\label{eq:app-K-momentum}
\end{equation}
Substituting, the generating function is
\begin{equation}
	\mathcal{L}(z) = L(w) + \tfrac{1}{2}m^\top J_{mw}^{-1}J_{wm}Q_{mm}\,m.
\end{equation}
For the case of heavy ball, we have that
\begin{equation}
	J_{wm}^{-1}J_{mw} = \tfrac{1-\beta}{-\beta(1-\beta)}\,I_d = -\tfrac{1}{\beta}I_d .
\end{equation}
The structured form flips the sign of the second block column of $J$ and divides it by $\beta$:
\begin{equation}
	\Kmat = J\begin{pmatrix} I_d & 0 \\ 0 & -\tfrac{1}{\beta}I_d \end{pmatrix}
	= \begin{pmatrix} -\eta(1-\beta)I_d & (1-\beta)I_d \\ (1-\beta)I_d & \tfrac{(1-\beta)^2}{\eta\beta}I_d \end{pmatrix}.
\end{equation}
And for the generating function, we have that the kinetic energy is multiplied by $-\beta$ :
\begin{equation}
	\mathcal{L}(w, m) = L(w) - \tfrac{\eta\beta}{2(1-\beta)}\norm{m}^2.
\end{equation}
\subsection{Other Structured Gradient Methods}
\label{app:structured-others}

We briefly cover some other structured gradient methods that admit edge couplings.

\paragraph{Gradient descent--ascent.} Simultaneous gradient descent--ascent on $f(u, v)$ \citep{arrow1958studies} descends in $u$ and ascends in $v$:
\begin{equation}
	u_{t+1} = u_t - \eta\nabla_u f(u_t, v_t),
	\qquad
	v_{t+1} = v_t + \eta\nabla_v f(u_t, v_t).
\end{equation}
For the generating function and structure matrix, we have:
\begin{equation}
	\mathcal{L}(u, v) = f(u, v),
	\qquad
	K = \begin{pmatrix} -\eta I & 0 \\ 0 & \eta I \end{pmatrix}.
\end{equation}
For the edge coupling, we have:
\begin{equation}
	\mathcal{B}(u, v, u', v') = f(u, v) + f(u', v') - \tfrac{1}{2\eta}\norm{u - u'}^2 + \tfrac{1}{2\eta}\norm{v - v'}^2 .
\end{equation}

\paragraph{Optimistic gradient descent.} Optimistic gradient descent \citep{popov1980modification, daskalakis2018training} extrapolates from the previous gradient:
\begin{equation}
	w_{t+1} = w_t - 2\eta\nabla L(w_t) + \eta\nabla L(w_{t-1}).
\end{equation}
In the coordinates $(w_t, w_{t-1})$, optimistic gradient descent is not a structured gradient method, since the loss function is evaluated at two different points. We can fix this with a change of coordinates. Let $q_t := \nabla L(w_{t-1})$ denote the previous gradient, and let $z = (w, q)$. Written in terms of differences, the update in these coordinates is
\begin{equation}
	\begin{aligned}
		w_{t+1} - w_t &= -2\eta\nabla L(w_t) + \eta q_t, \\
		q_{t+1} - q_t &= \nabla L(w_t) - q_t,
	\end{aligned}
\end{equation}
In this form, we have that optimistic gradient descent is a linear gradient method with the following generating function and structure matrix:
\begin{equation}
	\mathcal{G}(w, q) = L(w) + \tfrac{1}{2}\norm{q}^2,
	\qquad
	J = \begin{pmatrix} -2\eta I_d & \eta I_d \\ I_d & -I_d \end{pmatrix}.
\end{equation}
Since the generating function is a loss function plus a quadratic form, optimistic gradient descent can be written as a structured gradient method, with $\mathcal{L}$ and $K$ given by
\begin{equation}
	\mathcal{L}(w, q) = L(w) + \tfrac{\eta}{2}\norm{q}^2,
	\qquad
	\Kmat = \begin{pmatrix} -2\eta I_d & I_d \\ I_d & -\tfrac{1}{\eta}I_d \end{pmatrix}.
\end{equation}
For the edge coupling, we have:
\begin{equation}
	\begin{aligned}
		\mathcal{B}(w, q, w', q')
		&= L(w) + L(w') + \tfrac{\eta}{2}\bigl(\norm{q}^2 + \norm{q'}^2\bigr) \\
		&\quad - \tfrac{1}{2\eta}\norm{w - w'}^2 - (w - w')\cdot(q - q') - \eta\norm{q - q'}^2 .
	\end{aligned}
\end{equation}

%% file: appendix/C-nesterov.tex
\section{Nesterov Momentum}
\label{app:nesterov}

This appendix derives the edge coupling, rod flow, and central flow for Nesterov momentum.

\subsection{Setup}
\label{app:nesterov-setup}

Nesterov momentum \citep{nesterov1983method} is an alternative to heavy-ball momentum. The two differ only in where the gradient in the momentum update is evaluated---a difference that shows up in continuous time as a gradient-correction term \citep{shi2022understanding}. Rather than evaluating the gradient at the current iterate, Nesterov momentum evaluates it at a \emph{look-ahead} point: the location the iterate would reach if it were updated by the momentum term alone \citep{sutskever2013importance}. The update equations are:
\begin{equation}
	\begin{aligned}
		m_{t+1} &= \beta m_t + (1-\beta)\nabla L(w_t - \eta\beta m_t), \\
		w_{t+1} &= w_t - \eta m_{t+1}.
	\end{aligned}
	\label{eq:nag}
\end{equation}
Nesterov momentum is arguably more natural when expressed in look-ahead coordinates. Define the look-ahead point:
\begin{equation}
	\theta_t = w_t - \eta\beta m_t .
\end{equation}
In the coordinates $(\theta, m)$, the update equations no longer involve a look-ahead. The gradient is evaluated at the current iterate itself:
\begin{equation}
	\begin{aligned}
		m_{t+1} &= \beta m_t + (1-\beta)\nabla L(\theta_t), \\
		\theta_{t+1} &= \theta_t - \eta\bigl[\beta m_{t+1} + (1-\beta)\nabla L(\theta_t)\bigr].
	\end{aligned}
	\label{eq:nag-theta}
\end{equation}
We will work in look-ahead coordinates throughout.

\subsection{Edge Coupling}
\label{app:nesterov-coupling}

In look-ahead coordinates, Nesterov momentum is a structured gradient method. Written as differences, the update is
\begin{equation}
	\begin{aligned}
		\theta_{t+1} - \theta_t &= -\eta\bigl[\beta^2 m_t + (1-\beta^2)\nabla L(\theta_t)\bigr], \\
		m_{t+1} - m_t &= (1-\beta)\bigl[\nabla L(\theta_t) - m_t\bigr],
	\end{aligned}
\end{equation}
which is $\Kmat\nabla_{(\theta, m)}\mathcal{L}$ for the generating function and structure matrix
\begin{equation}
	\mathcal{L}(\theta, m) = L(\theta) - \tfrac{\eta\beta^2}{2(1-\beta)}\norm{m}^2,
	\qquad
	\Kmat =
	\begin{pmatrix}
		-\eta(1-\beta^2) I_d & (1-\beta) I_d \\
		(1-\beta) I_d & \tfrac{(1-\beta)^2}{\eta\beta^2} I_d
	\end{pmatrix},
	\label{eq:nag-K}
\end{equation}
The inverse of the structure matrix is
\begin{equation}
	\Kmat^{-1} =
	\begin{pmatrix}
		-\tfrac{1}{\eta} I_d & \tfrac{\beta^2}{1-\beta} I_d \\
		\tfrac{\beta^2}{1-\beta} I_d & \tfrac{\eta\beta^2(1+\beta)}{1-\beta} I_d
	\end{pmatrix}.
\end{equation}
Let $z = (\theta, m)$ and $z' = (\theta', m')$. We then obtain the phase-space edge coupling of Nesterov momentum \citep{regis2026variational}:
\begin{equation}
	\begin{aligned}
		\mathcal{B}^{\mathrm{NAG}}_{\eta,\beta}(z, z')
		&= L(\theta) + L(\theta') - \tfrac{1}{2\eta}\|\theta - \theta'\|^2
		+ \tfrac{\beta^2}{1-\beta}(m - m')\cdot(\theta - \theta') \\
		&\quad + \tfrac{\eta\beta^3}{2(1-\beta)}\|m - m'\|^2
		- \tfrac{\eta\beta^2}{1-\beta}\, m\cdot m' .
	\end{aligned}
	\label{eq:nag-B}
\end{equation}
Its critical points are the fixed points and period-two orbits of Nesterov momentum.

Define the centered coordinates
\begin{equation}
	\bar{z} = \begin{pmatrix} \thetabar \\ \mbar \end{pmatrix} = \frac{z + z'}{2}, \qquad \Delta = \begin{pmatrix} \delta \\ \gamma \end{pmatrix} = \frac{z - z'}{2}.
\end{equation}
In these coordinates, the centered phase-space edge coupling is
\begin{align}
	\Upsilon^{\mathrm{NAG}}_{\eta,\beta}(\bar{z}, \Delta)
	&:= \tfrac{1}{2}\mathcal{B}^{\mathrm{NAG}}_{\eta,\beta}(\bar{z} + \Delta, \bar{z} - \Delta) \nonumber \\
	&= \tfrac{1}{2}\bigl[L(\thetabar + \delta) + L(\thetabar - \delta)\bigr] - \tfrac{1}{\eta}\|\delta\|^2 + \tfrac{2\beta^2}{1-\beta}\,\gamma\cdot\delta \nonumber \\
	&\quad + \tfrac{\eta\beta^2(1+2\beta)}{2(1-\beta)}\|\gamma\|^2 - \tfrac{\eta\beta^2}{2(1-\beta)}\|\mbar\|^2 .
	\label{eq:nag-Ups}
\end{align}
To lighten notation, we will drop the superscript and subscripts for the remainder of this appendix---writing $\mathcal{B}$ and $\Upsilon$ for the edge couplings of Nesterov momentum.

\subsection{Rod Flow}
\label{app:nesterov-rod}

For the Nesterov midpoint rod flow, we have:
\begin{equation}
	\boxed{
		\ddt{}\begin{pmatrix} \bar{z} \\ \Delta \end{pmatrix}
		=
		\begin{pmatrix} \Kmat & 0 \\ 0 & -\Kmat \end{pmatrix}
		\nabla_{(\bar{z}, \Delta)} \Upsilon
	}
\end{equation}
Writing out the individual ODEs:
\begin{equation}
	\begin{aligned}
		\ddt{\thetabar} &= -\eta\bigl[\beta^2\mbar + (1-\beta^2)\gbar\bigr], \\
		\ddt{\mbar} &= (1-\beta)\bigl[\gbar - \mbar\bigr], \\
		\ddt{\delta} &= -2\delta + \eta\bigl[\beta^2\gamma + (1-\beta^2)\gjump\bigr], \\
		\ddt{\gamma} &= -(1+\beta)\gamma - (1-\beta)\gjump.
	\end{aligned}
	\label{eq:nag-rodflow}
\end{equation}
For the endpoint rod flow, we have:
\begin{equation}
	\boxed{
		\ddt{}\begin{pmatrix} z \\ z' \end{pmatrix}
		=
		\begin{pmatrix} 0 & \Kmat \\ \Kmat & 0 \end{pmatrix}
		\nabla_{(z, z')} \mathcal{B}
	}
\end{equation}
Writing out the individual ODEs:
\begin{equation}
	\begin{aligned}
		\ddt{\theta} &= -\bigl(\theta - \bigl[\theta' - \eta\bigl(\beta^2 m' + (1-\beta^2)\nabla L(\theta')\bigr)\bigr]\bigr), \\
		\ddt{m} &= -\bigl(m - \bigl[\beta m' + (1-\beta)\nabla L(\theta')\bigr]\bigr), \\
		\ddt{\theta'} &= -\bigl(\theta' - \bigl[\theta - \eta\bigl(\beta^2 m + (1-\beta^2)\nabla L(\theta)\bigr)\bigr]\bigr), \\
		\ddt{m'} &= -\bigl(m' - \bigl[\beta m + (1-\beta)\nabla L(\theta)\bigr]\bigr).
	\end{aligned}
	\label{eq:nag-endpoint}
\end{equation}
Each endpoint regresses toward the Nesterov image of the other endpoint.

\subsection{Central Flow}
\label{app:nesterov-central}

We can also derive a central flow for Nesterov momentum. As with heavy ball, we first eliminate the momentum oscillation. Completing the square,
\begin{equation}
	\frac{2\beta^2}{1-\beta}\gamma\cdot\delta
	+ \frac{\eta\beta^2(1+2\beta)}{2(1-\beta)}\|\gamma\|^2
	=
	\frac{\eta\beta^2(1+2\beta)}{2(1-\beta)}\left\|\gamma + \frac{2}{\eta(1+2\beta)}\delta\right\|^2
	- \frac{2\beta^2}{\eta(1-\beta)(1+2\beta)}\|\delta\|^2 .
\end{equation}
Stationarity in $\gamma$ gives $\gamma = -\tfrac{2}{\eta(1+2\beta)}\delta$. Eliminating $\gamma$ gives the reduced edge coupling
\begin{equation}
	\Upsilon^{\mathrm{red}}(\bar{z}, \delta)
	=
	\tfrac{1}{2}\bigl[L(\thetabar + \delta) + L(\thetabar - \delta)\bigr]
	- \tfrac{1+\beta}{\eta(1-\beta)(1+2\beta)}\|\delta\|^2
	- \tfrac{\eta\beta^2}{2(1-\beta)}\|\mbar\|^2 .
\end{equation}
We can then approximate the loss about the midpoint $\thetabar$ by its local quadratic approximation. Averaging over the endpoints $\thetabar \pm \delta$ cancels the first-order terms, giving
\begin{equation}
	\tfrac{1}{2}\bigl[L(\thetabar + \delta) + L(\thetabar - \delta)\bigr] \approx L(\thetabar) + \tfrac{1}{2} H(\thetabar)[\delta, \delta].
\end{equation}
Substituting into the reduced edge coupling and again replacing $\delta\otimes\delta$ with a general positive semidefinite matrix $\Sigma\succeq0$, we obtain the truncated reduced edge coupling
\begin{equation}
	\widetilde{\Upsilon}^{\mathrm{red}}(\bar{z}, \Sigma)
	=
	L(\thetabar)
	- \tfrac{\eta\beta^2}{2(1-\beta)}\|\mbar\|^2
	- \tfrac12\langle \Lambda_{\mathrm{NAG}}(\thetabar), \Sigma\rangle,
\end{equation}
where
\begin{equation}
	\Lambda_{\mathrm{NAG}}(\thetabar)
	:=
	\tfrac{2(1+\beta)}{\eta(1-\beta)(1+2\beta)} I_d - H(\thetabar).
\end{equation}
We can then maximize the truncated reduced edge coupling over $\Sigma\succeq0$. This gives the following static conditions:
\begin{equation}
	\Sigma\succeq0,
	\qquad
	\Lambda_{\mathrm{NAG}}\succeq0,
	\qquad
	\langle \Lambda_{\mathrm{NAG}}, \Sigma\rangle = 0.
\end{equation}
Taking derivatives of the truncated reduced edge coupling with respect to the midpoint variables gives:
\begin{align}
	\partial_{\thetabar}\widetilde{\Upsilon}^{\mathrm{red}}
	&= \nabla L(\thetabar) + \tfrac{1}{2}\nabla_{\thetabar}\langle H(\thetabar), \Sigma\rangle, \\
	\partial_{\mbar}\widetilde{\Upsilon}^{\mathrm{red}}
	&= -\tfrac{\eta\beta^2}{1-\beta}\mbar .
\end{align}
As with the Nesterov rod flow, we apply the matrix $\Kmat$ to the gradient of the truncated reduced edge coupling to obtain the central flow ODE for Nesterov momentum:
\begin{equation}
	\ddt{\bar{z}} = \Kmat\,\partial_{\bar{z}}\widetilde{\Upsilon}^{\mathrm{red}}(\bar{z}, \Sigma).
\end{equation}
Writing the ODEs for $\thetabar$ and $\mbar$ individually:
\begin{equation}
	\begin{aligned}
		\ddt{\thetabar}
		&= -\eta\left[\beta^2\mbar + (1-\beta^2)\Bigl(\nabla L(\thetabar) + \tfrac{1}{2}\nabla_{\thetabar}\langle H(\thetabar), \Sigma\rangle\Bigr)\right], \\
		\ddt{\mbar}
		&= (1-\beta)\left[\nabla L(\thetabar) + \tfrac{1}{2}\nabla_{\thetabar}\langle H(\thetabar), \Sigma\rangle - \mbar\right].
	\end{aligned}
	\label{eq:nag-central}
\end{equation}
To determine $\Sigma$ we invoke persistence: feasibility and complementarity must hold along the trajectory.

Let $U$ be an orthonormal basis for the kernel of $\Lambda_{\mathrm{NAG}}$, and $k$ the number of critical directions. Static complementarity requires:
\begin{equation}
	\Sigma = UXU^\top, \qquad X\succeq0 .
\end{equation}
Computing the rate of change of the slack:
\begin{equation}
		\dot{\Lambda}_{\mathrm{NAG}} = - \nabla_{\bar{\theta}} H[\dot{\bar{\theta}}]
\end{equation}
Persistence of feasibility requires that the slack not decrease along the sharp directions:
\begin{equation}
	U^\top\dot{\Lambda}_{\mathrm{NAG}}U\succeq0 .
\end{equation}
Persistence of complementarity requires that there be no oscillations in directions whose slack is increasing:
\begin{equation}
	\langle U^\top\dot{\Lambda}_{\mathrm{NAG}}U, X\rangle = 0 .
\end{equation}
These dynamic conditions, together with $X\succeq0$, define a semidefinite complementarity problem that determines $X$. Altogether, we have that the following conditions define the central flow for Nesterov momentum:
\begin{equation}
	\boxed{
		\begin{aligned}
			\ddt{\bar{z}} &= \Kmat\, \partial_{\bar{z}} \widetilde{\Upsilon}^{\mathrm{red}}(\bar{z}, \Sigma), \qquad \Sigma = U X U^\top, \\[4pt]
			X &\succeq 0, \qquad U^\top \dot{\Lambda}_{\mathrm{NAG}} U \succeq 0, \qquad \langle U^\top \dot{\Lambda}_{\mathrm{NAG}} U, X \rangle = 0.
		\end{aligned}
	}
\end{equation}

%% file: appendix/D-adamw.tex
\section{AdamW}
\label{app:adamw}

This appendix derives the edge coupling, rod flow, and central flow for AdamW.

\subsection{Setup}
\label{app:adamw-setup}

AdamW \citep{loshchilov2019decoupled} is a modification of Adam that decouples weight decay from the gradient-based update. Rather than adding an $\ell_2$ penalty to the loss function---which would pass the decay through the momentum and the preconditioner---AdamW shrinks the weights directly. Omitting bias correction, the update equations are:
\begin{equation}
	\begin{aligned}
		m_{t+1}
		&= \beta_1 m_t + (1-\beta_1)\nabla L(w_t), \\
		\nu_{t+1}
		&= \beta_2 \nu_t + (1-\beta_2)\nabla L(w_t)^{\odot 2}, \\
		w_{t+1}
		&= \kappa\, w_t - \eta P_{t+1}^{-1}m_{t+1},
	\end{aligned}
	\label{eq:adamw}
\end{equation}
where $\zeta \geq 0$ is the weight-decay coefficient and
\begin{equation}
	\kappa := 1 - \eta\zeta
\end{equation}
is the factor by which the decay shrinks the weights at each step. Following the PyTorch implementation \citep{paszke2019pytorch}, the decay is scaled by the step size, so it enters the update only through the product $\eta\zeta$. At $\zeta = 0$, we have $\kappa = 1$---and AdamW reduces down to Adam.

\subsection{Edge of Stability}
\label{app:adamw-eos}

We derive the edge-of-stability condition for AdamW.

We will treat the preconditioner $P$ as fixed. Consider a quadratic loss $L(w) = \tfrac{1}{2} w^\top H w$. The AdamW updates are then
\begin{align}
	m_{t+1} &= \beta_1 m_t + (1-\beta_1) H w_t, \\
	w_{t+1} &= \kappa w_t - \eta P^{-1} m_{t+1},
\end{align}
Since the update equations are linear in $(w, m)$, we can express them as a single matrix equation:
\begin{equation}
	\begin{pmatrix} w_{t+1} \\ m_{t+1} \end{pmatrix}
	= A \begin{pmatrix} w_t \\ m_t \end{pmatrix},
	\qquad
	A =
	\begin{pmatrix}
		\kappa I_d - \eta(1-\beta_1)P^{-1}H & -\eta\beta_1 P^{-1} \\
		(1-\beta_1)H & \beta_1 I_d
	\end{pmatrix}.
\end{equation}
Let $u$ be an eigenvector of $P^{-1} H$ with eigenvalue $\lambda$, so that $H u = \lambda P u$. For each eigenvector $u$, applying $A$ to the vectors $(u, 0)$ and $(0, Pu)$ gives
\begin{align}
	A \begin{pmatrix} u \\ 0 \end{pmatrix}
	&= \bigl(\kappa - \eta(1-\beta_1)\lambda\bigr) \begin{pmatrix} u \\ 0 \end{pmatrix}
	+ (1-\beta_1)\lambda \begin{pmatrix} 0 \\ Pu \end{pmatrix}, \\
	A \begin{pmatrix} 0 \\ Pu \end{pmatrix}
	&= -\eta\beta_1 \begin{pmatrix} u \\ 0 \end{pmatrix}
	+ \beta_1 \begin{pmatrix} 0 \\ Pu \end{pmatrix}.
\end{align}
We have that $A$ maps the span of these two vectors to itself. In the basis $\{(u, 0), (0, Pu)\}$, it acts as the $2 \times 2$ matrix
\begin{equation}
	A_\lambda =
	\begin{pmatrix}
		\kappa - \eta(1-\beta_1)\lambda & -\eta\beta_1 \\
		(1-\beta_1)\lambda & \beta_1
	\end{pmatrix},
\end{equation}
whose characteristic polynomial is
\begin{equation}
	p(\alpha) = \alpha^2 - \bigl(\kappa + \beta_1 - \eta(1-\beta_1)\lambda\bigr)\alpha + \kappa\beta_1 .
\end{equation}
The iterates are stable along $u$ when both roots of $p$ lie strictly inside the unit circle---with instability setting in when a root crosses the unit circle. Consider the crossing at $\alpha = -1$, which corresponds to the iterates oscillating. The value $\alpha = -1$ is a root of $p$ when
\begin{equation}
	p(-1) = (1+\kappa)(1+\beta_1) - \eta(1-\beta_1)\lambda = 0.
\end{equation}
Rearranging to isolate $\lambda$, we obtain the edge of stability condition for AdamW:
\begin{equation}
	\boxed{
		\lammax(P^{-1/2}HP^{-1/2}) =  \Bigl(1 - \frac{\eta\zeta}{2}\Bigr)\frac{2(1+\beta_1)}{\eta(1-\beta_1)}.
	}
	\label{eq:adamw-eos}
\end{equation}
\subsection{Edge Coupling}
\label{app:adamw-coupling}

We would like to derive an edge coupling for AdamW. We will again treat the preconditioner $P$ as fixed. Written as difference equations, the update equations of AdamW are
\begin{equation}
	\begin{aligned}
		w_{t+1} - w_t &= -\eta\zeta w_t - \eta\beta_1 P^{-1}m_t - \eta(1-\beta_1)P^{-1}\nabla L(w_t), \\
		m_{t+1} - m_t &= (1-\beta_1)\nabla L(w_t) - (1-\beta_1)m_t.
	\end{aligned}
\end{equation}
The update can be split into two terms: a term that is linear in $z$ and a term that is linear in the gradient of the loss function.
\begin{equation}
	z_{t+1} - z_t
	=  (1-\beta_1)\begin{pmatrix} -\eta P^{-1} \\ I_d \end{pmatrix}\nabla L(w_t) + R z_t,
	\qquad
	R
	=
	\begin{pmatrix}
		-\eta\zeta I_d & -\eta\beta_1 P^{-1} \\
		0 & -(1-\beta_1)I_d
	\end{pmatrix}.
	\label{eq:adamw-split}
\end{equation}
The weight decay enters only through the $(w, w)$ block of $R$.

We will assume that the generating function has the form $\mathcal{L}(z) = L(w) + \tfrac{1}{2} z^\top Q z$. We seek a symmetric structure matrix $\Kmat$ and a symmetric matrix $Q$ such that $z_{t+1} - z_t = \Kmat \nabla\mathcal{L}(z_t)$. Taking the derivative of $\mathcal{L}$, we have that:
\begin{equation}
	\Kmat \nabla\mathcal{L}(z) = \Kmat \begin{pmatrix} I_d \\ 0 \end{pmatrix} \nabla L(w) + \Kmat Q z.
\end{equation}
Setting this equal to the update yields:
\begin{equation}
\Kmat \begin{pmatrix} I_d \\ 0 \end{pmatrix} \nabla L(w) + \Kmat Q z =  (1-\beta_1)\begin{pmatrix} -\eta P^{-1} \\ I_d \end{pmatrix}\nabla L(w) + Rz
\end{equation}
Since the above equation must hold for arbitrary loss functions, the gradient terms and the linear terms must match separately. Matching the gradient terms, we have that
\begin{equation}
\Kmat \begin{pmatrix} I_d \\ 0 \end{pmatrix} = (1-\beta_1)\begin{pmatrix} -\eta P^{-1} \\ I_d \end{pmatrix},
\end{equation}
which fixes the first block column of $\Kmat$---it must equal the coefficient of $\nabla L(w)$ in the update. And because $\Kmat$ is symmetric, this also fixes the first block row as well. The only remaining unknown of $K$ is the $(m, m)$ block:
\begin{equation}
	\Kmat
	=
	\begin{pmatrix}
		-\eta(1-\beta_1)P^{-1} & (1-\beta_1)I_d \\
		(1-\beta_1)I_d & \Kmat_{mm}
	\end{pmatrix}.
\end{equation}
Matching the linear terms, we have that
\begin{equation}
\Kmat Q = R.
\end{equation}
Since $Q$ is symmetric, its $(m, w)$ block is the transpose of its $(w, m)$ block, so there are four unknowns: $\Kmat_{mm}$, $Q_{ww}$, $Q_{wm}$, and $Q_{mm}$. Writing out $\Kmat Q = R$ block by block, we obtain four equations---one for each block of $R$:
\begin{align}
		-\eta(1-\beta_1)P^{-1}Q_{ww} + (1-\beta_1)Q_{wm}^\top &= -\eta\zeta I_d, \\
		-\eta(1-\beta_1)P^{-1}Q_{wm} + (1-\beta_1)Q_{mm} &= -\eta\beta_1 P^{-1}, \\
		(1-\beta_1)Q_{ww} + \Kmat_{mm}Q_{wm}^\top &= 0, \\
		(1-\beta_1)Q_{wm} + \Kmat_{mm}Q_{mm} &= -(1-\beta_1)I_d.
\end{align}
We can solve these four equations for the four unknown sub-blocks. In order for the above system of equations to have a solution, it is crucial that every \textit{known} matrix in the equations above commutes with $P$. Otherwise, the number of independent scalar equations would exceed the effective number of scalar unknowns. It is also crucial that $P$ be symmetric: the solutions for the diagonal blocks $\Kmat_{mm}$, $Q_{ww}$, and $Q_{mm}$ are expressed in terms of $P$, and these blocks must themselves be symmetric.

Solving the third equation for $Q_{ww}$, we obtain:
\begin{equation}
	Q_{ww} = -\tfrac{1}{1-\beta_1}\Kmat_{mm}Q_{wm}^\top.
\end{equation}
We can then substitute into the first equation to obtain an equation in only $Q_{wm}$ and $\Kmat_{mm}$:
\begin{equation}
	-\eta(1-\beta_1)P^{-1}\bigl(-\tfrac{1}{1-\beta_1}\Kmat_{mm}Q_{wm}^\top\bigr) + (1-\beta_1)Q_{wm}^\top = -\eta\zeta I_d.
\end{equation}
Similarly, using the second equation, we can solve for $Q_{mm}$:
\begin{equation}
	Q_{mm} = \eta P^{-1}Q_{wm} - \tfrac{\eta\beta_1}{1-\beta_1}P^{-1}.
\end{equation}
We can then substitute this into the fourth equation to obtain:
\begin{equation}
	(1-\beta_1)Q_{wm} + \Kmat_{mm}\bigl(\eta P^{-1}Q_{wm} - \tfrac{\eta\beta_1}{1-\beta_1}P^{-1}\bigr) = -(1-\beta_1)I_d.
\end{equation}
As we are assuming that the diagonal sub-blocks can be expressed in terms of $P$, they must commute with $P$. We then have that $P^{-1}Q_{ww}$ is symmetric---from which it follows that $Q_{wm}$ is symmetric as well. Rearranging, we have the following system of equations in $Q_{wm}$ and $\Kmat_{mm}$:
\begin{equation}
	\begin{aligned}
		\bigl[(1-\beta_1)I_d + \eta P^{-1}\Kmat_{mm}\bigr]Q_{wm} &= -\eta\zeta I_d, \\
		\bigl[(1-\beta_1)I_d + \eta P^{-1}\Kmat_{mm}\bigr]Q_{wm} &= \tfrac{\eta\beta_1}{1-\beta_1}P^{-1}\Kmat_{mm} - (1-\beta_1)I_d.
	\end{aligned}
\end{equation}
The left-hand sides are the same. Setting the right-hand sides equal gives us $\Kmat_{mm}$:
\begin{equation}
	-\eta\zeta I_d = \tfrac{\eta\beta_1}{1-\beta_1}P^{-1}\Kmat_{mm} - (1-\beta_1)I_d \implies \Kmat_{mm} = \tfrac{(1-\beta_1)(\kappa-\beta_1)}{\eta\beta_1}P.
\end{equation}
We then have for the structure matrix:
\begin{equation}
	\Kmat_{\nu}
	=
	\begin{pmatrix}
		-\eta(1-\beta_1)P(\nu)^{-1} & (1-\beta_1)I_d \\
		(1-\beta_1)I_d &
		\tfrac{(1-\beta_1)(\kappa-\beta_1)}{\eta\beta_1}P(\nu)
	\end{pmatrix}.
	\label{eq:adamw-K}
\end{equation}
The inverse of the structure matrix is
\begin{equation}
	\Kmat_\nu^{-1}
	=
	\frac{1}{\kappa(1-\beta_1)}
	\begin{pmatrix}
		-\tfrac{\kappa-\beta_1}{\eta}P & \beta_1 I_d \\
		\beta_1 I_d & \eta\beta_1 P^{-1}
	\end{pmatrix}.
\end{equation}
As we have found $K_{mm}$, by substituting into the above equations, one can easily solve for $Q$:
\begin{equation}
	Q
	=
	\frac{1}{\kappa(1-\beta_1)}
	\begin{pmatrix}
		\zeta(\kappa-\beta_1)P & -\eta\beta_1\zeta I_d \\
		-\eta\beta_1\zeta I_d & -\eta\beta_1 P^{-1}
	\end{pmatrix}.
\end{equation}
Substituting this $Q$ into our ansatz gives the generating function for AdamW:
\begin{equation}
	\mathcal{L}(w,m)
	= L(w)
	- \tfrac{\eta\beta_1}{2(1-\beta_1)} \, m^\top P^{-1} m
	+ \tfrac{\zeta}{2(1-\beta_1)}
	\Bigl[ \|w\|_P^2 - \tfrac{\beta_1}{\kappa} \, \|w + \eta P^{-1} m\|_P^2 \Bigr],
	\label{eq:adamw-L}
\end{equation}
where $\|v\|_P^2 := v^\top P v$. Compared with Adam, the new terms in the AdamW generating function are all proportional to $\zeta$, so they vanish when $\zeta = 0$---as expected.

As we have found $\mathcal{L}$ and $K$, we can obtain the edge coupling for AdamW:
\begin{equation}
	\begin{aligned}
		\mathcal{B}^{\mathrm{AdamW}}_{\eta,\beta_1,\nu}(z,z')
		&= \mathcal{B}_{\eta,\beta_1,\nu}(z,z') \\
		&\quad + \tfrac{\zeta}{2\kappa(1-\beta_1)}
		\Bigl[\kappa\bigl(\|w\|_P^2 + \|w'\|_P^2\bigr)
		- 2\beta_1\,(w + \eta P^{-1}m)^\top P\,(w' + \eta P^{-1}m')\Bigr].
	\end{aligned}
	\label{eq:adamw-B}
\end{equation}
where $\mathcal{B}_{\eta,\beta_1,\nu}(z,z')$ denotes the edge coupling for Adam.

For the centered edge coupling, we have:
\begin{equation}
	\begin{aligned}
		\Upsilon^{\mathrm{AdamW}}_{\eta,\beta_1,\nu}(\zbar, \Delta)
		&:= \tfrac{1}{2}\mathcal{B}^{\mathrm{AdamW}}_{\eta,\beta_1,\nu}(\zbar + \Delta, \zbar - \Delta) \\
		&= \Upsilon_{\eta,\beta_1,\nu}(\zbar, \Delta)
		+ \frac{\zeta}{2\kappa(1-\beta_1)}
		\Bigl[\kappa\|\wbar\|_P^2 - \beta_1\|\wbar + \eta P^{-1}\mbar\|_P^2\Bigr] \\
		&\quad + \frac{\zeta}{2\kappa(1-\beta_1)}
		\Bigl[\kappa\|\delta\|_P^2 + \beta_1\|\delta + \eta P^{-1}\gamma\|_P^2\Bigr].
	\end{aligned}
	\label{eq:adamw-Ups}
\end{equation}
As we did in \cref{app:nesterov}, we will drop the superscript and subscripts for the remainder of this appendix---writing $\mathcal{B}$ and $\Upsilon$ for the edge couplings of AdamW.

\subsection{Rod Flow}
\label{app:adamw-rod}

As with Adam, we treat the second moment $\nubar$ as a smooth auxiliary variable. Applying the structure matrix to the gradient of the centered edge coupling, we obtain the rod flow for AdamW:
\begin{equation}
	\boxed{
		\begin{aligned}
		\ddt{}
		\begin{pmatrix}
			\bar{z} \\ \Delta
		\end{pmatrix}
		&=
		\begin{pmatrix}
			\Kmat_{\bar{\nu}} & 0 \\
			0 & -\Kmat_{\bar{\nu}}
		\end{pmatrix}
		\nabla_{(\bar{z},\Delta)}
		\Upsilon, \\
		\ddt{\nubar} &= (1-\beta_2)
		\bigl[\gbar^{\odot 2}+\gjump^{\odot 2}-\nubar\bigr].
		\end{aligned}}
\end{equation}
Writing out the individual ODEs:
\begin{equation}
	\begin{aligned}
		\ddt{\wbar} &= -\eta\zeta\wbar - \eta P^{-1}\bigl[\beta_1\mbar+(1-\beta_1)\gbar\bigr], \\
		\ddt{\mbar} &= (1-\beta_1)\bigl[\gbar-\mbar\bigr], \\
		\ddt{\nubar} &= (1-\beta_2)
		\bigl[\gbar^{\odot 2}+\gjump^{\odot 2}-\nubar\bigr], \\
		\ddt{\delta} &= -(1+\kappa)\delta+\eta P^{-1}\bigl[\beta_1\gamma+(1-\beta_1)\gjump\bigr], \\
		\ddt{\gamma} &= -(1+\beta_1)\gamma-(1-\beta_1)\gjump.
	\end{aligned}
	\label{eq:adamw-rodflow}
\end{equation}
For the endpoint rod flow, we have:
\begin{equation}
	\boxed{
		\begin{aligned}
		\ddt{}\begin{pmatrix} z \\ z' \end{pmatrix}
		&=
		\begin{pmatrix} 0 & \Kmat_{\nubar} \\ \Kmat_{\nubar} & 0 \end{pmatrix}
		\nabla_{(z, z')} \mathcal{B}, \\
		\ddt{\nubar} &= (1-\beta_2)\bigl[\tfrac12\bigl(\nabla L(w)^{\odot 2} + \nabla L(w')^{\odot 2}\bigr) - \nubar\bigr].
		\end{aligned}}
\end{equation}
Writing out the individual ODEs:
\begin{equation}
	\begin{aligned}
		\ddt{w} &= -\bigl(w - \bigl[\kappa w' - \eta P^{-1}\bigl(\beta_1 m' + (1-\beta_1)\nabla L(w')\bigr)\bigr]\bigr), \\
		\ddt{m} &= -\bigl(m - \bigl[\beta_1 m' + (1-\beta_1)\nabla L(w')\bigr]\bigr), \\
		\ddt{\nubar} &= (1-\beta_2)\bigl[\tfrac12\bigl(\nabla L(w)^{\odot 2} + \nabla L(w')^{\odot 2}\bigr) - \nubar\bigr], \\
		\ddt{w'} &= -\bigl(w' - \bigl[\kappa w - \eta P^{-1}\bigl(\beta_1 m + (1-\beta_1)\nabla L(w)\bigr)\bigr]\bigr), \\
		\ddt{m'} &= -\bigl(m' - \bigl[\beta_1 m + (1-\beta_1)\nabla L(w)\bigr]\bigr).
	\end{aligned}
	\label{eq:adamw-endpoint}
\end{equation}
Each endpoint regresses toward the AdamW image of the other endpoint.

\subsection{Central Flow}
\label{app:adamw-central}

We can also derive a central flow for AdamW. As with Adam, we first eliminate the momentum oscillation. For fixed $\delta$, the terms involving $\gamma$ can be written as:
\begin{equation}
	\begin{aligned}
		&\tfrac{2\beta_1}{1-\beta_1}\gamma\cdot\delta
		+\tfrac{\eta\beta_1}{2(1-\beta_1)}\gamma^\top P^{-1}\gamma
		+\tfrac{\zeta\beta_1}{2\kappa(1-\beta_1)}\bigl(2\eta\,\gamma\cdot\delta+\eta^2\gamma^\top P^{-1}\gamma\bigr) \\
		&\qquad=
		\tfrac{(1+\kappa)\beta_1}{\kappa(1-\beta_1)}\gamma\cdot\delta
		+\tfrac{\eta\beta_1}{2\kappa(1-\beta_1)}\gamma^\top P^{-1}\gamma \\
		&\qquad=
		\tfrac{\eta\beta_1}{2\kappa(1-\beta_1)}
		\left(\gamma+\frac{1+\kappa}{\eta}P\delta\right)^\top
		P^{-1}
		\left(\gamma+\frac{1+\kappa}{\eta}P\delta\right)
		-\tfrac{(1+\kappa)^2\beta_1}{2\eta\kappa(1-\beta_1)}\,\delta^\top P\delta .
	\end{aligned}
\end{equation}
One can see that the minimum is attained at $\gamma=-\tfrac{1+\kappa}{\eta}P\delta$. Eliminating $\gamma$ from the centered edge coupling gives the reduced edge coupling:
\begin{equation}
	\begin{aligned}
		\Upsilon^{\mathrm{red}}(\bar{z},\delta)
		&=
		\tfrac12\bigl[L(\wbar+\delta)+L(\wbar-\delta)\bigr]
		-\tfrac{\eta\beta_1}{2(1-\beta_1)}\mbar^\top P^{-1}\mbar \\
		&\quad
		+\tfrac{\zeta}{2(1-\beta_1)}\Bigl[\|\wbar\|_P^2-\tfrac{\beta_1}{\kappa}\|\wbar+\eta P^{-1}\mbar\|_P^2\Bigr]
		-\tfrac{(1+\kappa)(1+\beta_1)}{2\eta(1-\beta_1)}\,\delta^\top P\delta .
	\end{aligned}
\end{equation}
We again replace the loss with its local quadratic approximation about $\wbar$ and replace $\delta\otimes\delta$ with a generic positive semidefinite matrix $\Sigma$. The truncated reduced edge coupling is
\begin{equation}
	\widetilde{\Upsilon}^{\mathrm{red}}(\bar{z},\Sigma)
	=
	\mathcal{L}(\bar{z})
	-\tfrac12
	\langle\Lambda_{\mathrm{AdamW}},\Sigma\rangle,
\end{equation}
where $\mathcal{L}$ is the generating function for AdamW and the AdamW slack is given as:
\begin{equation}
	\Lambda_{\mathrm{AdamW}}(\bar{w},\bar{\nu})
	:=
	\tfrac{(1+\kappa)(1+\beta_1)}{\eta(1-\beta_1)}P(\bar{\nu})
	-H(\bar{w}).
\end{equation}
Compared with Adam, the weight decay lowers the threshold by the factor $1-\eta\zeta/2$, in agreement with the edge of stability condition. Maximizing over $\Sigma\succeq0$ gives the static conditions
\begin{equation}
	\Sigma\succeq0,
	\qquad
	\Lambda_{\mathrm{AdamW}}\succeq0,
	\qquad
	\langle\Lambda_{\mathrm{AdamW}},\Sigma\rangle=0.
\end{equation}
Let $U$ be an orthonormal basis for $\ker\Lambda_{\mathrm{AdamW}}$. We can express the covariance matrix as $\Sigma=UXU^\top$ with $X\succeq0$.

We then obtain the central flow ODEs for AdamW by taking the gradient of the truncated edge coupling with respect to the midpoint variables, treating the second moment as an auxiliary variable.
\begin{equation}
	\begin{aligned}
		\ddt{\bar{w}}
		&=
		-\eta\zeta\bar{w}
		-\eta P^{-1}
		\left[
		\beta_1\bar{m}
		+(1-\beta_1)\Bigl(
		\nabla L(\bar{w})
		+\tfrac12\nabla_{\bar{w}}
		\langle H(\bar{w}),\Sigma\rangle
		\Bigr)
		\right],
		\\
		\ddt{\bar{m}}
		&=
		(1-\beta_1)
		\left[
		\nabla L(\bar{w})
		+\tfrac12\nabla_{\bar{w}}
		\langle H(\bar{w}),\Sigma\rangle
		-\bar{m}
		\right],
		\\
		\ddt{\bar{\nu}}
		&=
		(1-\beta_2)
		\left[
		(\nabla L(\bar{w}))^{\odot2}
		+\operatorname{diag}
		\left(
		H(\bar{w})\Sigma H(\bar{w})
		\right)
		-\bar{\nu}
		\right].
	\end{aligned}
	\label{eq:adamw-central}
\end{equation}
As with Adam, the second-moment equation uses the approximation
\begin{equation}
	\gbar^{\odot2}+\gjump^{\odot2}
	\approx
	(\nabla L(\wbar))^{\odot2}
	+(H(\wbar)\delta)^{\odot2},
\end{equation}
which follows from the local quadratic approximation of the loss about $\wbar$ and by replacing $\delta\otimes\delta$ with $\Sigma$.

We determine the covariance matrix by imposing persistence of feasibility and complementarity. The time derivative of the slack is:
\begin{equation}
	\dot{\Lambda}_{\mathrm{AdamW}}
	=
	\tfrac{(1+\kappa)(1+\beta_1)}{\eta(1-\beta_1)}\nabla_{\bar{\nu}}P(\nubar)[\dot{\nubar}]
	- \nabla_{\bar{w}}H(\wbar)[\dot{\wbar}].
\end{equation}
At fixed $(\bar{w},\bar{\nu})$, both the $\dot{\bar{w}}$ and $\dot{\bar{\nu}}$ terms are affine in $\Sigma$. For our persistence conditions, we have:
\begin{equation}
	U^\top\dot{\Lambda}_{\mathrm{AdamW}}U\succeq0,
	\qquad
	\left\langle
	U^\top\dot{\Lambda}_{\mathrm{AdamW}}U,X
	\right\rangle=0.
\end{equation}
The above persistence conditions in combination with $X \succeq 0$ define a semidefinite complementarity problem for $X$. Altogether, we have that the following conditions define the central flow for AdamW:
\begin{equation}
	\boxed{
		\begin{aligned}
			\ddt{\bar{z}} &= \Kmat_{\bar{\nu}}\, \partial_{\bar{z}} \widetilde{\Upsilon}^{\mathrm{red}}(\bar{z}, \Sigma), \qquad \Sigma = U X U^\top, \\[4pt]
			\ddt{\bar{\nu}} &= (1-\beta_2)\Bigl[(\nabla L(\bar{w}))^{\odot2} + \operatorname{diag}\bigl(H(\bar{w})\Sigma H(\bar{w})\bigr) - \bar{\nu}\Bigr], \\[4pt]
			X &\succeq 0, \qquad U^\top \dot{\Lambda}_{\mathrm{AdamW}} U \succeq 0, \qquad \langle U^\top \dot{\Lambda}_{\mathrm{AdamW}} U, X \rangle = 0.
		\end{aligned}
	}
\end{equation}

%% file: appendix/E-rodflow.tex
\section{Rod Flow}
\label{app:rodflow}

This appendix collects the details of rod flow that the main text defers.

\subsection{Extent Matrix Formulation}
\label{app:sigma-lift}

The edge-coupling-derived rod flow of \cref{sec:rodflow-gda} evolves the vector $\delta$, whereas the original rod flow \citep{regis2026rod} evolves the extent matrix $\Sigma$. To more easily compare the two formulations, it helps to express the edge-coupling-derived rod flow in terms of $\Sigma$.

Let $\Sigma = \delta\otimes\delta$. Note that
\begin{equation}
	\operatorname{tr}\Sigma = \|\delta\|^2 .
\end{equation}
The vector $\delta$ can be identified with the principal eigenvector of $\Sigma$ scaled by the square root of its principal eigenvalue. Since the centered edge coupling is even in $\delta$, we can write the centered edge coupling as a function of $\Sigma$:
\begin{equation}
	\Psic(\wbar, \Sigma) = \tfrac{1}{2}\bigl[L(\wbar + \delta) + L(\wbar - \delta)\bigr] - \tfrac{1}{\eta}\operatorname{tr}\Sigma .
\end{equation}
We now want to derive the dynamics of $\Sigma$. As with $\delta$, $\Sigma$ must perform gradient ascent. However, when taking the derivative with respect to $\Sigma$, we need to be careful about which metric we use. While the Euclidean metric was appropriate for $\wbar$ and $\delta$, it is not the right choice for $\Sigma$. Instead, the relevant metric is the one induced on $\Sigma$ by the map $\delta \mapsto \delta \otimes \delta$. This turns out to be the \emph{Bures--Wasserstein metric} \citep{bhatia2019bures}, the same metric that gives the optimal transport distance between centered Gaussian distributions.

Consider a function $F(\Sigma)$, and let $\nabla_{\Sigma} F$ denote the Euclidean derivative of $F$. The Bures--Wasserstein gradient of $F$ is obtained from the Euclidean derivative $\nabla_\Sigma F$ via
\begin{equation}
 \operatorname{grad}_{\mathrm{BW}} F(\Sigma) = 2\bigl(\nabla_\Sigma F\,\Sigma + \Sigma\,\nabla_\Sigma F\bigr).
\end{equation}
Consider the Euclidean derivative $\nabla_\Sigma \Psi$. By using the chain rule, we can relate the Euclidean derivatives of $\Psi$ with respect to $\delta$ and $\Sigma$:
\begin{equation}
\nabla_\delta \Psi = 2	(\nabla_\Sigma \Psi)\,\delta
\end{equation}
We can then deduce:
\begin{equation}
\begin{aligned}
	\operatorname{grad}_{\mathrm{BW}} \Psi
	&= 2\bigl(\nabla_\Sigma\Psi\,\Sigma + \Sigma\,\nabla_\Sigma\Psi\bigr) \\
	&= 2\bigl(\nabla_\Sigma\Psi\,[\delta \otimes \delta] + [\delta \otimes \delta]\,\nabla_\Sigma\Psi\bigr) \\
	&= \nabla_\delta\Psi\otimes\delta + \delta\otimes\nabla_\delta\Psi \\
	&= (\gjump - \tfrac{2}{\eta} \delta)\otimes\delta + \delta\otimes (\gjump - \tfrac{2}{\eta} \delta) \\
	&= \gjump\otimes\delta + \delta\otimes\gjump - \tfrac{4}{\eta}\Sigma
\end{aligned}
\end{equation}
Setting $\ddt{\Sigma} = \eta\operatorname{grad}_{\mathrm{BW}}\Psi$, we then obtain the following rod flow ODE:
\begin{equation}
	\ddt{\Sigma} = \eta\bigl(\gjump\otimes\delta + \delta\otimes\gjump\bigr) - 4\Sigma .
	\label{eq:lift}
\end{equation}
The right-hand side can be expressed in terms of $\Sigma$ alone. By the fundamental theorem of calculus, the difference between the endpoint gradients can be expressed as integral with respect to the Hessian:
\begin{equation}
\nabla L(\bar{w} + \delta) - \nabla L(\bar{w} - \delta) = \int_{-1}^{1} \nabla^2 L(\wbar + s\delta)\,\delta ds
\end{equation}
Let $\widetilde{H}$ denote the rod-averaged Hessian.
\begin{equation}
\widetilde{H}(\bar{w}, \delta) = \tfrac{1}{2}\int_{-1}^{1} \nabla^2 L(\wbar + s\delta)\,\dd s
\end{equation}
We then have that the gradient half-difference can be expressed as:
\begin{equation}
\gjump = \widetilde{H} \delta
\end{equation}
Substituting, we then have that the rod flow ODE for $\Sigma$ can be expressed as:
\begin{equation}
\begin{aligned}
	\ddt{\Sigma}
	&= \eta\bigl(\widetilde{H}\Sigma + \Sigma\widetilde{H}\bigr) - 4\Sigma \\
	&= \eta\Bigl[\bigl(\widetilde{H} - \tfrac{2}{\eta}I_d\bigr)\Sigma + \Sigma\bigl(\widetilde{H} - \tfrac{2}{\eta}I_d\bigr)\Bigr].
	\label{eq:lift-lyap}
\end{aligned}
\end{equation}

\subsection{The Noise Floor}
\label{app:floor}

In the stable regime, the sharp component of the gradient \textit{should} decay exponentially under gradient descent. In practice, however, rounding noise prevents this decay: the sharp component of the discrete iterates instead plateaus at a small but nonzero value.

To motivate why that would be the case, consider the following model of the rounding error. Every time the optimizer stores its state, each stored component $z_i$ is rounded to a nearby representable number. Let $\mathsf{u}$ represent the unit round-off ($\mathsf{u} = 2^{-24}$ for float32). We can model the rounding error as uniform on the interval $[-\mathsf{u}|z_i|, \mathsf{u}|z_i|]$---and independent across coordinates and iterations. As the round-off noise is uniformly-distributed, it has a variance of $\mathsf{u}^2 z_i^2/3$.

Let $\hat{u}_j$ denote an eigenvector of the Hessian. Projected onto $\hat{u}_j$, the rounding of the stored position and momentum injects per-step variances of
\begin{equation}
	V_j = \frac{\mathsf{u}^2}{3}\sum_i (\hat{u}_j)_i^2\,\wbar_i^2,
	\qquad
	V^{\mathrm{mom}}_j = \frac{\mathsf{u}^2}{3}\sum_i (\hat{u}_j)_i^2\,\mbar_i^2
	\label{eq:roundoff-source}
\end{equation}
into that direction.

Why must rounding noise be the culprit? Could the plateau of the sharp component not arise from the dynamics of gradient descent itself?

\begin{figure}[tbp]
	\centering
	\includegraphics[width=\textwidth]{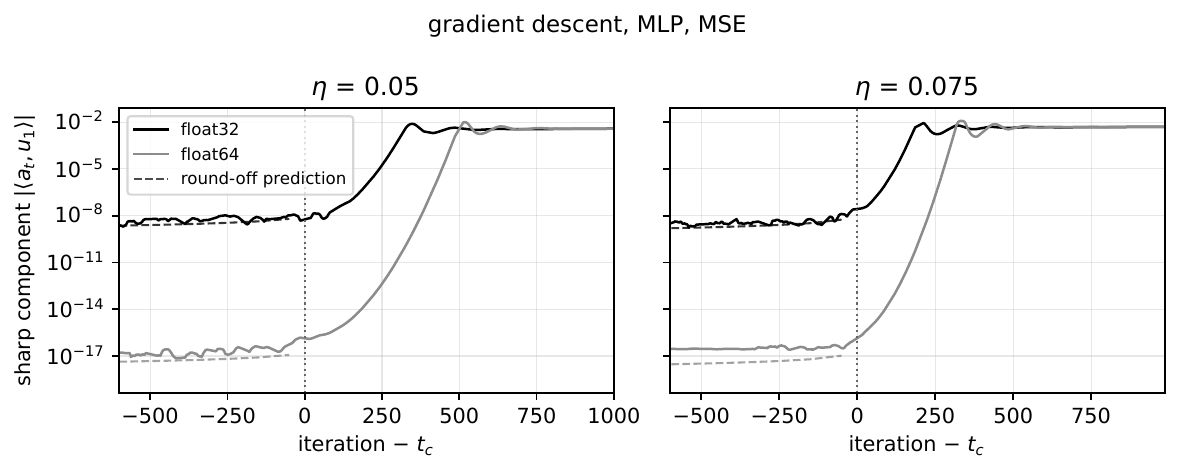}
	\caption{\textbf{Round-Off Sets the Floor.} Gradient descent on an MLP at $\eta = 0.05$ (\textbf{left}) and $\eta = 0.075$ (\textbf{right}), run in single and double precision from the same initialization. Shown is the sharp component $|\inner{a_t}{u_1}|$ of the oscillation amplitude. Before the crossing, the sharp component sits at the floor predicted by the rounding model, which is about eight orders of magnitude lower in double precision. As a result, the double-precision runs self-stabilize later than the single-precision runs, though eventually they settle on the same plateau.}
	\label{fig:roundoff}
\end{figure}

Two observations single out round-off. First, if the dynamics caused the plateau of the sharp component, we would expect the plateau to scale with the step size. Instead, the measured floor is the same for gradient descent across step sizes spanning several orders of magnitude. Second, repeating the gradient descent run in double precision from the same initialization lowers the floor by several orders of magnitude, to the level predicted for float64 (\cref{fig:roundoff}).

The model above predicts the noise injected at each step. However, to predict the \textit{stationary} covariance, we must also account for the dynamics of the optimizer.

Let $z_t = (w_t, m_t)$ denote the phase-space point of the optimizer, and let $\xi_t$ be the \textit{accumulated} deviation from the trajectory that the optimizer would have followed if there were no rounding error.

Since the deviation is of the order of the rounding error, a linearization suffices: we can replace the loss by its local quadratic approximation. The update map then becomes an affine map of $z$. Let $A$ denote the matrix of its linear part (\cref{tab:floor-maps}).

Let $C_t = \E[\xi_t\xi_t^\top]$ be the covariance of the deviation, taken over the rounding errors. Each step multiplies the deviation by $A$ and adds a fresh rounding error that is independent of the deviation:
\begin{equation}
	C_{t+1} = AC_t A^\top + \tfrac{\mathsf{u}^2}{3}\diag\bigl(\zbar^{\odot 2}\bigr).
	\label{eq:roundoff-lyap}
\end{equation}

\begin{table}[t]
	\centering
	\caption{\textbf{Linearized Update Maps.} The linear part $A$ of each optimizer's update under the quadratic approximation about the midpoint. The preconditioner $P$ is held fixed and bias correction is omitted.}
	\label{tab:floor-maps}
	\small
	\renewcommand{\arraystretch}{1.3}
	\begin{tabular}{@{}llc@{}}
		\toprule
		Optimizer & State & $A$ \\
		\midrule
		Gradient descent & $w$ & $I_d - \eta H$ \\
		\addlinespace[10pt]
		RMSProp & $w$ & $I_d - \eta P^{-1}H$ \\
		\addlinespace[10pt]
		Heavy ball & $(w, m)$ &
		$\begin{pmatrix} I_d - \eta(1-\beta)H & -\eta\beta I_d \\ (1-\beta)H & \beta I_d \end{pmatrix}$ \\
		\addlinespace[10pt]
		Nesterov & $(w, m)$ &
		$\begin{pmatrix} I_d - \eta(1-\beta)H & -\eta\beta\bigl(I_d - \eta(1-\beta)H\bigr) \\ (1-\beta)H & \beta\bigl(I_d - \eta(1-\beta)H\bigr) \end{pmatrix}$ \\
		\addlinespace[10pt]
		Adam & $(w, m)$ &
		$\begin{pmatrix} I_d - \eta(1-\beta_1)P^{-1}H & -\eta\beta_1 P^{-1} \\ (1-\beta_1)H & \beta_1 I_d \end{pmatrix}$ \\
		\addlinespace[10pt]
		AdamW & $(w, m)$ &
		$\begin{pmatrix} \kappa I_d - \eta(1-\beta_1)P^{-1}H & -\eta\beta_1 P^{-1} \\ (1-\beta_1)H & \beta_1 I_d \end{pmatrix}$ \\
		\bottomrule
	\end{tabular}
\end{table}

\begin{figure}[tbp]
	\centering
	\includegraphics[width=0.9\textwidth]{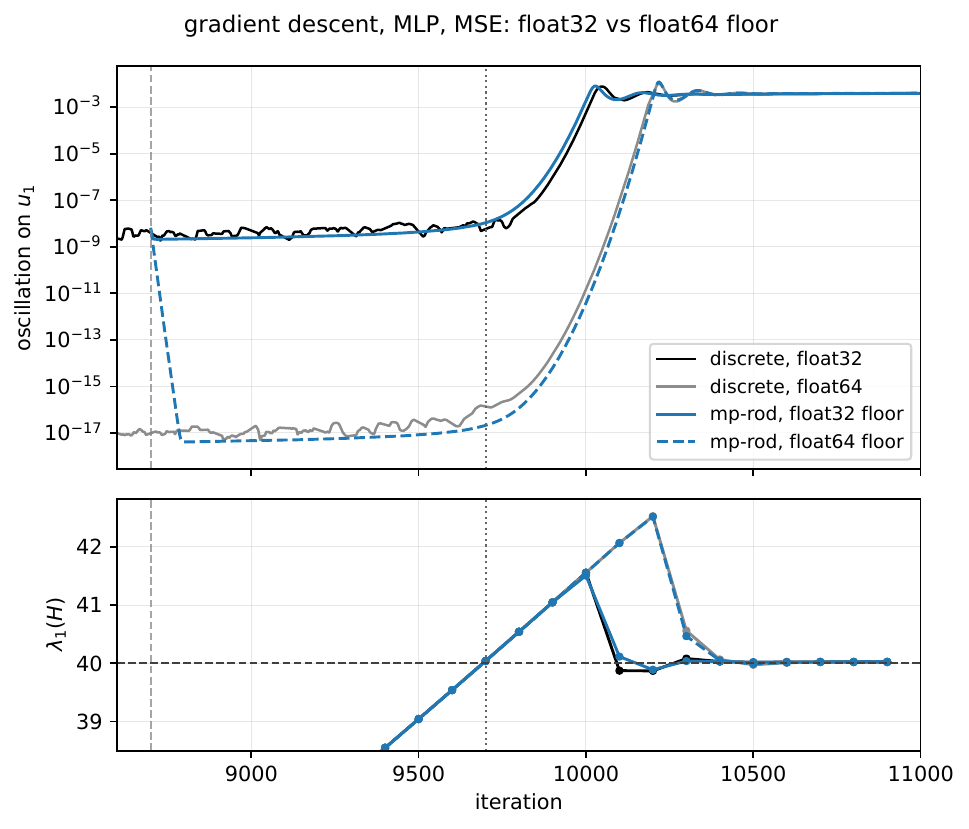}
	\caption{\textbf{A Floor for Each Precision.} Midpoint rod flow with the single- and double-precision rounding floors, compared against gradient descent run in each precision from the same initialization (MLP, $\eta = 0.05$). \textbf{Top:} The magnitude of the oscillation along the top eigenvector $\hat{u}_1$ over time. Gradient descent and rod flow agree closely at each precision. \textbf{Bottom:} The sharpness at the onset of self-stabilization. Self-stabilization lags behind the crossing of the sharpness threshold, and the lag is longer in double precision for both gradient descent and rod flow.}
	\label{fig:flow-floor}
\end{figure}

\begin{figure}[tbp]
	\centering
	\includegraphics[width=0.9\textwidth]{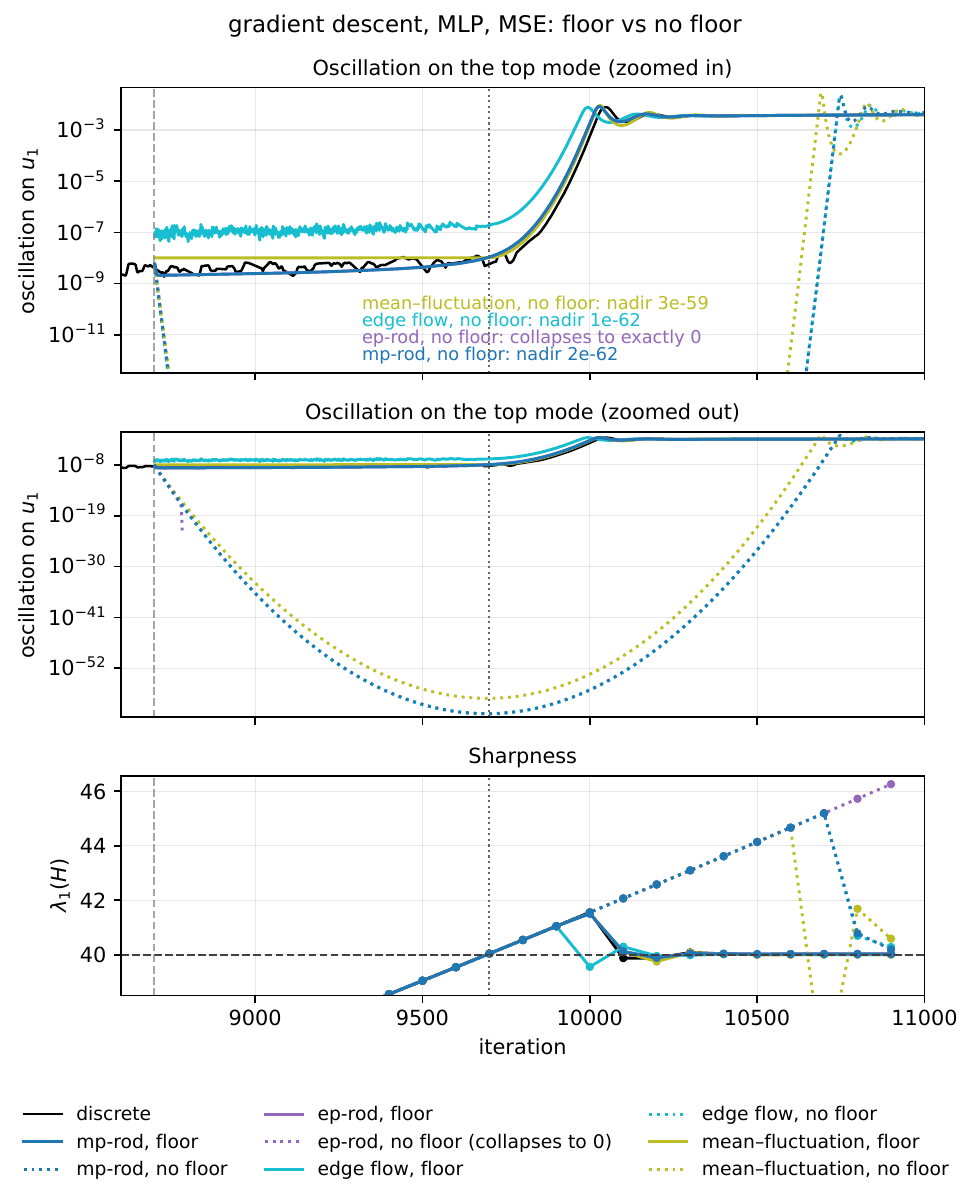}
	\caption{\textbf{Floor Versus No Floor.} Comparison of midpoint rod flow, endpoint rod flow, edge flow, and the mean--fluctuation model against gradient descent on an MLP (learning rate $\eta = 0.05$). Each model is shown with a noise floor (solid) and without it (dotted). The floor of the mean--fluctuation model is ours (\cref{app:exp-integration}). \textbf{Top and middle:} Oscillation along the top eigenvector $\hat{u}_1$, shown at two zoom levels. Without a floor, every flow's oscillation decays by more than fifty orders of magnitude before the crossing, and endpoint rod flow's collapses to exactly zero. \textbf{Bottom:} Without a floor, the flows arrest late. With it, every flow arrests at the same time as the gradient descent iterates.}
	\label{fig:floor}
\end{figure}

Rod flow's noise floor gives its extent the same stationary variance that round-off error induces in the discrete iterates.

Assume that the Hessian is fixed. Let $C^\ast$ denote the stationary covariance of the deviation. Along each eigencomponent of $H$, it satisfies the following equation:
\begin{equation}
	C_j^\ast = A_j C_j^\ast A_j^\top + \diag\bigl(V_j, V^{\mathrm{mom}}_j\bigr).
	\label{eq:floor-R}
\end{equation}
Let $R_j$ be the position sub-block of the stationary covariance:
\begin{equation}
R_j = (C^\ast_j)_{ww}
\end{equation}
$R_j$ is the stationary covariance along $\hat{u}_j$.

Let $\chi_j = \inner{\delta}{\hat{u}_j}$ be the component of the extent along $\hat{u}_j$. For simplicity, assume that $\gamma$ is in a fixed ratio to $\delta$. Then $\chi_j$ decays at the rate
\begin{equation}
	s_j = \frac{\det(I + A_j)}{1 + (A_j)_{mm}},
\end{equation}
so one forward Euler substep multiplies $\chi_j$ by
\begin{equation}
	\rho_j = 1 - s_j \, dt.
\end{equation}
After every substep, the noise floor increases the magnitude of each $\chi_j$ in quadrature while preserving its sign:
\begin{equation}
	\chi_j \leftarrow \sgn(\chi_j)\sqrt{\chi_j^2 + R_j\bigl(1 - \rho_j^2\bigr)}.
	\label{eq:floor}
\end{equation}

\subsection{Bias Correction}
\label{app:bias-correction}

Adam initializes its first and second moment estimates at zero, which biases them toward zero early in training. To correct for this, Adam introduces the bias-correction factors
\begin{equation}
	\mathrm{bc}_1 = 1 - \beta_1^{\,t+1},
	\qquad
	\mathrm{bc}_2 = 1 - \beta_2^{\,t+1}.
\end{equation}
These define the bias-corrected moments
\begin{align}
	\hat{m}_{t+1} &= \frac{m_{t+1}}{\mathrm{bc}_1}, \\
	\hat{\nu}_{t+1} &= \frac{\nu_{t+1}}{\mathrm{bc}_2}.
\end{align}
For both rod flow and central flow for Adam, the bias-corrected flows are obtained by replacing $m$ and $\nu$ with $\hat{m}$ and $\hat{\nu}$, respectively.

%% file: appendix/F-centralflow.tex
\section{Central Flow}
\label{app:centralflow}

This appendix collects the details of central flow that the main text defers.

\subsection{Convex Quadratic Programs}
\label{app:qp}

In the main text, we determined the covariance by defining a semidefinite complementarity problem (SDCP). Here, we show that the SDCP is equivalent to a quadratic program. This equivalence explains why the SDCP determines the covariance.

First, it helps to keep the following taxonomy in mind. We can categorize the objects in this appendix into six types, which form three dual pairs:
\begin{itemize}[leftmargin=*,itemsep=3pt,topsep=6pt]
	\item \textit{Displacements} are tangent vectors at $\bar{w}$. Examples include $\dot{\wbar}$ and the columns of $U$.
	\item \textit{Gradients} are cotangent vectors at $\bar{w}$ and are dual to displacements. Examples include $\nabla L$ and the momentum $\mbar$.
	\item \textit{Covariances} are symmetric $d\times d$ matrices built from pairs of displacements. Examples include the covariance matrix $\Sigma$ and the inverse preconditioner $P^{-1}$.
	\item \textit{Duals of covariances} are symmetric $d\times d$ matrices that act on pairs of displacements. Examples include the Hessian $H$, the preconditioner $P$, and the slack $\Lambda$.
	\item \textit{Critical covariances} are symmetric $k\times k$ matrices built from pairs of displacements along the critical directions. An example is the critical covariance matrix $X$.
	\item \textit{Duals of critical covariances} are symmetric $k\times k$ matrices that act on pairs of displacements along the critical directions. An example is the critical slack $U^\top\Lambda U$.
\end{itemize}

For simplicity, we will consider the central flow for gradient descent. The ODE is given as
$$ \frac{d\bar{w}}{dt} = - \eta \nabla L(\bar{w}) - \tfrac{\eta}{2} \nabla_{\bar{w}} \langle H(\bar{w}), \Sigma \rangle$$

It helps to introduce the notation $\mathcal{T}$ to denote the sharpness gradient as a linear map:
\begin{equation}
	\Tt[\Sigma] := \nabla_{\wbar}\inner{H(\wbar)}{\Sigma},
\end{equation}
$\mathcal{T}$ is a linear operator from the space of covariance matrices to the cotangent space at $\bar{w}$. Let $v$ be a tangent vector at $\bar{w}$. Now consider the inner product of $v$ with the sharpness gradient $\mathcal{T}[\Sigma]$:
\begin{equation}
	\langle \mathcal{T}[\Sigma], v \rangle.
\end{equation}
This inner product measures how much the curvature of $L$ in the directions of $\Sigma$ increases when we displace along $v$. We define the adjoint $\mathcal{T}^\ast$ as the linear operator from the tangent space at $\bar{w}$ to the space dual to covariance matrices that satisfies
\begin{equation}
	\langle \mathcal{T}^\ast[v], \Sigma \rangle = \langle \mathcal{T}[\Sigma], v \rangle
\end{equation}
for all tangent vectors $v$ and covariance matrices $\Sigma$. The adjoint can be identified with the directional derivative of the Hessian:
\begin{equation}
	\mathcal{T}^\ast[v] = \frac{\dd}{\dd s}H(\wbar + sv)\big|_{s=0}.
\end{equation}

Now consider the rate of change of the slack. Its critical block is affine in $X$:
\begin{equation}
	U^\top\dot{\Lambda}U = \eta(\mathsf{A} + \mathsf{B}[X]),
	\label{eq:affine-rate}
\end{equation}
where $\eta\mathsf{A}$ is the rate under the free flow ($X = 0$), and $\mathsf{B}$ is a linear operator on symmetric $k\times k$ matrices such that $\eta\mathsf{B}[X]$ gives the rate of change of the slack due to the oscillations:
\begin{equation}
	\mathsf{A} = \,U^\top\mathcal{T}^\ast[\nabla L]\,U,
	\qquad
	\mathsf{B}[X] = \tfrac{1}{2}\,U^\top\mathcal{T}^\ast\big[\mathcal{T}[UXU^\top]\big]\,U.
\end{equation}
Note that for any symmetric $k\times k$ matrices $X$ and $X'$, we have
\begin{equation}
	\begin{aligned}
		\langle \mathsf{B}[X], X' \rangle
		&= \tfrac{1}{2} \big\langle U^\top\mathcal{T}^\ast\big[\mathcal{T}[UXU^\top]\big]U, X' \big\rangle \\
		&= \tfrac{1}{2} \big\langle \mathcal{T}^\ast\big[\mathcal{T}[UXU^\top]\big], UX'U^\top \big\rangle \\
		&= \tfrac{1}{2} \big\langle \mathcal{T}[UXU^\top], \mathcal{T}[UX'U^\top] \big\rangle.
	\end{aligned}
\end{equation}
The final expression is symmetric in $X$ and $X'$, so $\mathsf{B}$ is self-adjoint. And setting $X' = X$ gives
\begin{equation}
 \langle \mathsf{B}[X], X \rangle = \tfrac{1}{2}\|\mathcal{T}[UXU^\top]\|^2 \geq 0,
\end{equation}
so $\mathsf{B}$ is positive semidefinite.

We let $\mathrm{SDCP}(\mathsf{A}, \mathsf{B})$ denote the SDCP defined in terms of $\mathsf{A}$ and $\mathsf{B}$:
\begin{equation}
	X\succeq0,
	\qquad
	\mathsf{A} + \mathsf{B}[X]\succeq0,
	\qquad
	\bigl\langle \mathsf{A} + \mathsf{B}[X],\ X\bigr\rangle = 0.
	\label{eq:sdcp}
\end{equation}
The above SDCP can be shown to be equivalent to a convex quadratic program. Consider the objective
\begin{equation}
	f(X) := \frac{1}{\eta^2}\Bigl\|\ddt{\wbar}\Bigr\|^2 = \Bigl\|\nabla L(\wbar) + \tfrac{1}{2}\Tt[UXU^\top]\Bigr\|^2.
	\label{eq:qp-gd}
\end{equation}
$f(X)$ is the squared speed of the midpoint $\bar{w}$ when the critical covariance matrix is $X$. Because the midpoint velocity is affine in $X$, $f$ is convex in $X$. Taking the gradient of $f$ with respect to $X$:
\begin{equation}
\begin{aligned}
\nabla_{X} f &= \nabla_{X} \Bigl\|\nabla L(\wbar) + \tfrac{1}{2}\Tt[UXU^\top]\Bigr\|^2 \\
&= \nabla_X \langle\nabla L(\wbar) + \tfrac{1}{2}\Tt[UXU^\top], \nabla L(\wbar) + \tfrac{1}{2}\Tt[UXU^\top] \rangle \\
&= \nabla_X \Bigl( \|\nabla L(\wbar)\|^2 + \langle \Tt[UXU^\top], \nabla L(\wbar) \rangle + \tfrac{1}{4}\|\Tt[UXU^\top]\|^2 \Bigr) \\
&= \nabla_X \Bigl( \langle U^\top \mathcal{T}^\ast[\nabla L(\wbar)]U, X \rangle + \tfrac{1}{4}\bigl\langle U^\top \mathcal{T}^\ast\bigl[\Tt[UXU^\top]\bigr]U, X \bigr\rangle \Bigr) \\
&= U^\top \mathcal{T}^\ast [\nabla L(\wbar) + \tfrac{1}{2}\Tt[UXU^\top]]U \\
&= \mathsf{A} + \mathsf{B}[X],
\end{aligned}
\end{equation}
which is the rate of change of the slack in the critical block, divided by $\eta$. We can also deduce that
\begin{equation}
f(X) = \|\nabla L(\wbar)\|^2 + \langle \mathsf{A}, X \rangle + \tfrac{1}{2}\langle \mathsf{B}[X], X \rangle.
\end{equation}
Consider the problem of minimizing $f$ over positive semidefinite matrices $X$:
\begin{equation}
	\min_{X\succeq0}\ f(X).
	\label{eq:qp}
\end{equation}
When there are no constraints, a convex function is minimized where its gradient is zero. However, the unconstrained minimizer might not lie in the PSD cone, so the gradient need not vanish at the constrained minimizer. Instead, the constrained minimizer must satisfy $\nabla_X f(X) \succeq 0$. Otherwise, one could increase the component of $X$ along an eigenvector of the gradient with a negative eigenvalue, which would decrease the objective while keeping $X$ in the PSD cone. Furthermore, along any eigenvector of the gradient with a positive eigenvalue, the component of $X$ must be zero. Otherwise, one could decrease that component to decrease the objective. This means that $\nabla_X f(X)$ and $X$ must be orthogonal.

Since $f$ is convex, these conditions are also sufficient for optimality. We have that minimizing $f$ subject to $X \succeq 0$ is equivalent to
\begin{equation}
	X \succeq 0, \quad \nabla_X f(X) \succeq 0, \quad \langle \nabla_X f(X), X \rangle = 0.
\end{equation}
Substituting our expression for $\nabla_X f$ recovers $\mathrm{SDCP}(\mathsf{A}, \mathsf{B})$. It's important to note that if the map $X \mapsto \mathcal{T}[UXU^\top]$ is injective, then $\mathsf{B}$ is positive definite---from which it follows that $f$ is a strongly convex quadratic. Because minimizing a strongly convex function over a closed convex set always yields a unique solution, $X$ is uniquely determined.

Interestingly, because the objective $f$ is the squared speed of the midpoint $\bar{w}$, there is an alternative interpretation: among all covariance matrices whose oscillations are contained within the critical subspace, the central flow selects the one for which the midpoint moves as slowly as possible.

The same structure recurs for the other optimizers (\cref{tab:qp-inputs}). In each case, the rate of change of the critical block of the slack is affine in $X$, and the persistence conditions give rise to $\mathrm{SDCP}(\mathsf{A}, \mathsf{B})$, with $\mathsf{A}$ and $\mathsf{B}$ specific to the optimizer. Whenever $\mathsf{B}$ is self-adjoint and positive semidefinite, the SDCP gives the optimality conditions of a convex quadratic program---which has a unique minimizer when $\mathsf{B}\succ0$.

\begin{table}[t]
	\centering
	\caption{\textbf{Inputs to the Quadratic Program.} The rate of change of the critical block of the slack, $U^\top\dot{\Lambda}U = \eta(\mathsf{A} + \mathsf{B}[X])$, for each optimizer. $\eta\mathsf{A}$ is the rate of change under the free flow. $\eta\mathsf{B}[X]$ is the rate of change due to the oscillations. For RMSProp, Adam, and AdamW, $\Lambda$ denotes that optimizer's slack.}
	\label{tab:qp-inputs}
	\small
	\renewcommand{\arraystretch}{1.3}
	\begin{tabular}{@{}lll@{}}
		\toprule
		Optimizer & $\mathsf{A}$ & $\mathsf{B}[X]$ \\
		\midrule
		Gradient descent & $U^\top\Tt^*[\nabla L]\,U$ & $\tfrac{1}{2}\,U^\top\Tt^*\Tt[UXU^\top]\,U$ \\
		\addlinespace[6pt]
		Heavy ball & $U^\top\Tt^*\bigl[\beta\mbar + (1-\beta)\nabla L\bigr]U$ & $\tfrac{1-\beta}{2}\,U^\top\Tt^*\Tt[UXU^\top]\,U$ \\
		\addlinespace[6pt]
		Nesterov & $U^\top\Tt^*\bigl[\beta^2\mbar + (1-\beta^2)\nabla L(\thetabar)\bigr]U$ & $\tfrac{1-\beta^2}{2}\,U^\top\Tt^*\Tt[UXU^\top]\,U$ \\
		\addlinespace[6pt]
		RMSProp & $U^\top\Tt^*\bigl[P^{-1}\nabla L\bigr]U$ & $\tfrac{1}{2}\,U^\top\Tt^*\bigl[P^{-1}\Tt[UXU^\top]\bigr]U$ \\
		& ${}+\tfrac{1-\beta_2}{\eta}\,U^\top\nabla_{\nubar}\Lambda\bigl[(\nabla L)^{\odot2} - \nubar\bigr]U$ & ${}+\tfrac{1-\beta_2}{\eta}\,U^\top\nabla_{\nubar}\Lambda\bigl[\diag(HUXU^\top H)\bigr]U$ \\
		\addlinespace[6pt]
		Adam & $U^\top\Tt^*\bigl[P^{-1}\bigl(\beta_1\mbar + (1-\beta_1)\nabla L\bigr)\bigr]U$ & $\tfrac{1-\beta_1}{2}\,U^\top\Tt^*\bigl[P^{-1}\Tt[UXU^\top]\bigr]U$ \\
		& ${}+\tfrac{1-\beta_2}{\eta}\,U^\top\nabla_{\nubar}\Lambda\bigl[(\nabla L)^{\odot2} - \nubar\bigr]U$ & ${}+\tfrac{1-\beta_2}{\eta}\,U^\top\nabla_{\nubar}\Lambda\bigl[\diag(HUXU^\top H)\bigr]U$ \\
		\addlinespace[6pt]
		AdamW & $U^\top\Tt^*\bigl[\zeta\wbar + P^{-1}\bigl(\beta_1\mbar + (1-\beta_1)\nabla L\bigr)\bigr]U$ & $\tfrac{1-\beta_1}{2}\,U^\top\Tt^*\bigl[P^{-1}\Tt[UXU^\top]\bigr]U$ \\
		& ${}+\tfrac{1-\beta_2}{\eta}\,U^\top\nabla_{\nubar}\Lambda\bigl[(\nabla L)^{\odot2} - \nubar\bigr]U$ & ${}+\tfrac{1-\beta_2}{\eta}\,U^\top\nabla_{\nubar}\Lambda\bigl[\diag(HUXU^\top H)\bigr]U$ \\
		\bottomrule
	\end{tabular}
\end{table}

In terms of computational implementation, we found that we were able to use the same SDCP solver as \citet{Cohen+25CentralFlow} with minimal modification.

\subsection{Central Flows as Projected Dynamical Systems}
\label{app:projected}

For gradient descent, central flow can be interpreted as a projected gradient flow. We extend this interpretation to the other optimizers, where it becomes a constrained flow.

Consider the central flow for gradient descent. Define the objective $\Fval$ as the supremum over $\Sigma\succeq0$ of the truncated edge coupling:
\begin{equation}
	\Fval(\wbar) := \sup_{\Sigma\succeq0}\widetilde{\Psi}(\wbar, \Sigma) = L(\wbar) + \ind_{\feas}(\wbar),
\end{equation}
where $\feas = \{\wbar : \Lambda(\wbar)\succeq0\}$ is the feasible region and $\ind_{\feas}$ is its indicator function:
\begin{equation}
	\ind_{\feas}(\wbar) =
	\begin{cases}
		0 & \text{if } \wbar \in \feas, \\
		+\infty & \text{otherwise.}
	\end{cases}
\end{equation}
The indicator $\ind_{\feas}(\wbar)$ creates an infinite potential barrier that constrains the flow to the feasible region. The central flow is a gradient flow on $\Fval$, though only in the weak sense of a subdifferential inclusion \citep{brezis1973operateurs}:
\begin{equation}
	\ddt{\wbar} \in -\eta\,\partial\Fval(\wbar) = -\eta\bigl(\nabla L(\wbar) + N_{\feas}(\wbar)\bigr),
	\label{eq:value-flow}
\end{equation}
where $N_{\feas}(\wbar)$ is the normal cone to $\feas$ at $\wbar$, whose elements act as normal forces constraining the flow to the feasible region. Among the velocities that the inclusion allows, central flow selects the one of least norm (\cref{app:qp}). It is the \emph{slow solution} of the inclusion \citep{henry1973existence, cornet1983existence}.

We can interpret the other central flows similarly. For each optimizer, we define an objective by taking the supremum over $\Sigma\succeq0$ of the truncated reduced edge coupling:
\begin{equation}
	\Fval(\bar{z}) := \sup_{\Sigma\succeq0}\widetilde{\Upsilon}^{\mathrm{red}}(\bar{z}, \Sigma) = \mathcal{L}(\bar{z}) + \ind_{\feas}(\bar{w}),
	\label{eq:value-functions}
\end{equation}
with $\mathcal{L}$ and $\feas$ given in \cref{tab:value-functions}. The central flow then satisfies the differential inclusion
\begin{equation}
	\ddt{\bar{z}} \in \Kmat\,\partial\Fval(\bar{z}) = \Kmat\left(\nabla\mathcal{L}(\bar{z}) + \begin{pmatrix} N_{\feas}(\bar{z}) \\ 0 \end{pmatrix}\right),
	\label{eq:kflow-inclusion}
\end{equation}
where $\Kmat$ is the structure matrix of the optimizer.

\begin{table}[t]
	\centering
	\caption{\textbf{Central Flow Objectives.} For each optimizer, we give the generating function $\mathcal{L}$ and the feasible region $\feas$ of the objective. For the preconditioned methods, we give the metric $W_{\nu}$ through which the normal cone acts on the second moment.}
	\label{tab:value-functions}
	\small
	\renewcommand{\arraystretch}{1.3}
	\begin{tabular}{@{}llll@{}}
		\toprule
		Optimizer & $\mathcal{L}$ & $\feas$ & $W_{\nu}$ \\
		\midrule
		Gradient descent & $L(\wbar)$ & $\{\Lambda(\wbar)\succeq0\}$ & --- \\
		\addlinespace[6pt]
		Heavy ball & $L(\wbar) - \tfrac{\eta\beta}{2(1-\beta)}\|\mbar\|^2$ & $\{\Lambda_{\mathrm{HB}}(\wbar)\succeq0\}$ & --- \\
		\addlinespace[6pt]
		Nesterov & $L(\thetabar) - \tfrac{\eta\beta^2}{2(1-\beta)}\|\mbar\|^2$ & $\{\Lambda_{\mathrm{NAG}}(\thetabar)\succeq0\}$ & --- \\
		\addlinespace[6pt]
		RMSProp & $L(\wbar)$ & $\{\Lambda_{\mathrm{RMSProp}}(\wbar, \nubar)\succeq0\}$ & $\tfrac{8(1-\beta_2)}{\eta}\,P^2\operatorname{Diag}\bigl(\sqrt{\nubar}\bigr)$ \\
		\addlinespace[6pt]
		Adam & $L(\wbar) - \tfrac{\eta\beta_1}{2(1-\beta_1)}\mbar^\top P^{-1}\mbar$ & $\{\Lambda_{\mathrm{Adam}}(\wbar, \nubar)\succeq0\}$ & $\tfrac{8(1-\beta_2)(1+\beta_1)}{\eta(1-\beta_1)}\,P^2\operatorname{Diag}\bigl(\sqrt{\nubar}\bigr)$ \\
		\addlinespace[6pt]
		AdamW & $L(\wbar) - \tfrac{\eta\beta_1}{2(1-\beta_1)}\mbar^\top P^{-1}\mbar$ & $\{\Lambda_{\mathrm{AdamW}}(\wbar, \nubar)\succeq0\}$ & $\tfrac{4(1-\beta_2)(1+\kappa)(1+\beta_1)}{\eta(1-\beta_1)}\,P^2\operatorname{Diag}\bigl(\sqrt{\nubar}\bigr)$ \\
		& ${}+ \tfrac{\zeta}{2(1-\beta_1)}\bigl[\|\wbar\|_P^2$ & & \\
		& $\qquad{}- \tfrac{\beta_1}{\kappa}\|\wbar + \eta P^{-1}\mbar\|_P^2\bigr]$ & & \\
		\bottomrule
	\end{tabular}
\end{table}

For the preconditioned methods, the constraint depends on the joint state $(\wbar, \nubar)$. There are then two ways to leave the feasible region: the sharpness can increase as $\wbar$ moves, or $\nubar$ can shrink. The covariance acts to counteract both effects. As a result, each normal force in $N_{\feas}(\bar{z}, \nubar)$ is a $2d$-dimensional vector, with components for both $\wbar$ and $\nubar$.

Just as we defined the $\wbar$ component of the normal force using the subdifferential of the truncated edge coupling with respect to $\wbar$, we define the $\nubar$ component using the subdifferential with respect to $\nubar$. The $\nubar$ component of the normal force, which we denote $N_{\feas}^{\nubar}$, is then
\begin{equation}
	\begin{aligned}
		N_{\feas}^{\nubar}
		&= \partial_{\nubar}\widetilde{\Psi}_{\eta,\nubar} \\
		&= -\partial_{\nubar}\tfrac{1}{2}\langle \Lambda_{\mathrm{RMSProp}}, \Sigma \rangle \\
		&= -\frac{1}{2\eta}\operatorname{Diag}(\nubar)^{-1/2}\diag(\Sigma).
	\end{aligned}
	\label{eq:nu-normal}
\end{equation}
The central flow ODE for $\nubar$ is
\begin{equation}
	\ddt{\nubar} = (1-\beta_2)\bigl[(\nabla L)^{\odot2} + \diag(H\Sigma H) - \nubar\bigr].
\end{equation}
Recall that the columns of $U$ lie in the critical subspace. Since these columns have preconditioned sharpness equal to $2/\eta$, we have $HU = \tfrac{2}{\eta}PU$. Because $P$ is diagonal, it follows that
\begin{equation}
	\diag(H\Sigma H) = \tfrac{4}{\eta^2}P^2\diag(\Sigma).
\end{equation}
Substituting into the ODE for $\nubar$ and comparing with \eqref{eq:nu-normal}, we deduce that
\begin{equation}
	\ddt{\nubar} = (1-\beta_2)\bigl[(\nabla L)^{\odot2} - \nubar\bigr] - W_{\nu}\,\partial_{\nubar}\widetilde{\Psi}_{\eta,\nubar},
	\label{eq:nu-metric}
\end{equation}
where the matrix
\begin{equation}
	W_{\nu} = \tfrac{8(1-\beta_2)}{\eta}\,P^2\operatorname{Diag}(\nubar)^{1/2}
\end{equation}
is effectively an inverse metric tensor that converts the normal force into its effect on the second moment. The central flow of RMSProp is therefore a \textit{projected dynamical system} \citep{dupuis1993dynamical, nagurney1996projected}, though not a projected gradient flow, because the unconstrained dynamics of $\nubar$ are not a gradient flow.

%% file: appendix/G-experimental-details.tex
\section{Experimental Details}
\label{app:experimental-details}

This appendix provides further details about our experimental setup. Code to reproduce our experiments is included as ancillary files with the arXiv version of this paper.

\subsection{Stable Flows}
\label{app:exp-stable}

The \textit{stable flow} of an optimizer is its na\"{\i}ve continuous-time limit, obtained by replacing the discrete difference $z_{t+1} - z_t$ in the update equations with the time derivative $\dot{z}$. For a structured gradient method, the stable flow is $\dot{z} = \Kmat\nabla\mathcal{L}(z)$.

Below the sharpness threshold, the stable flow closely tracks the discrete iterates. However, while the iterates self-stabilize such that their sharpness hovers at the threshold, the sharpness of the stable flow keeps rising past the threshold. We therefore use the stable flow as the baseline against which to compare rod flow and central flow. The stable flows of the six optimizers are as follows.

\paragraph{Gradient descent.}
\begin{equation}
	\begin{aligned}
		\ddt{w} &= -\eta\nabla L(w).
	\end{aligned}
\end{equation}

\paragraph{RMSProp.}
\begin{equation}
	\begin{aligned}
		\ddt{w} &= -\eta P^{-1}\nabla L(w), \\
		\ddt{\nu} &= (1-\beta_2)\bigl[\nabla L(w)^{\odot 2} - \nu\bigr].
	\end{aligned}
\end{equation}

\paragraph{Heavy ball.}
\begin{equation}
	\begin{aligned}
		\ddt{w} &= -\eta\bigl[\beta m + (1-\beta)\nabla L(w)\bigr], \\
		\ddt{m} &= (1-\beta)\bigl[\nabla L(w) - m\bigr].
	\end{aligned}
\end{equation}

\paragraph{Nesterov momentum.}
\begin{equation}
	\begin{aligned}
		\ddt{\theta} &= -\eta\bigl[\beta^2 m + (1-\beta^2)\nabla L(\theta)\bigr], \\
		\ddt{m} &= (1-\beta)\bigl[\nabla L(\theta) - m\bigr].
	\end{aligned}
\end{equation}

\paragraph{Adam.}
\begin{equation}
	\begin{aligned}
		\ddt{w} &= -\eta \hat{P}^{-1}\bigl[\beta_1 \hat{m} + (1-\beta_1)\nabla\widehat{L}(w)\bigr], \\
		\ddt{m} &= (1-\beta_1)\bigl[\nabla L(w) - m\bigr], \\
		\ddt{\nu} &= (1-\beta_2)\bigl[\nabla L(w)^{\odot 2} - \nu\bigr].
	\end{aligned}
\end{equation}
\paragraph{AdamW.}
\begin{equation}
	\begin{aligned}
		\ddt{w} &= -\eta\zeta w - \eta \hat{P}^{-1}\bigl[\beta_1 \hat{m} + (1-\beta_1)\nabla\widehat{L}(w)\bigr], \\
		\ddt{m} &= (1-\beta_1)\bigl[\nabla L(w) - m\bigr], \\
		\ddt{\nu} &= (1-\beta_2)\bigl[\nabla L(w)^{\odot 2} - \nu\bigr].
	\end{aligned}
\end{equation}
Note that the stable flows for Adam and AdamW are bias-corrected (\cref{app:bias-correction}).

\subsection{Runs}
\label{app:exp-runs}

We ran each of the six optimizers on three architectures (MLP, CNN, and ViT). The step size, handoff iteration, and crossing iteration for each run are given in \Cref{tab:runs}.

\paragraph{Data and models.} We used the first $5\,000$ images of the CIFAR-10 training set, normalized per channel, with no data augmentation. All runs on a given architecture start from the same initialization: PyTorch's default, with a fixed seed.

The MLP flattens each input image into a vector of $3\,072$ values. It has two hidden layers of width $50$, each consisting of a fully connected layer followed by a SiLU activation, and a linear output layer that maps to the ten outputs. In total, the MLP has $156\,710$ parameters.

The CNN operates directly on the image, represented as three channels of $32\times32$ pixels. It has two convolutional layers, each with $32$ channels and $3\times3$ kernels, and each padded by one pixel so that the convolution preserves the image size. Each convolution is followed by a $\tanh$ activation and $2\times2$ average pooling, which halves the image size from $32\times32$ to $16\times16$, then to $8\times8$. A final linear layer maps the resulting $32\times8\times8 = 2\,048$ features to the ten outputs. In total, the CNN has $30\,634$ parameters.

The ViT is a SimpleViT, as implemented in the \texttt{vit-pytorch} library. It splits each image into an $8\times8$ grid of $4\times4$ patches, giving $64$ tokens. Each patch, a vector of $48$ values, passes through a LayerNorm, a linear embedding of width $64$, and a second LayerNorm. The ViT then adds fixed two-dimensional sinusoidal position embeddings. It does not use a class token. The tokens pass through three transformer blocks. Each block applies multi-head self-attention, with $8$ heads of width $8$ and bias-free projections, followed by an MLP of width $256$ with GELU activations. Both sublayers are preceded by a LayerNorm and wrapped in a residual connection. After a final LayerNorm, the ViT mean-pools the tokens, and a linear head maps the result to the ten outputs. In total, the ViT has $153\,322$ parameters. To lower the sharpness at initialization, we halve the initial weights of the head.

We also replace each LayerNorm with a reimplementation in elementary PyTorch operations, so that the loss can be differentiated three times, as central flow requires.

\paragraph{Training.} We train full-batch with the MSE loss. We chose MSE because it's the loss function which induces sustained edge of stability dynamics. We also ran experiments with cross-entropy loss on gradient descent and RMSProp (\cref{app:xent}).

\begin{table}[t]
	\centering
	\caption{\textbf{Runs.} The step size $\eta$, handoff $t_h$, and crossing $t_c$ of each run.}
	\label{tab:runs}
	\small
	\renewcommand{\arraystretch}{1.2}
	\setlength{\tabcolsep}{4pt}
	\begin{tabular}{@{}lrrrrrrrrr@{}}
		\toprule
		& \multicolumn{3}{c}{MLP} & \multicolumn{3}{c}{CNN} & \multicolumn{3}{c}{ViT} \\
		\cmidrule(lr){2-4} \cmidrule(lr){5-7} \cmidrule(l){8-10}
		Optimizer & $\eta$ & $t_h$ & $t_c$ & $\eta$ & $t_h$ & $t_c$ & $\eta$ & $t_h$ & $t_c$ \\
		\midrule
		Gradient descent & $0.05$ & $8\,700$ & $9\,700$ & $0.05$ & $19\,600$ & $20\,600$ & $0.1$ & $4\,600$ & $5\,100$ \\
		RMSProp & $2\times10^{-6}$ & $1\,500$ & $2\,500$ & $2\times10^{-6}$ & $3\,700$ & $4\,800$ & $2\times10^{-6}$ & $3\,700$ & $4\,200$ \\
		Heavy ball & $0.3$ & $3\,800$ & $4\,800$ & $0.7$ & $1\,600$ & $2\,800$ & $0.2$ & $5\,700$ & $6\,200$ \\
		Nesterov & $0.2$ & $3\,100$ & $4\,100$ & $0.4$ & $1\,200$ & $2\,200$ & $0.2$ & $3\,700$ & $4\,300$ \\
		Adam & $10^{-5}$ & $3\,900$ & $4\,900$ & $5\times10^{-5}$ & $1\,000$ & --- & $10^{-5}$ & $4\,000$ & $4\,500$ \\
		AdamW & $10^{-5}$ & $6\,200$ & $7\,200$ & $5\times10^{-5}$ & $1\,800$ & $2\,700$ & $10^{-5}$ & $5\,600$ & $6\,100$ \\
		\bottomrule
	\end{tabular}
\end{table}

\paragraph{Hyperparameters.} All four momentum methods use $\beta = \beta_1 = 0.9$. Adam and AdamW use $\beta_2 = 0.999$. These are the standard hyperparameter settings used in practice. Adam and AdamW also use $\epsilon = 10^{-7}$ and bias correction. RMSProp uses the PyTorch defaults: $\beta_2 = 0.99$ and $\epsilon = 10^{-8}$, without bias correction. For AdamW, we chose a weight decay $\zeta$ large enough to noticeably delay the crossing into the edge of stability relative to Adam.

\paragraph{Step sizes.} We chose the step sizes using probe runs of the discrete optimizer, so that the crossing occurs neither too early nor too late. In practice, this meant choosing a step size such that the edge of stability was reached within a couple thousand iterations.

\paragraph{Crossing and handoff.} The crossing $t_c$ is the first iteration at which the sharpness at the discrete midpoint exceeds the threshold. Since the sharpness is computed every $100$ iterations, $t_c$ is a multiple of $100$. We placed the handoff $t_h$ using the crossing in a probe run. As a result, the handoff comes roughly $1\,000$ iterations before the crossing ($500$ for the ViT). On the CNN, Adam's sharpness plateaus just below the threshold and never crosses it. For this run, the handoff instead comes $1\,000$ iterations before the sharpness reaches its plateau.

\paragraph{Handoff state.} At the handoff, each flow is initialized from the last two discrete iterates. Its center is set to their midpoint. For rod flow, the extent is initialized from the half-differences of the last two iterates' positions and momenta. These half-differences are dominated by the drift of the center rather than by the oscillation, so the initial amplitude of the extent corresponds to the entire discrete step, not just the oscillating portion. However, this transient quickly fades---well before the edge of stability is reached.

\subsection{Measurements}
\label{app:exp-measurements}

\begin{table}[t]
	\centering
	\caption{\textbf{Effective Sharpness and Thresholds.} The effective sharpness of each optimizer and its threshold $\thr$.}
	\label{tab:thresholds}
	\small
	\renewcommand{\arraystretch}{1.3}
	\begin{tabular}{@{}lcc@{}}
		\toprule
		Optimizer & Effective sharpness & Threshold $\thr$ \\
		\midrule
		Gradient descent & $\lamone(H)$ & $\dfrac{2}{\eta}$ \\
		\addlinespace[4pt]
		RMSProp & $\lamone(P^{-1/2}HP^{-1/2})$ & $\dfrac{2}{\eta}$ \\
		\addlinespace[4pt]
		Heavy ball & $\lamone(H)$ & $\dfrac{2(1+\beta)}{\eta(1-\beta)}$ \\
		\addlinespace[4pt]
		Nesterov & $\lamone(H)$ & $\dfrac{2(1+\beta)}{\eta(1-\beta)(1+2\beta)}$ \\
		\addlinespace[4pt]
		Adam & $\lamone(\hat{P}^{-1/2}H\hat{P}^{-1/2})$ & $\dfrac{2(1+\beta_1)}{\eta(1-\beta_1)}$ \\
		\addlinespace[4pt]
		AdamW & $\lamone(\hat{P}^{-1/2}H\hat{P}^{-1/2})$ & $\Bigl(1 - \dfrac{\eta\zeta}{2}\Bigr)\dfrac{2(1+\beta_1)}{\eta(1-\beta_1)}$ \\
		\bottomrule
	\end{tabular}
\end{table}

\paragraph{Sharpness.} Every $100$ iterations, we compute the top six eigenvalues of the Hessian at the center of each trajectory. We use LOBPCG on Hessian-vector products, warm-started from the eigenvectors of the previous call. For RMSProp, Adam, and AdamW, we instead compute the eigenvalues of the pencil $(H, P)$, which are those of $P^{-1/2}HP^{-1/2}$ (with $\hat{P}$ in place of $P$ for Adam and AdamW). Each trajectory uses its own preconditioner: the iterates use the discrete optimizer's, and each flow uses the one given by its second moment $\nubar$. \Cref{tab:thresholds} gives the resulting effective sharpness of each optimizer, along with its threshold.

\paragraph{Oscillation envelope.} The half-displacement $\dhat_t$ of consecutive iterates contains not only the oscillation but also the drift of the midpoint. For gradient descent, writing $\gjump_t$ and $\gbar_t$ for the half-difference and the average of the gradients at $w_t$ and $w_{t+1}$, we have
\begin{equation*}
	\dhat_t = \tfrac{\eta}{2}\gjump_t + \tfrac{\eta}{2}\gbar_t,
	\qquad
	\tfrac{\eta}{2}\gbar_t = -\tfrac12(\wbar_{t+1} - \wbar_t).
\end{equation*}
Before the crossing, the drift dominates the half-displacement, so the onset of the oscillation cannot be seen in it. We therefore remove the drift and undo the alternating sign. For every optimizer, we define the oscillation envelope of the iterates as
\begin{equation}
	a_t := (-1)^t\bigl[\dhat_t + \tfrac12(\wbar_{t+1} - \wbar_t)\bigr].
	\label{eq:decomp}
\end{equation}
The figures compare the norm of $a_t$ with the norm of the extent $\norm{\delta}$ of the rod flows.

\paragraph{Distance.} We measure how closely a flow tracks the iterates by the Euclidean distance from its center to the discrete midpoint $\wbar_t$. For the momentum methods, this distance compares positions only. We likewise evaluate the loss at each center.

\subsection{Numerical Integration}
\label{app:exp-integration}

\paragraph{Time stepping.} The flows run in lockstep with the discrete optimizer: after each optimizer iteration, every flow takes four forward Euler substeps of size $\dd t = 0.25$, so one unit of flow time corresponds to one iteration.

\paragraph{Storing the extent.} Midpoint rod flow does not store its extent as a vector. Without the noise floor, the extent decays by up to several hundred e-folds between the handoff and the crossing, approaching the range of double precision, which is about $700$ e-folds. A stored vector could then underflow to zero and lose its direction.

We instead store a unit axis $\hat{v}$ and a log-amplitude $\log\norm{\Delta}$. Each substep applies forward Euler to the extent, $\Delta \leftarrow \norm{\Delta}(\hat{v} + \dd t\,f)$, where $f$ is the velocity of the extent per unit amplitude. The axis $\hat{v} + \dd t\,f$ is then renormalized, and the logarithm of its norm is added to the log-amplitude. By contrast, endpoint rod flow stores the endpoints themselves.

\paragraph{Noise floor.} Both rod flows include the noise floor of \cref{app:floor} along the three sharpest directions. The floor models the rounding error in the discrete optimizer's single-precision state.

\paragraph{Central flow.} Central flow solves its semidefinite complementarity problem at every substep. To find the critical directions, it recomputes the top eigenvectors of the (preconditioned) Hessian at each substep, warm-started from the previous substep. The eigenvectors whose eigenvalues lie within $2.5\%$ of the threshold are passed to the problem, so that it can catch an eigenvalue arriving at the threshold. With one such eigenvector, the problem is solved in closed form. With more, it is solved using the convex solver of \citet{Cohen+25CentralFlow}, at a relative accuracy of about $10^{-4}$.

\paragraph{Edge flow.} For gradient descent, we also ran edge flow \citep{marion2026edge}, which tracks a center $\wbar$, a unit axis $u$, and an amplitude $x$. These evolve as
\begin{equation}
	\begin{aligned}
		\dot{\wbar} &= -\tfrac{\eta}{2}\bigl[\nabla L(\wbar + xu) + \nabla L(\wbar - xu)\bigr], \\
		\dot{x}     &= \bigl(\eta\norm{H(\wbar)\,u} - 2\bigr)\,x, \\
		u           &\leftarrow H(\wbar)\,u \,/\, \norm{H(\wbar)\,u}.
	\end{aligned}
	\label{eq:edge-flow}
\end{equation}
The center follows the negative average of the gradients at the endpoints $\wbar \pm xu$, and the amplitude grows when the sharpness estimate $\norm{H(\wbar)\,u}$ exceeds $2/\eta$. The axis takes one step of power iteration at each substep. At the handoff, we set $u$ to the top eigenvector of the Hessian and $x = |\inner{\dhat}{u}|$. Following \citet{marion2026edge}, we kept the floor $x \geq |\inner{\nabla L(\wbar)}{u}|$.

\paragraph{Mean--fluctuation model.} We implemented the mean--fluctuation model from the first version of \citet{chodron2026free}. The second version keeps the same equations but computes the covariance differently. The model tracks a center $\wbar$ and a covariance $\Sigma$, which evolve as
\begin{equation}
	\begin{aligned}
		\dot{\wbar}  &= -\eta\bigl[\nabla L(\wbar) + \tfrac12\nabla_{\wbar}\inner{H(\wbar)}{\Sigma}\bigr], \\
		\dot{\Sigma} &= -\eta\bigl(H(\wbar)\,\Sigma + \Sigma\,H(\wbar)\bigr) + \eta^2\,H(\wbar)\,\Sigma\,H(\wbar).
	\end{aligned}
	\label{eq:mf-model}
\end{equation}
The center moves as in central flow, but $\Sigma$ follows the linearized gradient descent map, with no threshold. As in the first version, we restricted $\Sigma$ to the span of the top three Hessian eigenvectors (two for the ViT), which we recomputed every $2.5$ iterations rather than transported. At the handoff, we set $\Sigma$ to the projection of $\dhat \otimes \dhat$ onto this span. Neither version of the model has a noise floor, so we added one: we held the standard deviation of $\Sigma$ along each of these eigenvectors at $10^{-8}$ or above, about the size of the rounding noise along $\hat{u}_1$ before the crossing (\cref{fig:roundoff}).

\begin{figure}[tbp]
	\centering
	\includegraphics[width=0.8\textwidth]{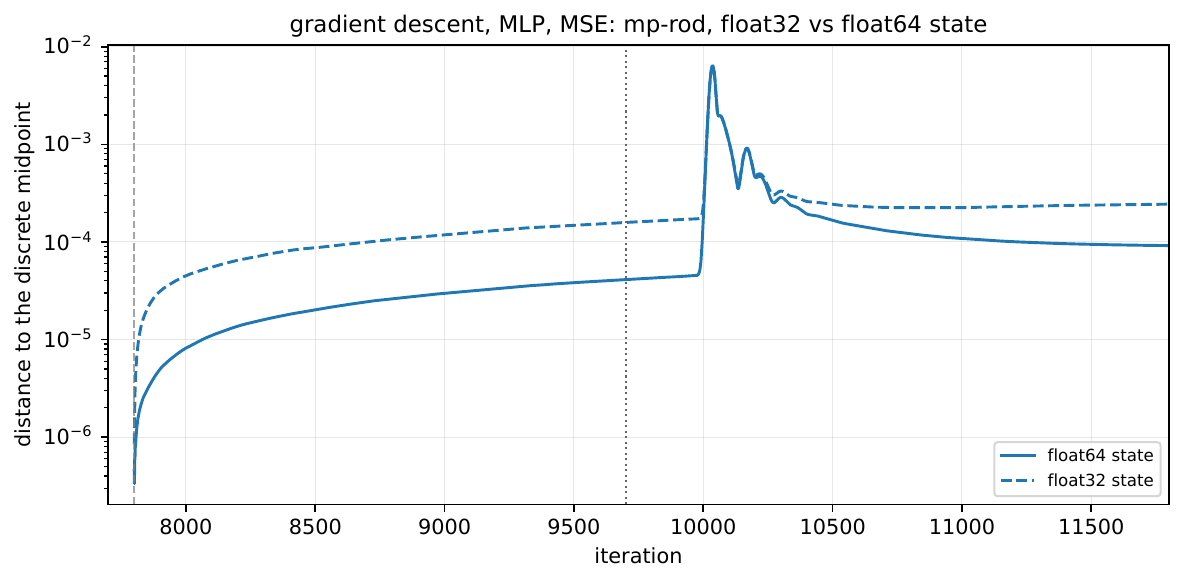}
	\caption{\textbf{Double-Precision Flow State.} Distance between the center of midpoint rod flow and the discrete midpoint, with the flow's state stored in single and double precision (gradient descent, MLP, $\eta = 0.05$). With single precision, the distance grows as rounding error accumulates, reaching several times that of the double-precision flow.}
	\label{fig:flow-dtype}
\end{figure}

\paragraph{Precision.} The discrete optimizers run in single precision. Every flow keeps its state in double precision and casts it to single precision only to evaluate the loss and its derivatives. The flows thus compute their gradients with the same arithmetic as the iterates, but rounding error does not accumulate in their state. Endpoint rod flow needs double precision for a second reason: a single-precision pair of endpoints cannot resolve the kicks of the noise floor.

On the CNN, the discrete runs are not bitwise reproducible. We kept PyTorch's default cuDNN settings, which allow the convolutions to use TF32 arithmetic and nondeterministic algorithms. This does not affect the comparisons, since the flows run in lockstep with the iterates.

%% file: appendix/H-more-experiments.tex
\clearpage
\section{Additional Experiments}
\label{app:more-experiments}

In this appendix, we present the results of our experiments, including additional experiments not featured in the main text.

\subsection{Flow Comparisons}
\label{app:flow-comparisons}

\begin{figure}[!htbp]
	\centering
	\textbf{\large Gradient Descent}\par\vspace{0.2cm}
	\includegraphics[width=\textwidth]{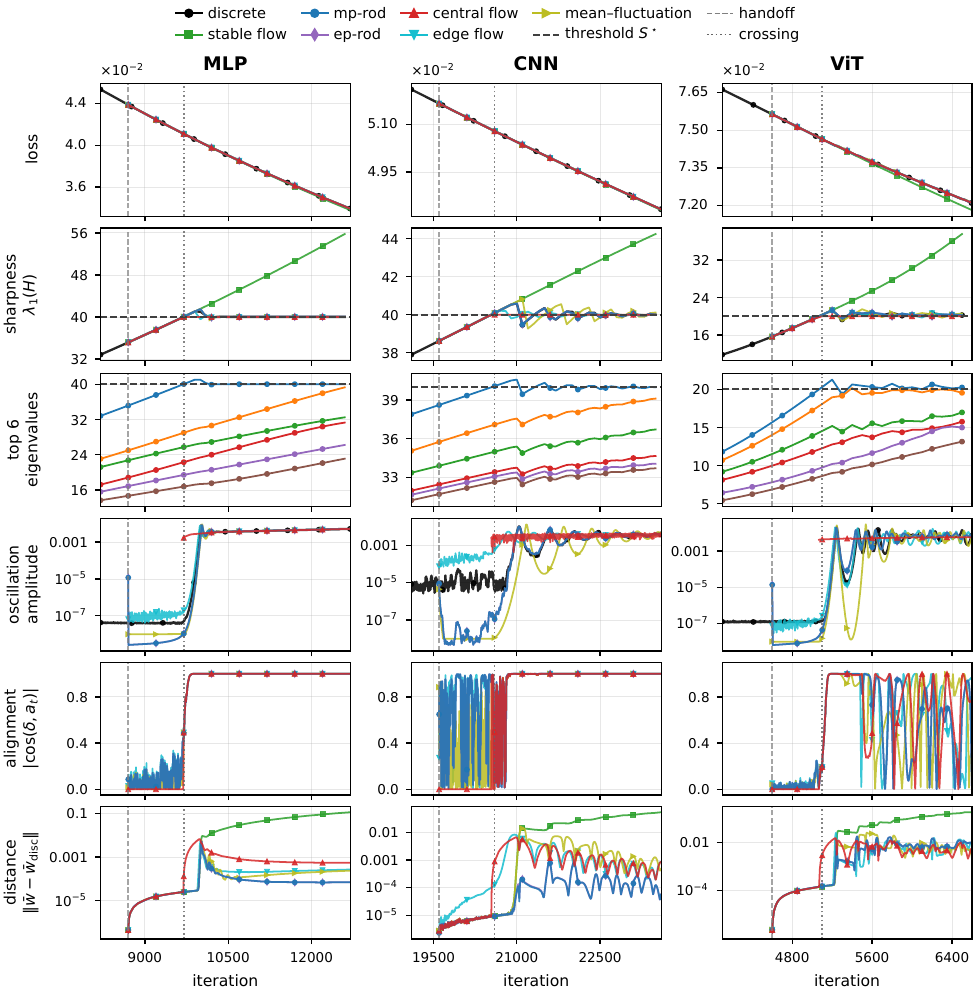}
	\caption{\textbf{Flow Comparison for Gradient Descent.} \textbf{MLP:} $\eta = 0.05$. \textbf{CNN:} $\eta = 0.05$. \textbf{ViT:} $\eta = 0.1$. The rows show, from top to bottom: the loss at each center; the sharpness at each center; the top six eigenvalues at the discrete optimizer's midpoint; the oscillation amplitude; the alignment of each flow's oscillation with the discrete-time oscillation; and the distance from each flow's center to the discrete optimizer's midpoint. Dashed vertical lines mark the handoff, and dotted vertical lines mark the crossing into the edge of stability.}
	\label{fig:flows-gd}
\end{figure}

\begin{figure}[!htbp]
	\centering
	\textbf{\large RMSProp}\par\vspace{0.2cm}
	\includegraphics[width=\textwidth]{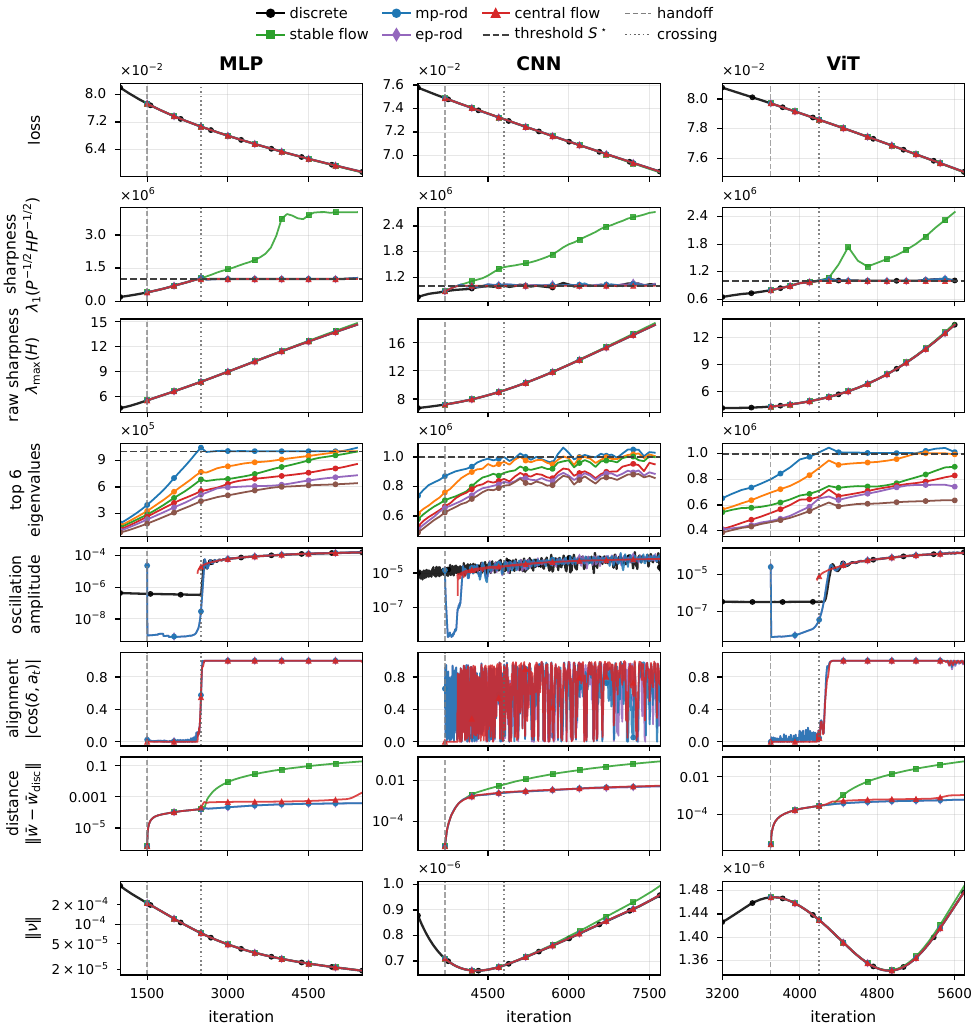}
	\caption{\textbf{Flow Comparison for RMSProp} (no bias correction). \textbf{MLP:} $\eta = 2\times10^{-6}$. \textbf{CNN:} $\eta = 2\times10^{-6}$. \textbf{ViT:} $\eta = 2\times10^{-6}$. All with $\beta_2 = 0.99$ and $\epsilon = 10^{-8}$. The rows show, from top to bottom: the loss at each center; the preconditioned sharpness at each center; the raw sharpness at each center; the top six preconditioned eigenvalues at the discrete optimizer's midpoint; the oscillation amplitude; the alignment of each flow's oscillation with the discrete-time oscillation; the distance from each flow's center to the discrete optimizer's midpoint; and the norm of the second-moment estimate. Dashed vertical lines mark the handoff, and dotted vertical lines mark the crossing into the edge of stability.}
	\label{fig:flows-rmsprop}
\end{figure}

\begin{figure}[!htbp]
	\centering
	\textbf{\large Heavy Ball}\par\vspace{0.2cm}
	\includegraphics[width=\textwidth]{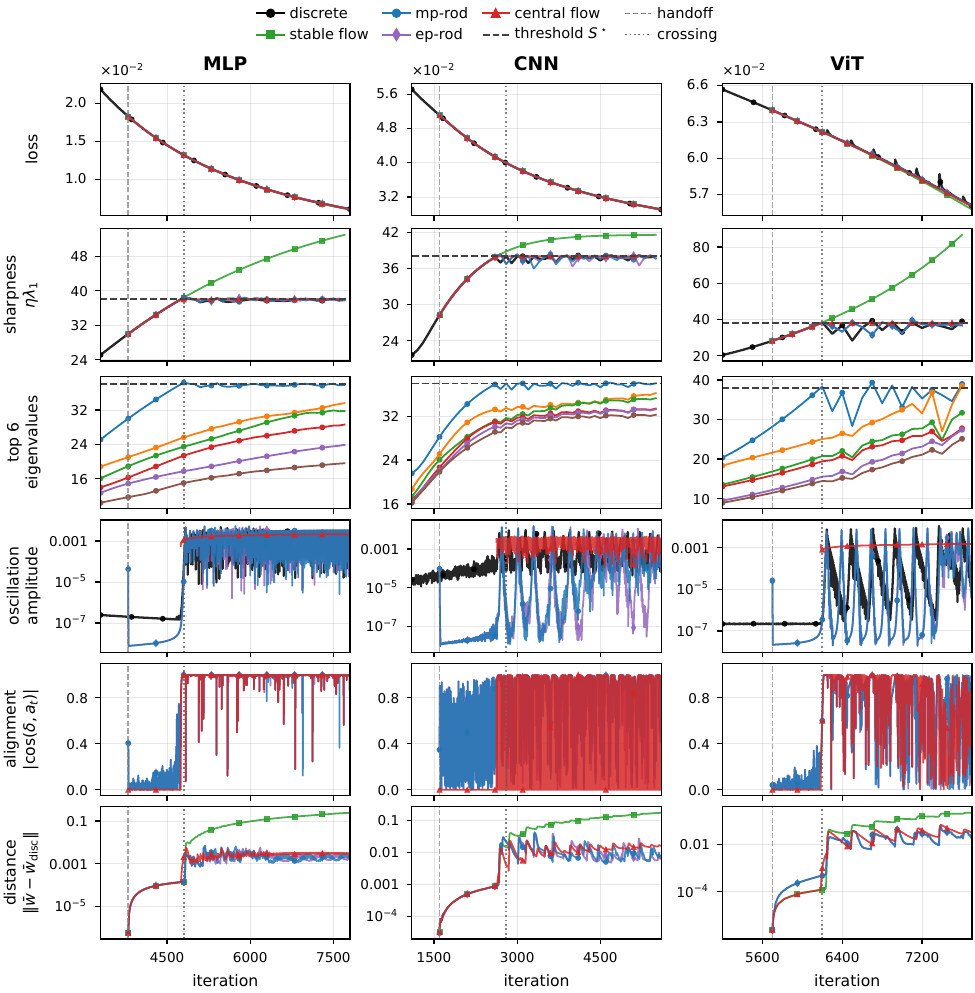}
	\caption{\textbf{Flow Comparison for Heavy Ball Momentum.} \textbf{MLP:} $\eta = 0.3$. \textbf{CNN:} $\eta = 0.7$. \textbf{ViT:} $\eta = 0.2$. All with $\beta = 0.9$. The rows show, from top to bottom: the loss at each center; the sharpness at each center; the top six eigenvalues at the discrete optimizer's midpoint; the oscillation amplitude; the alignment of each flow's oscillation with the discrete-time oscillation; and the distance from each flow's center to the discrete optimizer's midpoint. Dashed vertical lines mark the handoff, and dotted vertical lines mark the crossing into the edge of stability.}
	\label{fig:flows-hb}
\end{figure}

\begin{figure}[!htbp]
	\centering
	\textbf{\large Nesterov}\par\vspace{0.2cm}
	\includegraphics[width=\textwidth]{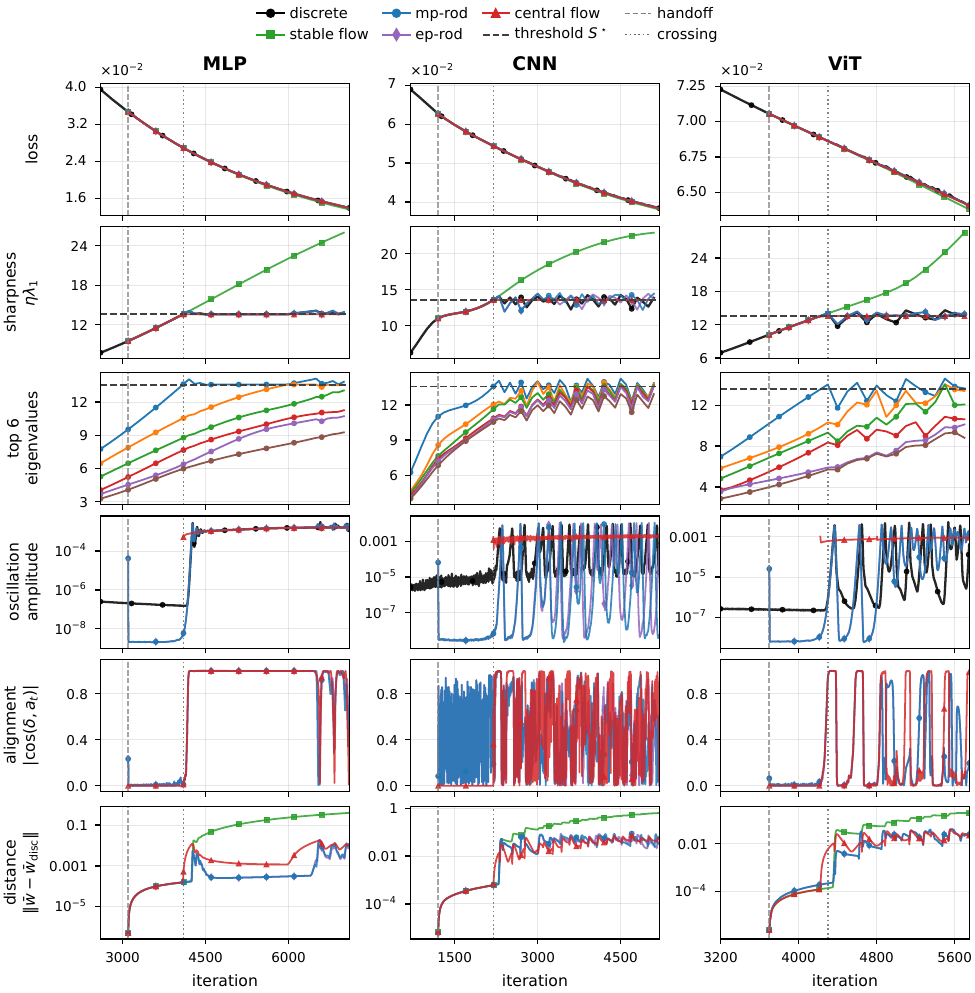}
	\caption{\textbf{Flow Comparison for Nesterov Momentum.} \textbf{MLP:} $\eta = 0.2$. \textbf{CNN:} $\eta = 0.4$. \textbf{ViT:} $\eta = 0.2$. All with $\beta = 0.9$. The rows show, from top to bottom: the loss at each center; the sharpness at each center; the top six eigenvalues at the discrete optimizer's midpoint; the oscillation amplitude; the alignment of each flow's oscillation with the discrete-time oscillation; and the distance from each flow's center to the discrete optimizer's midpoint. Dashed vertical lines mark the handoff, and dotted vertical lines mark the crossing into the edge of stability.}
	\label{fig:flows-nesterov}
\end{figure}

\begin{figure}[!htbp]
	\centering
	\textbf{\large Adam}\par\vspace{0.2cm}
	\includegraphics[width=\textwidth]{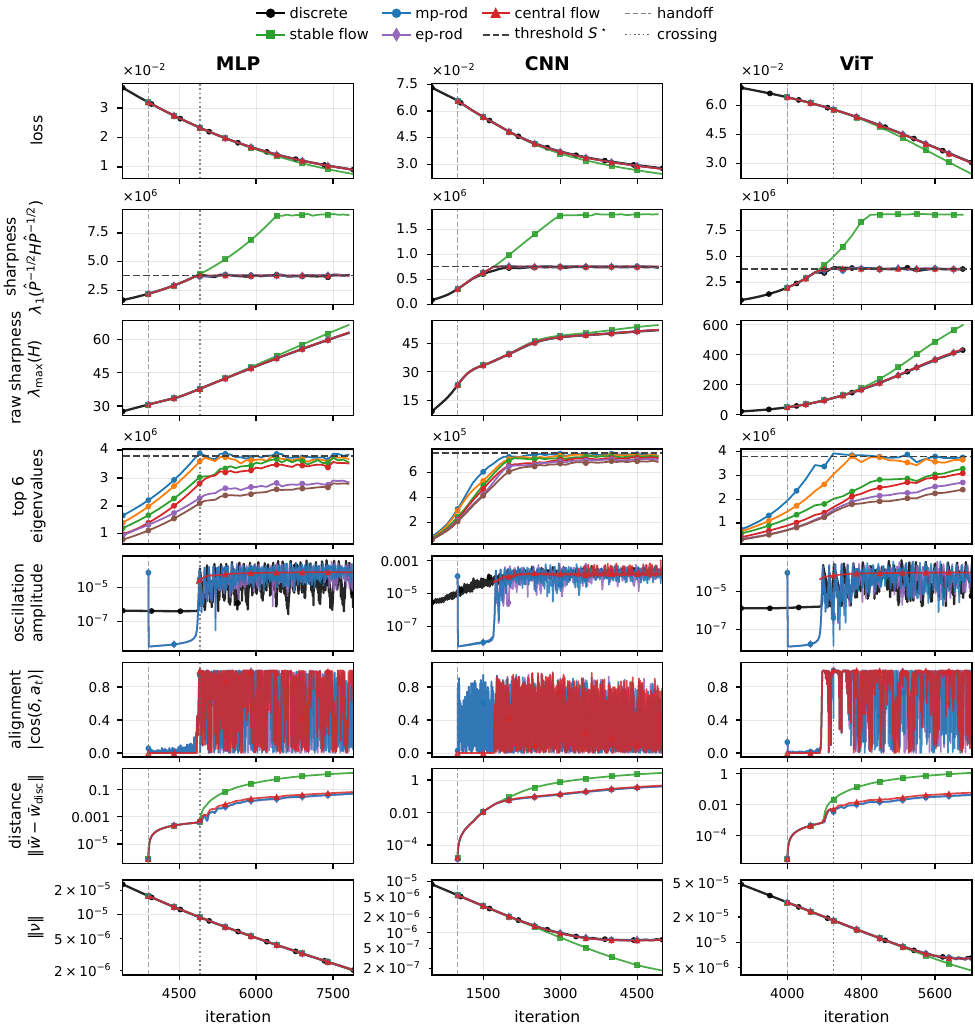}
	\caption{\textbf{Flow Comparison for Adam} (with bias correction). \textbf{MLP:} $\eta = 10^{-5}$. \textbf{CNN:} $\eta = 5\times10^{-5}$. \textbf{ViT:} $\eta = 10^{-5}$. All with $\beta_1 = 0.9$, $\beta_2 = 0.999$, and $\epsilon = 10^{-7}$. The rows show, from top to bottom: the loss at each center; the preconditioned sharpness at each center; the raw sharpness at each center; the top six preconditioned eigenvalues at the discrete optimizer's midpoint; the oscillation amplitude; the alignment of each flow's oscillation with the discrete-time oscillation; the distance from each flow's center to the discrete optimizer's midpoint; and the norm of the second-moment estimate. Dashed vertical lines mark the handoff, and dotted vertical lines mark the crossing into the edge of stability (none on the CNN, which never crosses).}
	\label{fig:flows-adam}
\end{figure}

\begin{figure}[!htbp]
	\centering
	\textbf{\large AdamW}\par\vspace{0.2cm}
	\includegraphics[width=\textwidth]{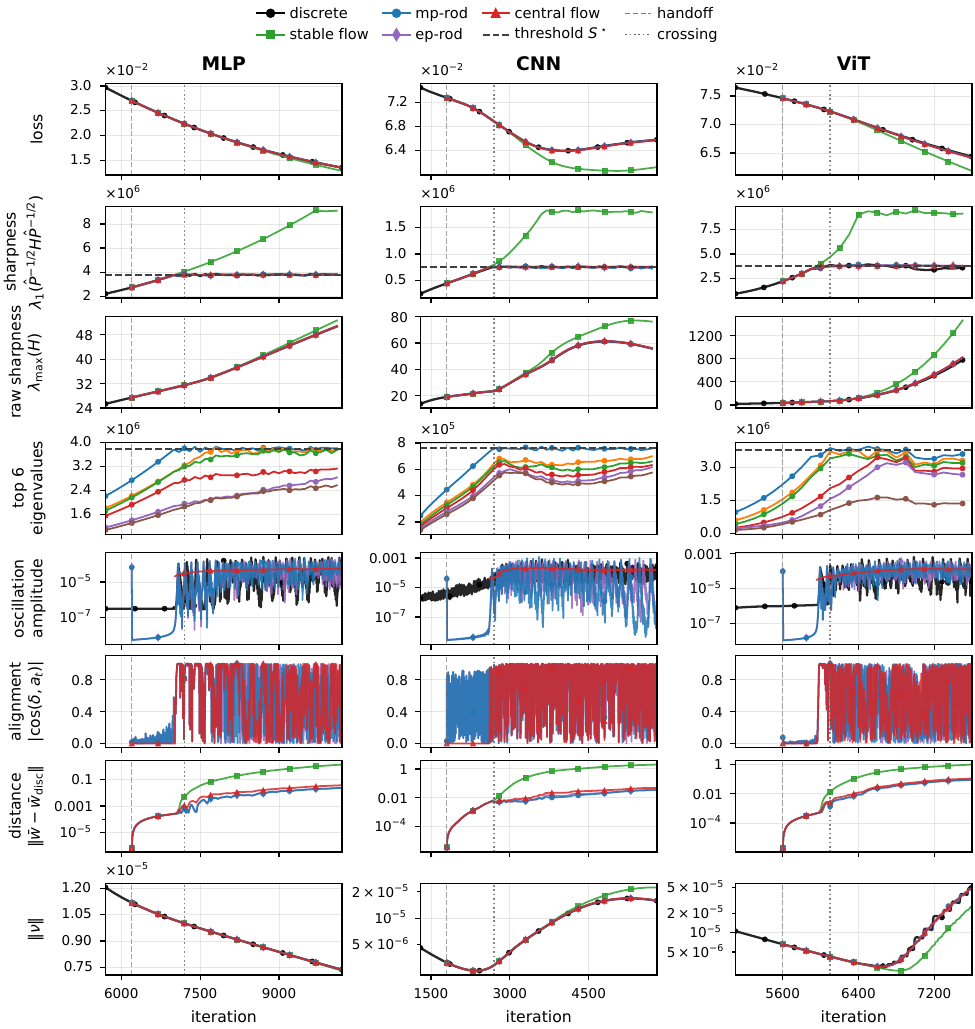}
	\caption{\textbf{Flow Comparison for AdamW} (with bias correction). \textbf{MLP:} $\eta = 10^{-5}$, $\zeta = 10$. \textbf{CNN:} $\eta = 5\times10^{-5}$, $\zeta = 40$. \textbf{ViT:} $\eta = 10^{-5}$, $\zeta = 10$. All with $\beta_1 = 0.9$, $\beta_2 = 0.999$, and $\epsilon = 10^{-7}$. The rows show, from top to bottom: the loss at each center; the preconditioned sharpness at each center; the raw sharpness at each center; the top six preconditioned eigenvalues at the discrete optimizer's midpoint; the oscillation amplitude; the alignment of each flow's oscillation with the discrete-time oscillation; the distance from each flow's center to the discrete optimizer's midpoint; and the norm of the second-moment estimate. Dashed vertical lines mark the handoff, and dotted vertical lines mark the crossing into the edge of stability.}
	\label{fig:flows-adamw}
\end{figure}

\Cref{fig:flows-gd,fig:flows-rmsprop,fig:flows-hb,fig:flows-nesterov,fig:flows-adam,fig:flows-adamw} compare the flows with the discrete iterates. Each experiment starts with a warmup phase in which the discrete iterates progressively sharpen. About $1\,000$ iterations before the edge of stability ($500$ for the ViT), the flows are initialized from the discrete iterates. The iterates and the flows then advance in lockstep for $4\,000$ iterations ($2\, 000$ for the ViT), which is long enough to capture the transition from the stable regime into the edge of stability and to let the fluctuations from the onset of self-stabilization dissipate. We compare the iterates and the flows using metrics such as the loss and the sharpness.

\clearpage

\subsection{Cross-Entropy Loss}
\label{app:xent}

\begin{figure}[!htbp]
	\centering
	\textbf{\large Gradient Descent}\par\vspace{0.2cm}
	\includegraphics{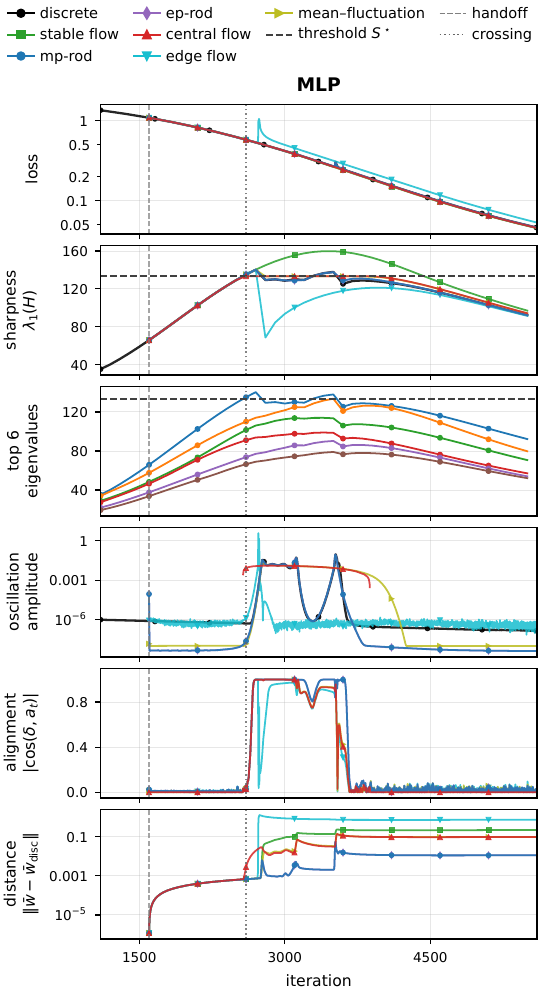}
	\caption{\textbf{Flow Comparison for Gradient Descent Under Cross-Entropy Loss.} \textbf{MLP:} $\eta = 0.015$. The rows show, from top to bottom: the loss at each center; the sharpness at each center; the top six eigenvalues at the discrete optimizer's midpoint; the oscillation amplitude; the alignment of each flow's oscillation with the discrete-time oscillation; and the distance from each flow's center to the discrete optimizer's midpoint. Dashed vertical lines mark the handoff, and dotted vertical lines mark the crossing into the edge of stability.}
	\label{fig:xent-gd}
\end{figure}

\begin{figure}[!htbp]
	\centering
	\textbf{\large RMSProp}\par\vspace{0.2cm}
	\includegraphics{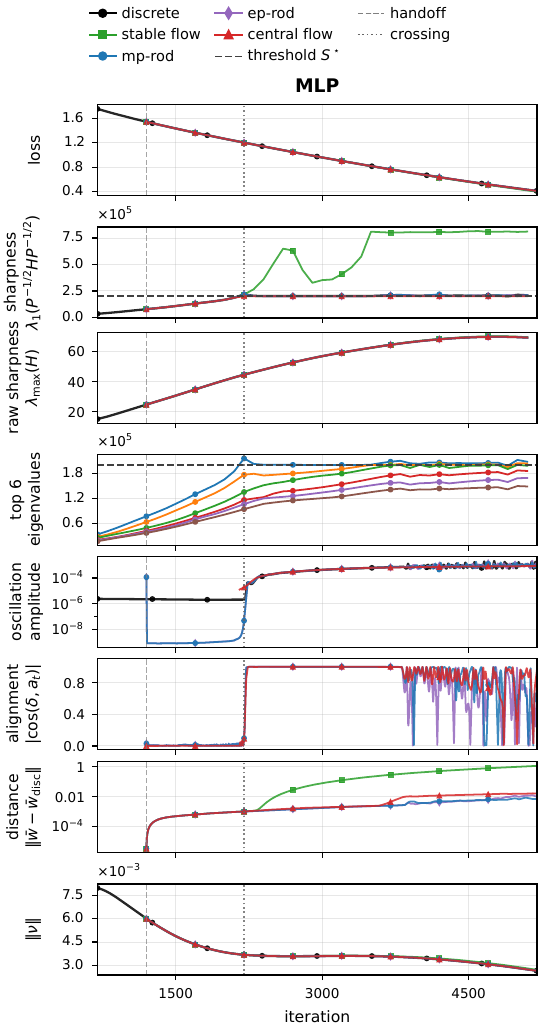}
	\caption{\textbf{Flow Comparison for RMSProp Under Cross-Entropy Loss} (no bias correction). \textbf{MLP:} $\eta = 10^{-5}$, $\beta_2 = 0.99$, $\epsilon = 10^{-8}$. The rows show, from top to bottom: the loss at each center; the preconditioned sharpness at each center; the raw sharpness at each center; the top six preconditioned eigenvalues at the discrete optimizer's midpoint; the oscillation amplitude; the alignment of each flow's oscillation with the discrete-time oscillation; the distance from each flow's center to the discrete optimizer's midpoint; and the norm of the second-moment estimate. Dashed vertical lines mark the handoff, and dotted vertical lines mark the crossing into the edge of stability.}
	\label{fig:xent-rmsprop}
\end{figure}

\Cref{fig:xent-gd,fig:xent-rmsprop} repeat the flow comparisons on the MLP with cross-entropy loss. We ran only gradient descent and RMSProp. For gradient descent, the edge of stability ends about $1\,000$ iterations after the crossing, when the sharpness falls back below the threshold. By the end of the comparison window, rod flow is twenty times closer to the iterates than the stable flow is, and central flow about twice as close. For RMSProp, both are more than fifty times closer.

\clearpage

\subsection{Rod Turning}
\label{app:rod-turning}

\begin{figure}[!htbp]
	\centering
	\textbf{\large Gradient Descent}\par\vspace{0.2cm}
	\includegraphics[width=0.65\textwidth]{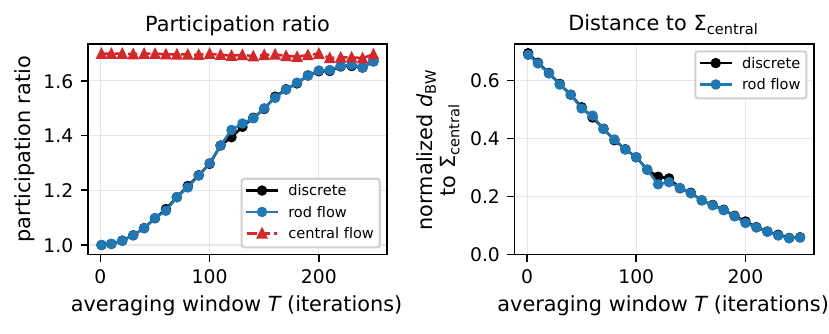}\par\smallskip
	\includegraphics[width=0.93\textwidth]{gradient-descent/rod_turning_gd_mlp_mse_k2.pdf}\par\medskip
	\includegraphics[width=0.65\textwidth]{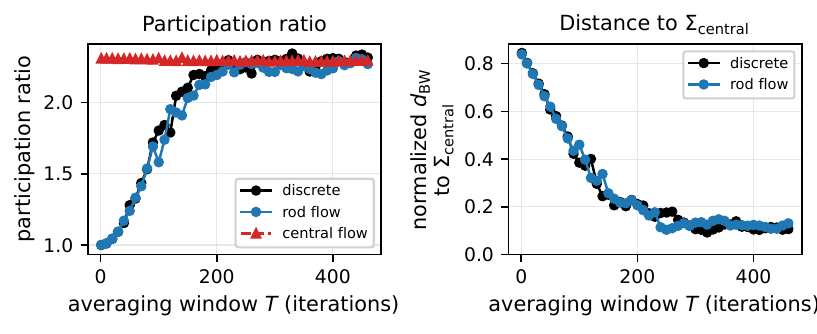}\par\smallskip
	\includegraphics[width=0.93\textwidth]{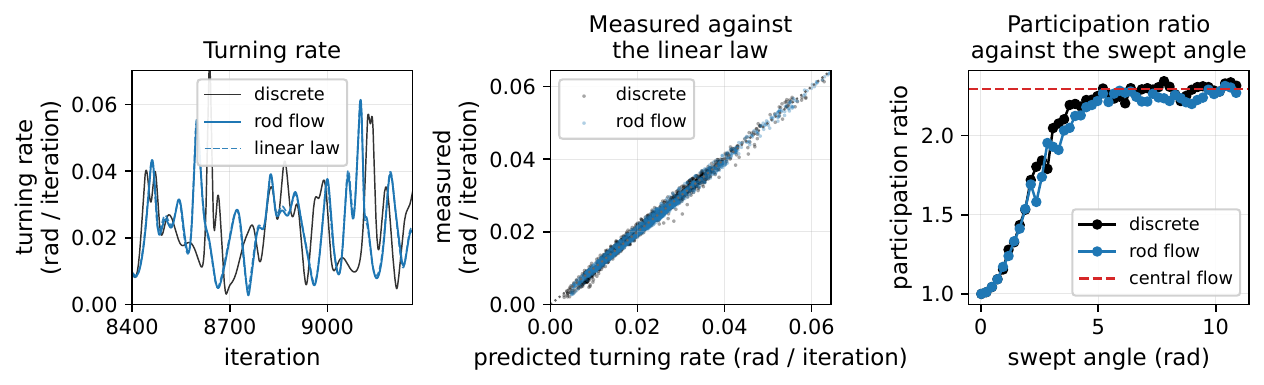}
	\caption{\textbf{Rod Turning for Gradient Descent.} MLP. \textbf{Top:} Two critical eigenvalues, $\eta = 0.05$. \textbf{Bottom:} Three critical eigenvalues, $\eta = 0.075$. In each half, the first row compares the time-averaged extent matrix with $\Sigma_{\mathrm{central}}$. The second row compares the turning rate with the linear law, as in \cref{fig:turning-gd}.}
	\label{fig:rodavg-gd}
\end{figure}

\begin{figure}[!htbp]
	\centering
	\textbf{\large RMSProp}\par\vspace{0.2cm}
	\includegraphics[width=0.65\textwidth]{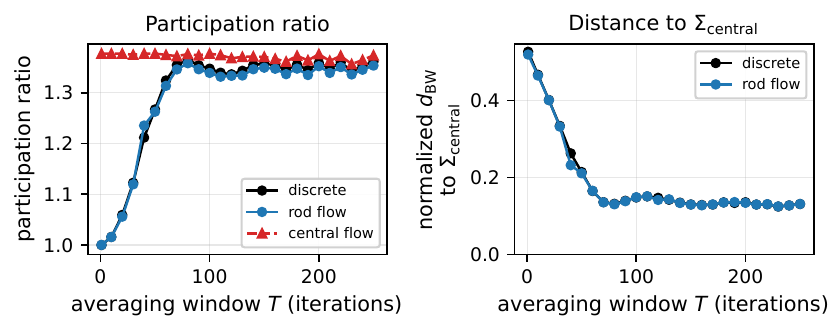}\par\smallskip
	\includegraphics[width=0.93\textwidth]{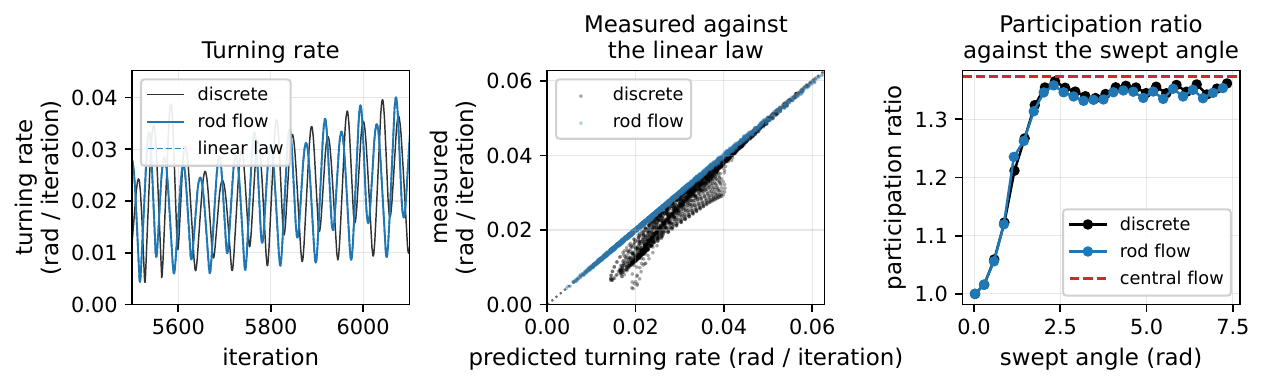}\par\medskip
	\includegraphics[width=0.65\textwidth]{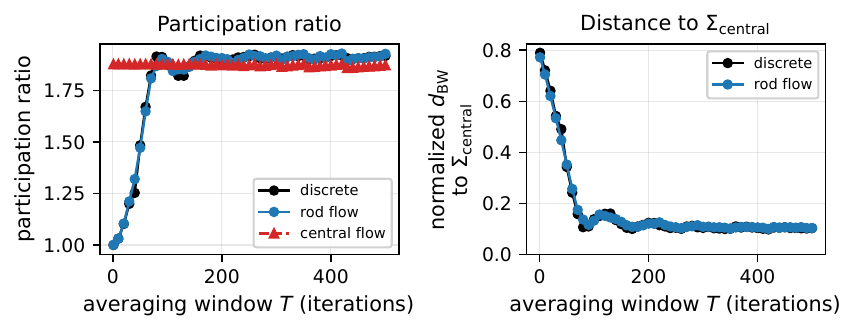}\par\smallskip
	\includegraphics[width=0.93\textwidth]{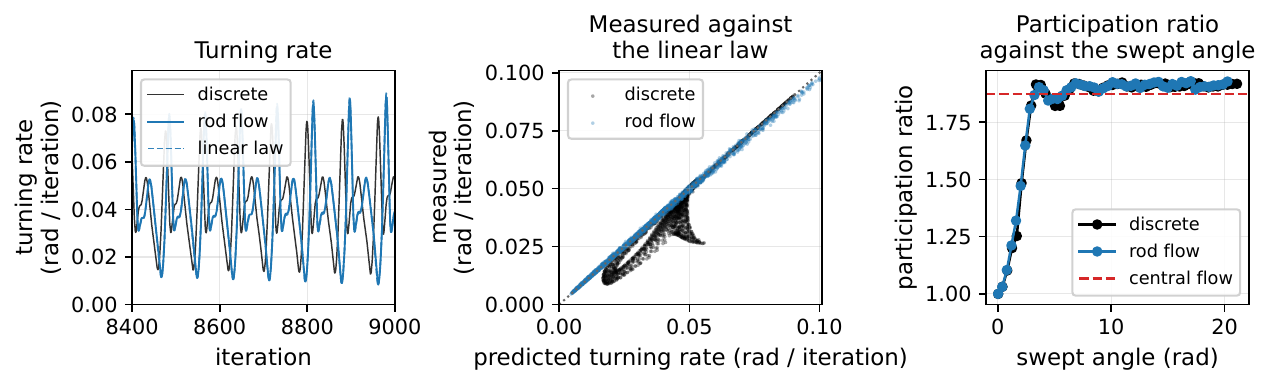}
	\caption{\textbf{Rod Turning for RMSProp} (no bias correction). MLP, $\eta = 2\times10^{-6}$, $\beta_2 = 0.99$. \textbf{Top:} Two critical eigenvalues. \textbf{Bottom:} Three critical eigenvalues. In each half, the first row compares the time-averaged extent matrix with $\Sigma_{\mathrm{central}}$. The second row compares the turning rate with the linear law, as in \cref{fig:turning-gd}.}
	\label{fig:rodavg-rmsprop}
\end{figure}

\begin{figure}[!htbp]
	\centering
	\textbf{\large Heavy Ball}\par\vspace{0.2cm}
	\includegraphics[width=0.65\textwidth]{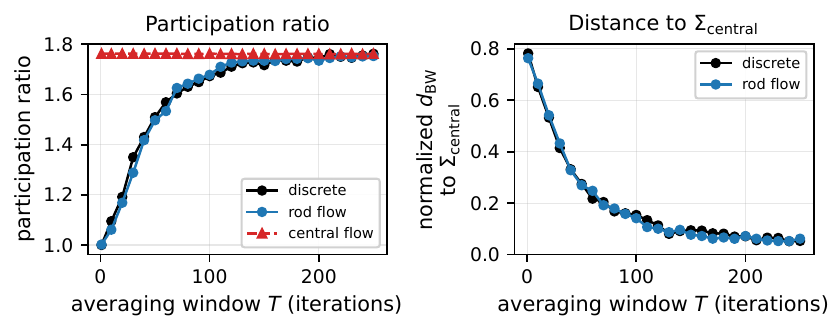}\par\smallskip
	\includegraphics[width=0.93\textwidth]{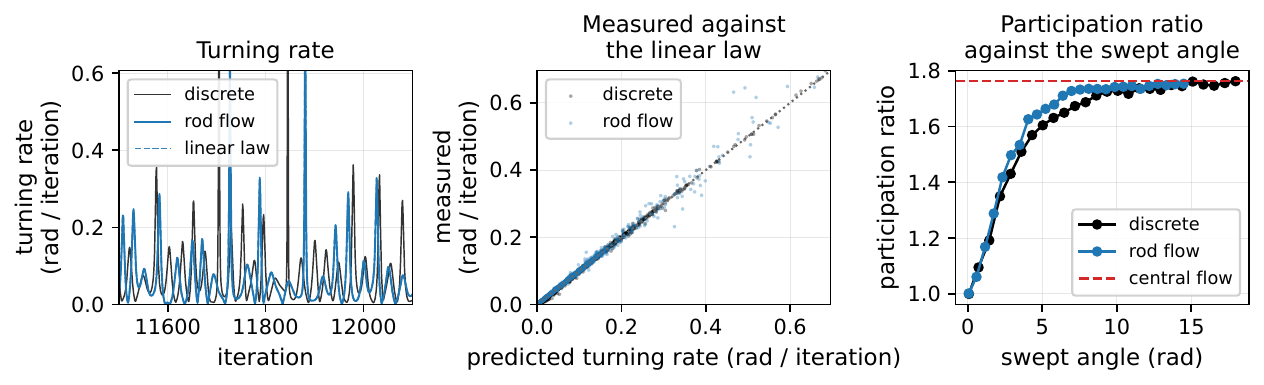}\par\medskip
	\includegraphics[width=0.65\textwidth]{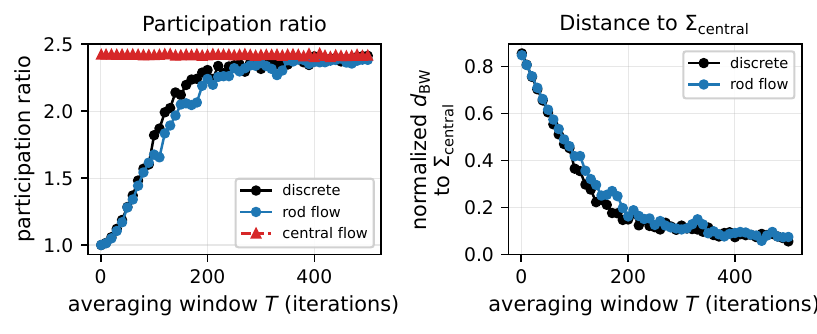}\par\smallskip
	\includegraphics[width=0.93\textwidth]{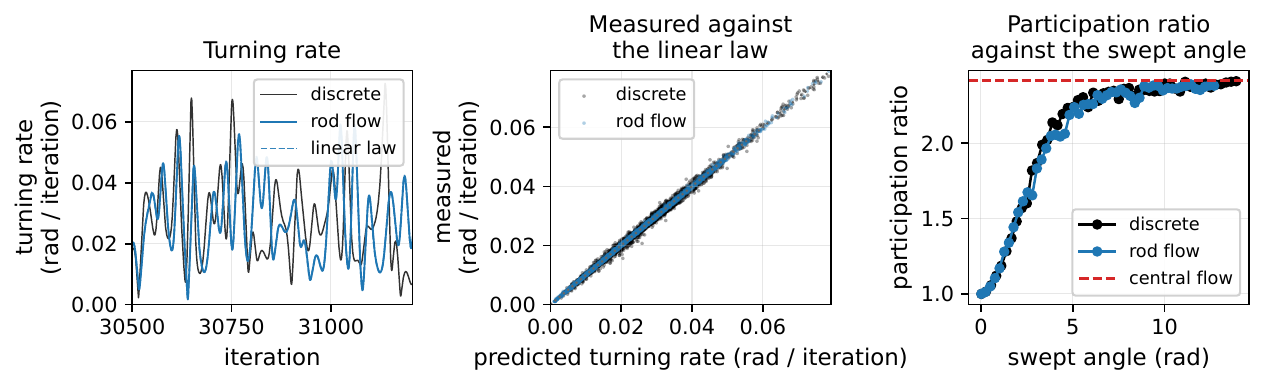}
	\caption{\textbf{Rod Turning for Heavy Ball Momentum.} MLP, $\eta = 0.3$, $\beta = 0.9$. \textbf{Top:} Two critical eigenvalues. \textbf{Bottom:} Three critical eigenvalues. In each half, the first row compares the time-averaged extent matrix with $\Sigma_{\mathrm{central}}$. The second row compares the turning rate with the linear law, as in \cref{fig:turning-gd}.}
	\label{fig:rodavg-hb}
\end{figure}

\begin{figure}[!htbp]
	\centering
	\textbf{\large Nesterov}\par\vspace{0.2cm}
	\includegraphics[width=0.65\textwidth]{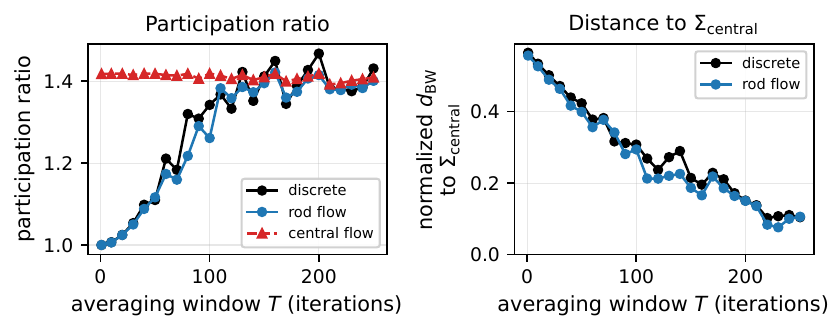}\par\smallskip
	\includegraphics[width=0.93\textwidth]{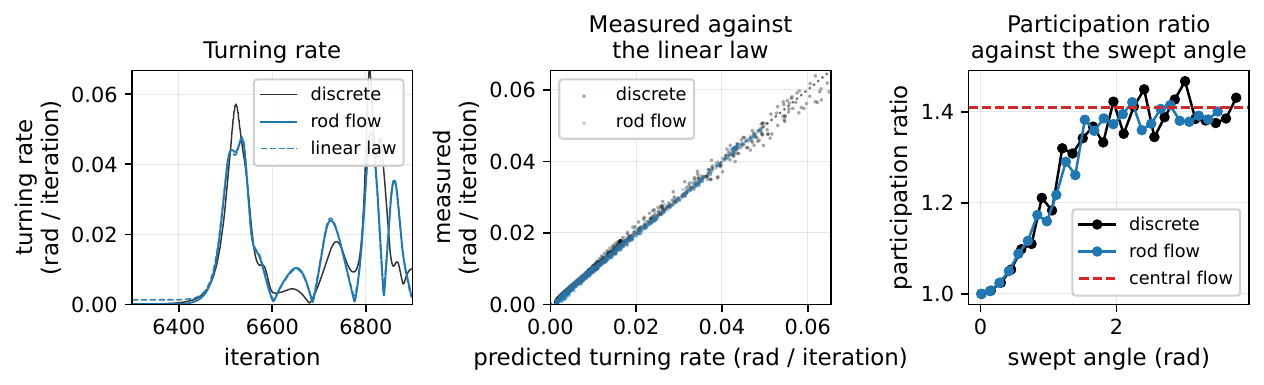}\par\medskip
	\includegraphics[width=0.65\textwidth]{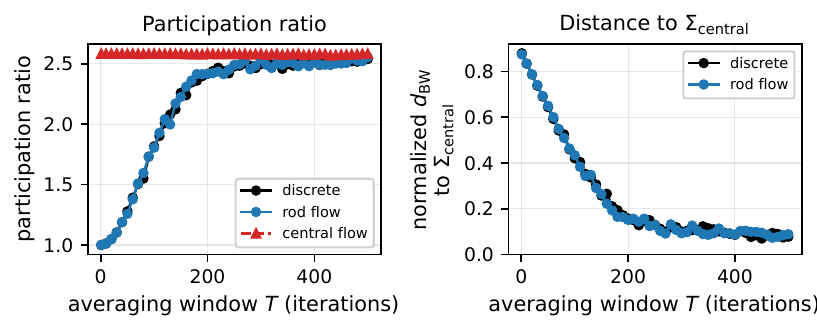}\par\smallskip
	\includegraphics[width=0.93\textwidth]{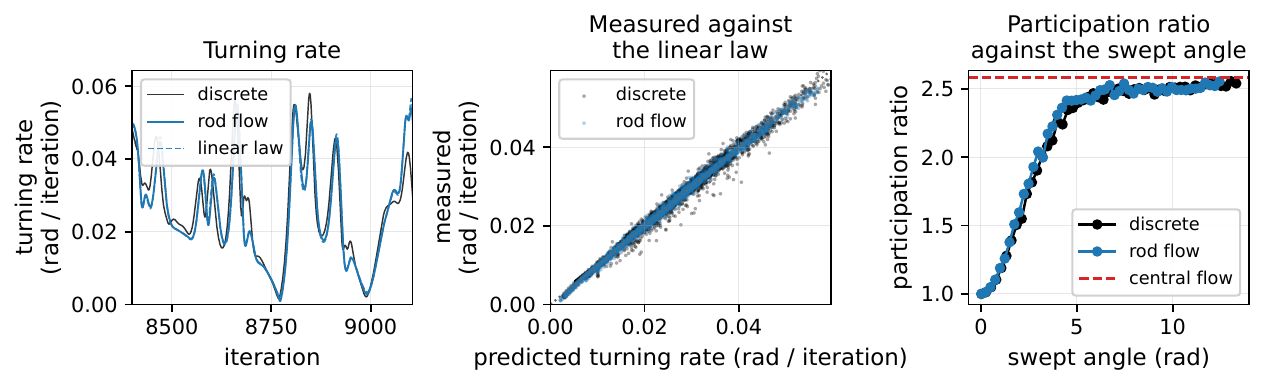}
	\caption{\textbf{Rod Turning for Nesterov Momentum.} MLP, $\eta = 0.2$, $\beta = 0.9$. \textbf{Top:} Two critical eigenvalues. \textbf{Bottom:} Three critical eigenvalues. In each half, the first row compares the time-averaged extent matrix with $\Sigma_{\mathrm{central}}$. The second row compares the turning rate with the linear law, as in \cref{fig:turning-gd}.}
	\label{fig:rodavg-nesterov}
\end{figure}

\begin{figure}[!htbp]
	\centering
	\textbf{\large Adam}\par\vspace{0.2cm}
	\includegraphics[width=0.65\textwidth]{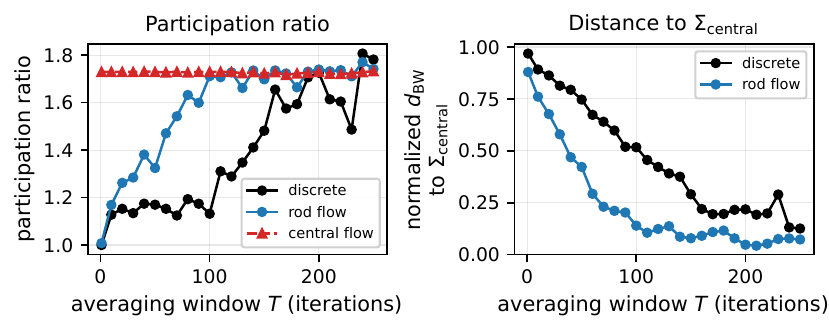}\par\smallskip
	\includegraphics[width=0.93\textwidth]{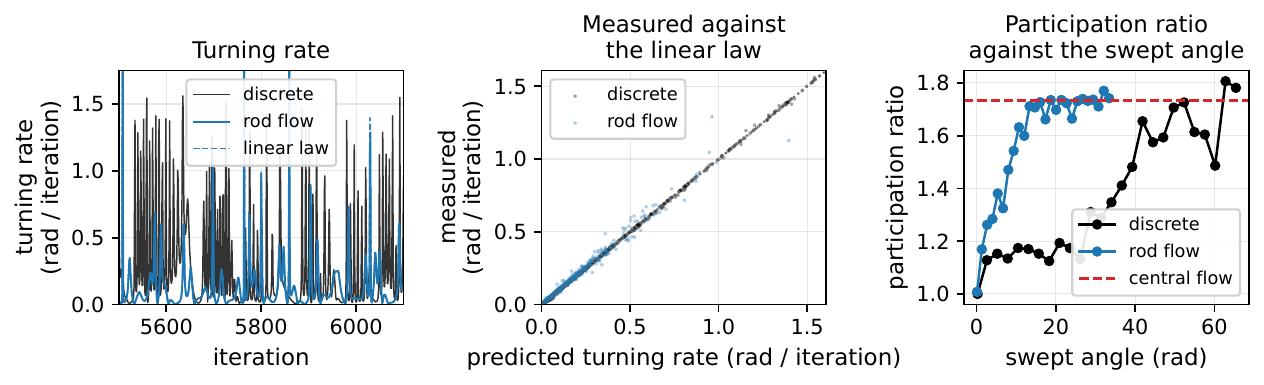}\par\medskip
	\includegraphics[width=0.65\textwidth]{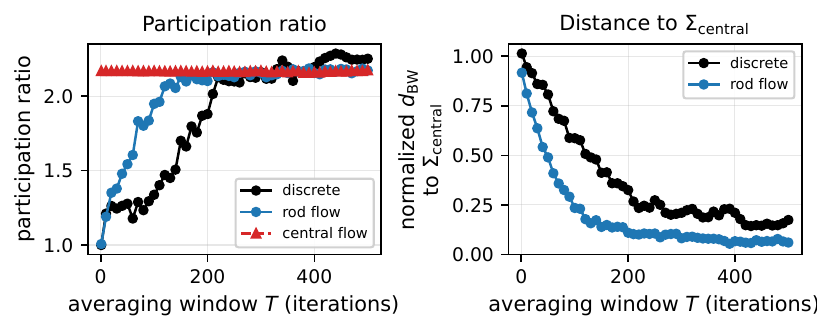}\par\smallskip
	\includegraphics[width=0.93\textwidth]{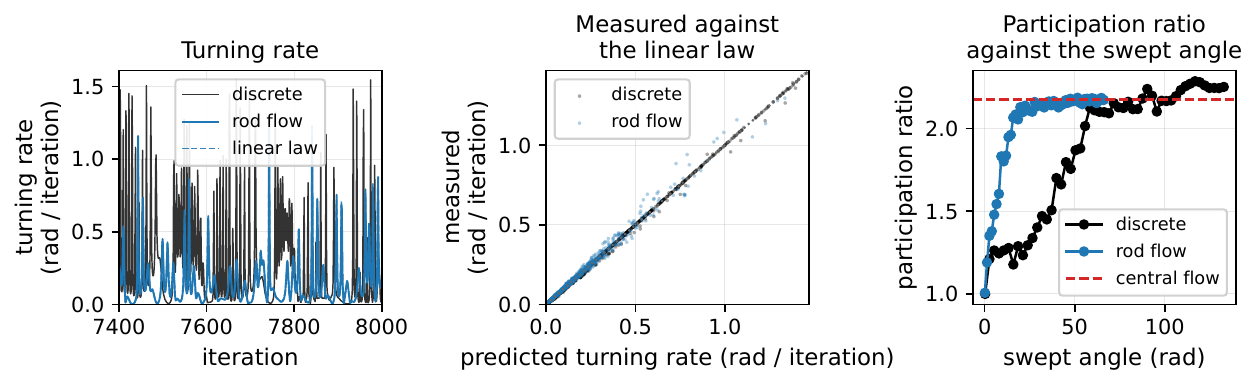}
	\caption{\textbf{Rod Turning for Adam} (with bias correction). MLP, $\eta = 10^{-5}$, $\beta_1 = 0.9$, $\beta_2 = 0.999$. \textbf{Top:} Two critical eigenvalues. \textbf{Bottom:} Three critical eigenvalues. In each half, the first row compares the time-averaged extent matrix with $\Sigma_{\mathrm{central}}$. The second row compares the turning rate with the linear law, as in \cref{fig:turning-gd}.}
	\label{fig:rodavg-adam}
\end{figure}

\begin{figure}[!htbp]
	\centering
	\textbf{\large AdamW}\par\vspace{0.2cm}
	\includegraphics[width=0.65\textwidth]{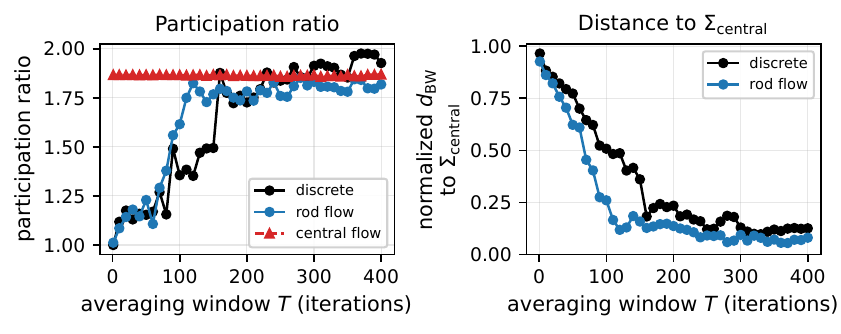}\par\smallskip
	\includegraphics[width=0.93\textwidth]{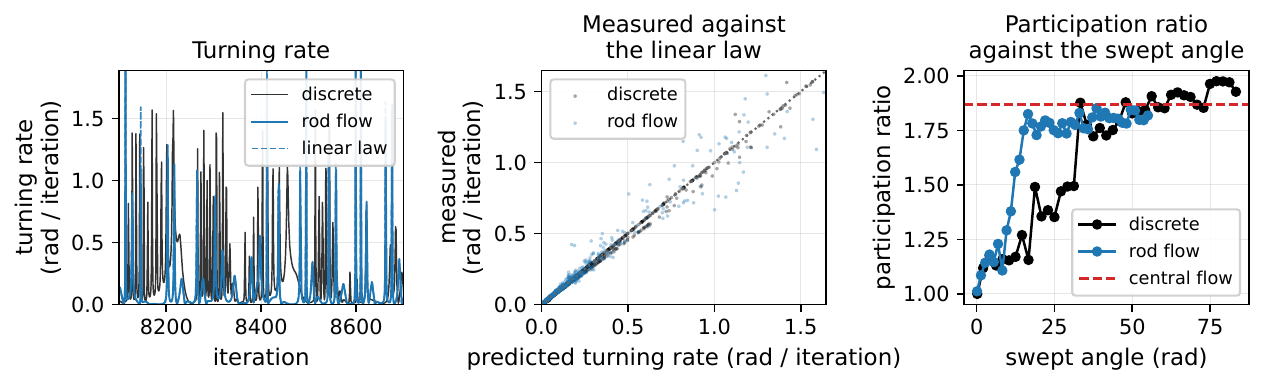}\par\medskip
	\includegraphics[width=0.65\textwidth]{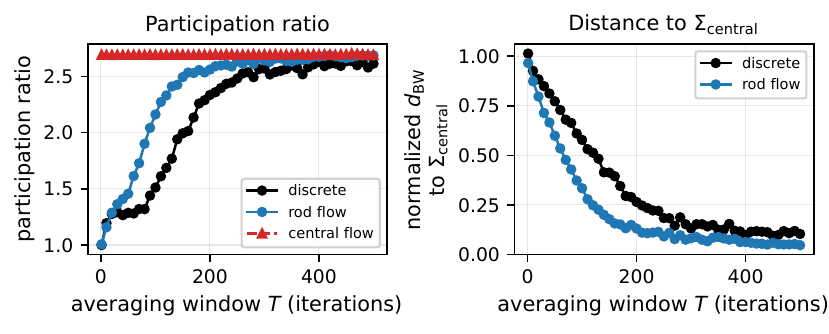}\par\smallskip
	\includegraphics[width=0.93\textwidth]{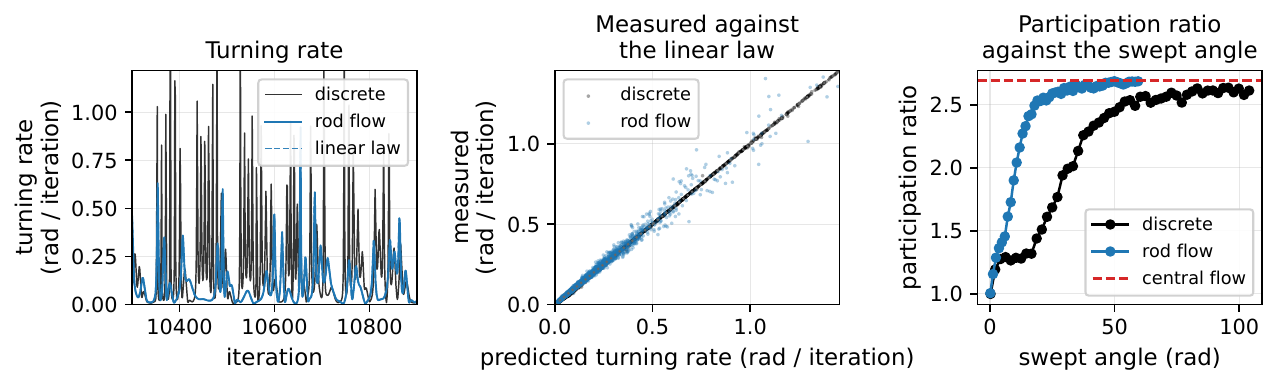}
	\caption{\textbf{Rod Turning for AdamW} (with bias correction). MLP, $\eta = 10^{-5}$, $\zeta = 10$, $\beta_1 = 0.9$, $\beta_2 = 0.999$. \textbf{Top:} Two critical eigenvalues. \textbf{Bottom:} Three critical eigenvalues. In each half, the first row compares the time-averaged extent matrix with $\Sigma_{\mathrm{central}}$. The second row compares the turning rate with the linear law, as in \cref{fig:turning-gd}.}
	\label{fig:rodavg-adamw}
\end{figure}

\Cref{fig:rodavg-gd,fig:rodavg-rmsprop,fig:rodavg-hb,fig:rodavg-nesterov,fig:rodavg-adam,fig:rodavg-adamw} repeat the rod-turning experiments of \cref{fig:time-average,fig:turning-gd} for each optimizer on the MLP. Each experiment again starts with a warmup phase, but the warmup now continues past the onset of the edge of stability, because the goal is to capture the window in which there are multiple critical eigenvalues.

At the end of the warmup, rod flow and central flow are initialized from the discrete iterates, and the iterates and the flows then advance in lockstep for a couple thousand iterations. The first $300$--$500$ of these iterations let the flows settle. The rest form the averaging span, throughout which there are exactly two critical eigenvalues (three in the bottom half of each figure). We cut the span into consecutive windows of $T$ iterations, and over each window we average the extent matrix $\Sigma_{\mathrm{rod}}$ of rod flow and its discrete counterpart $a_t\otimes a_t$. We compare each average with $\Sigma_{\mathrm{central}}$ at the middle of its window using the participation ratio and the normalized Bures--Wasserstein distance. We also compare the turning rates of rod flow and of the discrete oscillation with those predicted by the linear law (derived in \cref{sec:turning,sec:timescales-momentum}).

In every run, the participation ratio of the time-averaged extent matrix completes $90\%$ of its rise within $250$ iterations, and at the widest window it is within $4\%$ of that of central flow, for both the discrete iterates and rod flow. As the window widens, the normalized distance to $\Sigma_{\mathrm{central}}$ falls from $0.5$--$1.0$ for a single step to $0.05$--$0.17$.

In every run, the measured turning rate tracks the linear law, with a correlation of at least $0.95$. On average, the two agree to within $3\%$ for both rod flow and the discrete oscillation, except that RMSProp's discrete oscillation turns about $10\%$ slower than predicted.

\clearpage
\subsection{Position and Momentum Extents}
\label{app:two-cycle}

\begin{figure}[!htbp]
	\centering
	\includegraphics[width=\textwidth]{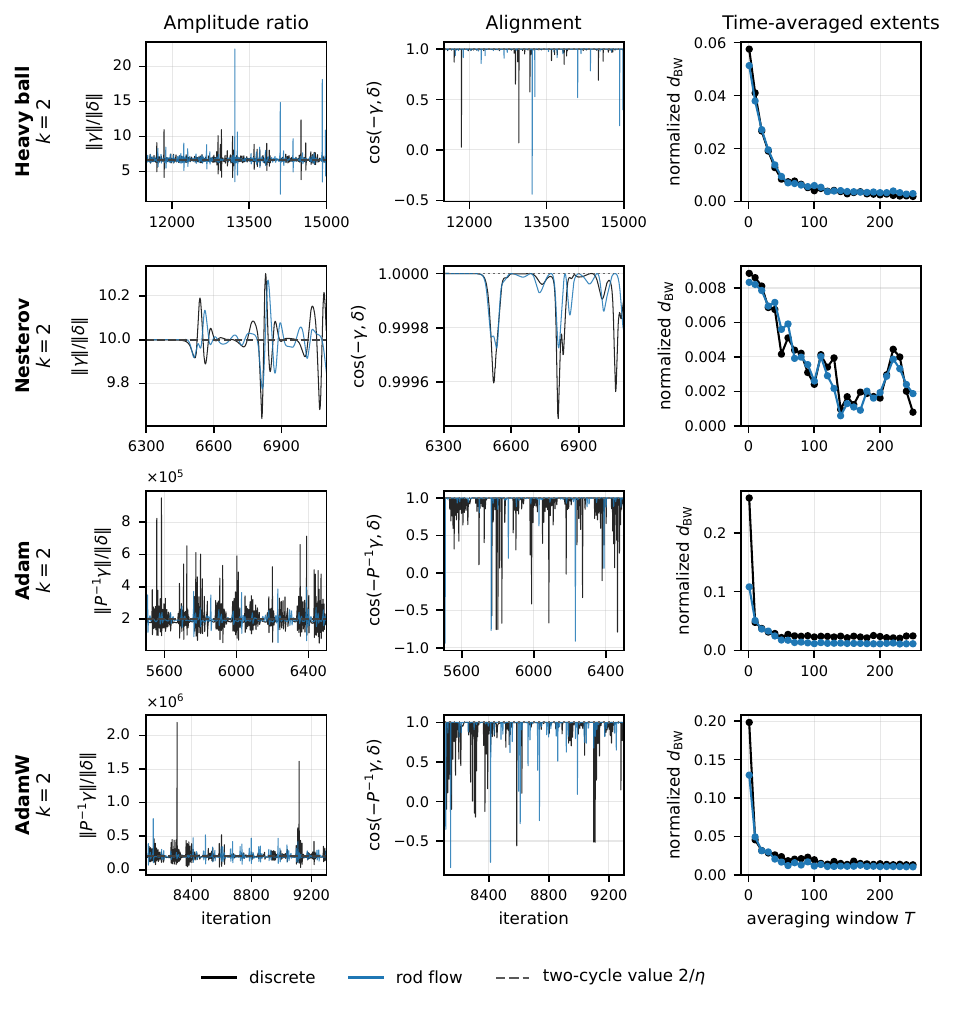}
	\caption{\textbf{Position and Momentum Extents of the Momentum Optimizers.} MLP, with two critical eigenvalues. \textbf{Heavy ball:} $\eta = 0.3$, $\beta = 0.9$. \textbf{Nesterov:} $\eta = 0.2$, $\beta = 0.9$. \textbf{Adam} (with bias correction): $\eta = 10^{-5}$, $\beta_1 = 0.9$, $\beta_2 = 0.999$. \textbf{AdamW:} as Adam, with $\zeta = 10$. The columns show, from left to right: the ratio of the momentum amplitude to the position amplitude, with its value on a period-two orbit (dashed); the alignment between the position and momentum extents; and the distance between their time averages, against the width of the averaging window. For Adam and AdamW, the momentum extent is preconditioned.}
	\label{fig:two-cycle-momentum}
\end{figure}

\begin{figure}[!htbp]
	\centering
	\includegraphics[width=\textwidth]{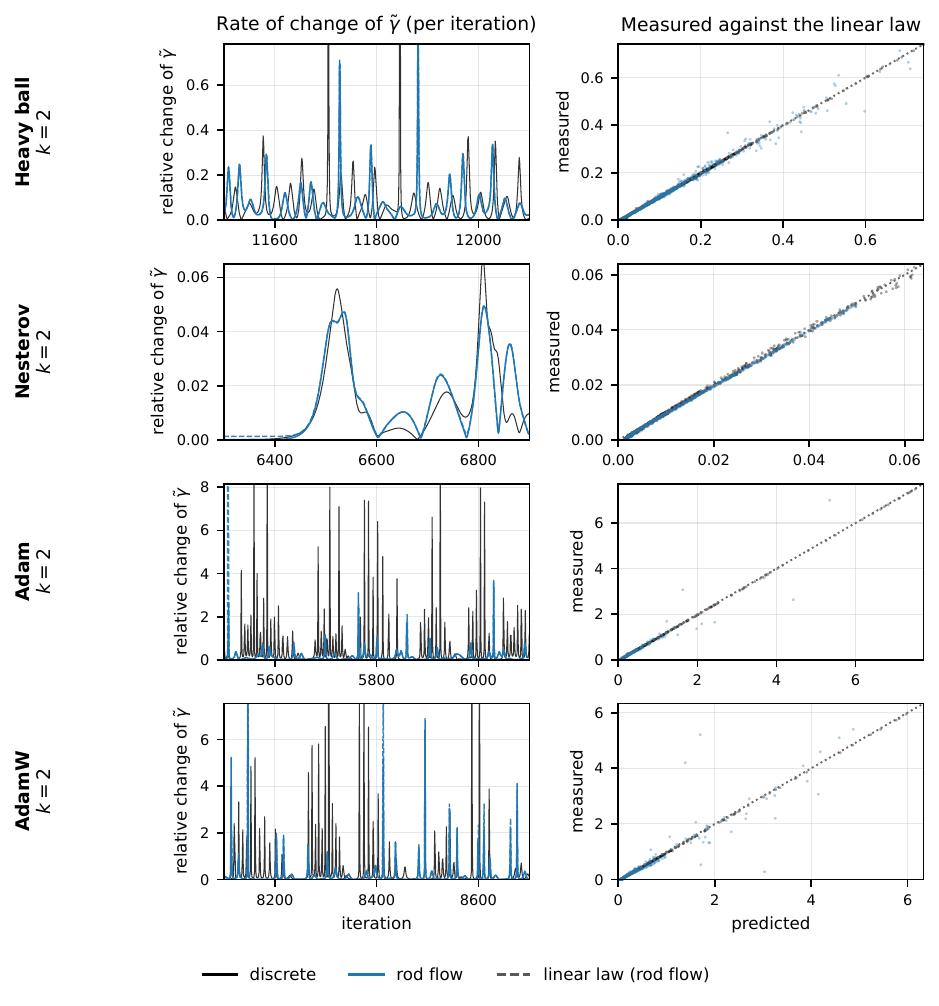}
	\caption{\textbf{Rate of Change of the Relative Momentum.} Runs as in \cref{fig:two-cycle-momentum}. The relative momentum $\tilde{\gamma}$ is the momentum extent divided by the position amplitude. \textbf{Left:} The relative change of $\tilde{\gamma}$ per iteration, for the discrete oscillation and for rod flow, with the linear law for rod flow (dashed). \textbf{Right:} Measured against the linear law.}
	\label{fig:gamma-turning-momentum}
\end{figure}

\Cref{fig:two-cycle-momentum,fig:gamma-turning-momentum} repeat the experiments of \cref{sec:timescales-momentum} for each momentum optimizer. Each run has the same warmup and averaging span as the corresponding two-critical-eigenvalue run in \cref{app:rod-turning}. We ran only rod flow alongside the iterates, since central flow represents only the position oscillations.

Over this span, we compare the momentum extent with the position extent for both the iterates and rod flow. We use three metrics: the amplitude ratio between the two extents, the alignment between them, and the normalized Bures--Wasserstein distance between the time averages of their outer products, rescaled as in \cref{fig:two-cycle-hb}. We also compare the rate of change of the relative momentum $\tilde{\gamma}$ with that predicted by the linear law.

In every run, the median amplitude ratio is within $5\%$ of $2/\eta$, its value on a period-two orbit. The median alignment is above $0.998$ for both the iterates and rod flow, though both quantities fluctuate in bursts for Adam and AdamW. As the averaging window widens from one iteration to $250$, the distance between the time-averaged extent matrices falls ten- to thirtyfold, except for Nesterov momentum, where it is below $0.01$ even without time averaging. The rate of change of $\tilde{\gamma}$ also tracks the linear law, for both the iterates and rod flow.

\clearpage
\subsection{Near Two-Periodicity}
\label{app:two-periodic}

\citet{regis2026adamrod} suggested that period-doubling bifurcations might explain why their momentum rod flow failed to track the iterates at large $\beta_1$. Their rationale was that since rod flow is intended to model nearly two-periodic behavior, there is reason to suspect that it might fail if the iterates oscillated in four-cycles.

However, at the hyperparameters used in our experiments, theory suggests nearly two-periodic behavior. \citet{fong2026adam} analyze Adam at the edge of stability under the condition $\beta_2 > \beta_1^2$. They show that when $\beta_2 \le \beta_1^2$, Adam can have orbits of period four. Their experiments agree: across fully connected networks, ResNets, ViTs, LSTMs, and GPT-2, and across values of $\beta_1$ and $\beta_2$, they found that much of the motion at the edge of stability is a fast, approximately period-two oscillation.

Since our setting satisfies $\beta_2 = 0.999 > \beta_1^2 = 0.81$ by a wide margin, we also expect nearly two-periodic behavior. To test this, we measure how close the discrete oscillation is to a two-cycle on the MLP when there are two critical eigenvalues. At every iteration, we record the oscillation envelope $a_t$ in the basis of the top five eigenvectors of the (preconditioned) Hessian. Since $a_t$ undoes the sign alternation of a two-cycle, the oscillation itself is $o_t = (-1)^t a_t$.

Let $S(f)$ be the power spectrum of $o_t$. To estimate $S$, we use Welch's method on Hann windows of $256$ iterations and sum over the coordinates. A two-cycle puts all of its power at $f = \tfrac12$ cycles per iteration. Period doubling would add power at $f = \tfrac14$ for a cycle of period four, at $\tfrac38$ for period eight, and at $\tfrac{7}{16}$ for period sixteen.

\begin{figure}[!htbp]
	\centering
	\includegraphics[width=\textwidth]{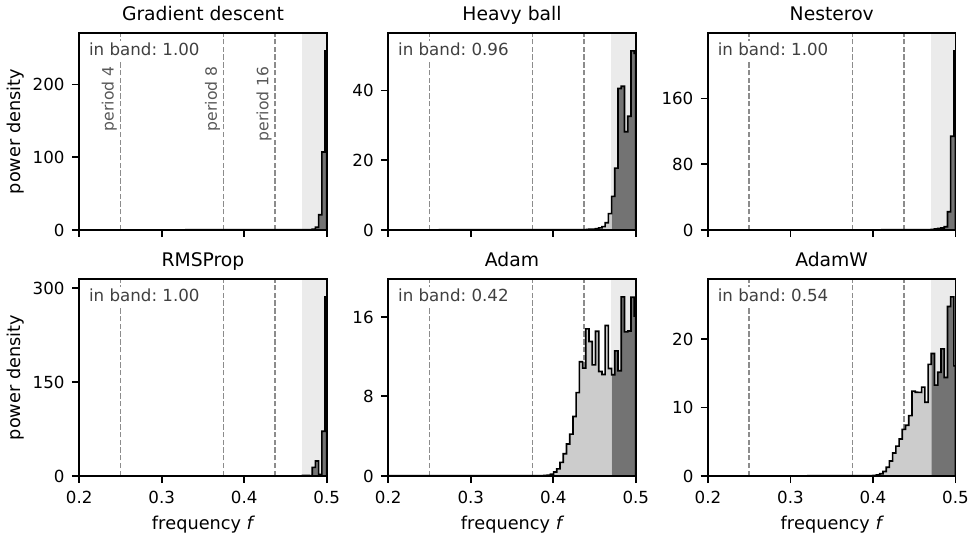}
	\caption{\textbf{Power Spectrum of the Discrete Oscillation.} MLP with two critical eigenvalues. Runs as in the top halves of \cref{fig:rodavg-gd,fig:rodavg-rmsprop,fig:rodavg-hb,fig:rodavg-nesterov,fig:rodavg-adam,fig:rodavg-adamw}. Each panel shows the power spectrum of the oscillation of the iterates, normalized to unit area, so that the area under the curve over any range of frequencies is the share of the power in that range. The gray band marks the frequencies within $0.03$ of $\tfrac12$, and the number in each panel gives the share of the power in this band (dark area). Dashed lines mark where period doubling to periods four, eight, and sixteen would add power.}
	\label{fig:two-cycle-spectra}
\end{figure}

Our runs agree with \citet{fong2026adam} (\cref{fig:two-cycle-spectra}). For every optimizer, the power is concentrated near $f = \tfrac12$. What departs from a steady two-cycle is the regularity of the oscillation. Less than half of Adam's power lies within $0.03$ of $f = \tfrac12$, compared with nearly all of it for gradient descent, RMSProp, heavy ball, and Nesterov momentum. The rest forms a broad hump extending down to $f \approx 0.4$. This hump peaks near $\tfrac{7}{16}$, but it is many times wider than a spectral line, so the oscillation also varies on a time scale of ten to thirty iterations, though not periodically. AdamW behaves similarly, but slightly more regularly.
\clearpage

\subsection{The Ring and the Onset}
\label{app:ring-onset}

\begin{figure}[!htbp]
	\centering
	\includegraphics[width=\textwidth]{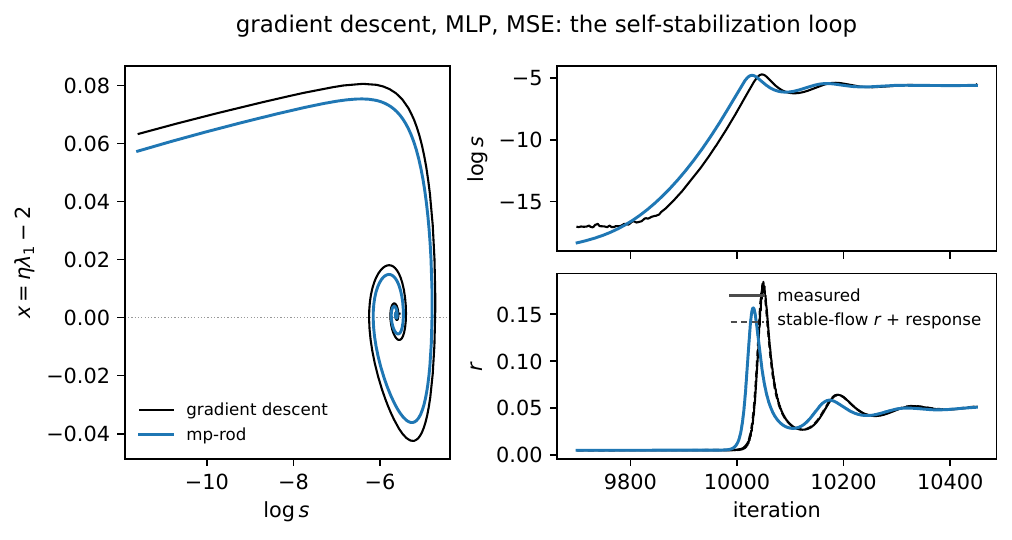}\par\medskip
	\includegraphics[width=\textwidth]{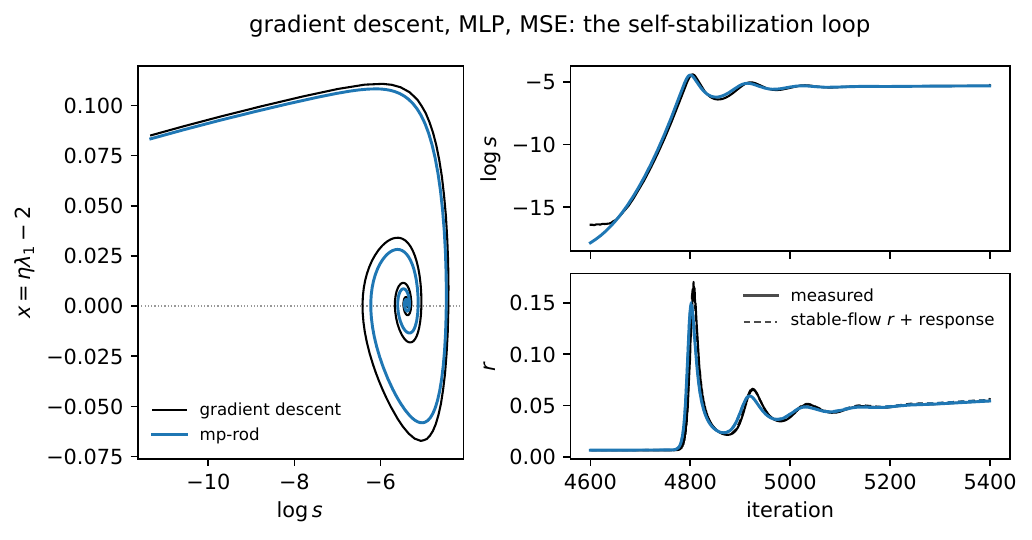}
	\caption{\textbf{The Self-Stabilization Loop.} Gradient descent and rod flow on an MLP. \textbf{Top:} $\eta = 0.05$. \textbf{Bottom:} $\eta = 0.075$. In each half, the left panel shows the phase portrait of the loop in the log-amplitude and the excess sharpness. The right panels show the log-amplitude and the progressive sharpening constant over time. The dashed curve is the progressive sharpening constant of the stable flow plus the response of the center to the oscillation, with no free parameters.}
	\label{fig:ring-loop}
\end{figure}

\begin{figure}[!htbp]
	\centering
	\includegraphics[width=\textwidth]{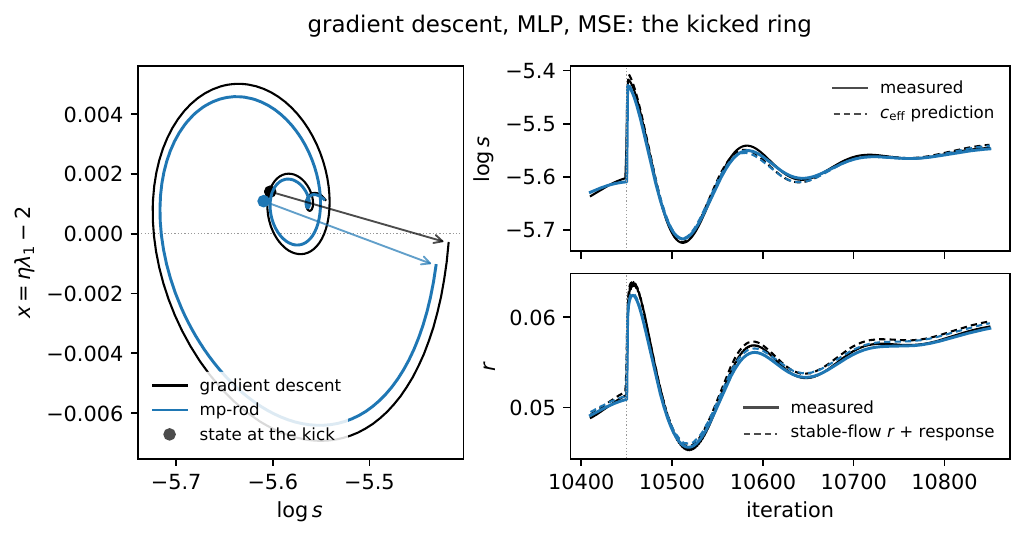}\par\medskip
	\includegraphics[width=\textwidth]{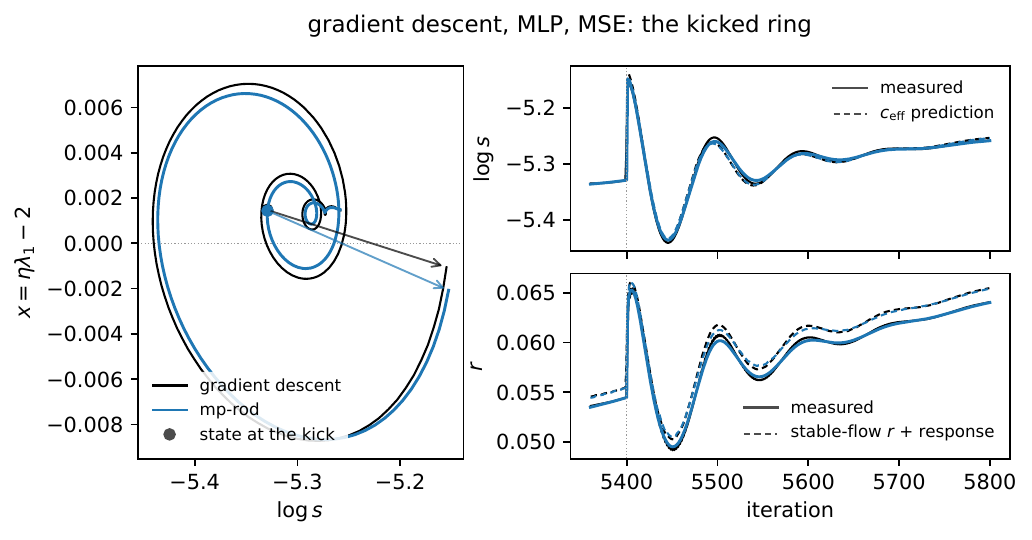}
	\caption{\textbf{The Kicked Ring.} Gradient descent and rod flow on an MLP, kicked by multiplying the amplitude of the oscillation by $1.2$. \textbf{Top:} $\eta = 0.05$. \textbf{Bottom:} $\eta = 0.075$. In each half, the left panel shows the phase portrait, with dots marking the state at the kick and arrows marking the kick. The right panels show the log-amplitude and the progressive sharpening constant over time. The dashed curves are predictions. For the log-amplitude, only the frequency and the decay rate are predicted; the amplitude, phase, and trend are fitted.}
	\label{fig:ring-kick}
\end{figure}

\Cref{fig:ring-loop,fig:ring-kick} test the self-stabilization loop \eqref{eq:lv} for gradient descent and rod flow on an MLP with a single critical eigenvalue. Each experiment starts with a warmup phase in which the discrete iterates progressively sharpen. Rod flow and the stable flow are initialized from the discrete iterates $1\,000$ iterations before the edge of stability. The iterates and the flows then advance in lockstep for about $2\,200$ iterations, long enough to capture the onset of self-stabilization and to let the ring that follows it decay. We kick the oscillation $750$--$800$ iterations after the crossing, and follow the ring that the kick sets off for the last $450$ iterations. At each center, we measure $s$, $x$, $r$, and $c$ at every iteration.

The orbit-average relation holds: over each orbit, $\langle s^2\rangle$ is within $0.3\%$ of $2r/(\eta c)$ for gradient descent and within $3\%$ for rod flow. However, the orbits themselves are not closed. They spiral inward, decaying at a rate of $0.011$--$0.014$ per iteration, and their period is $1.4$--$1.6$ times the period $2\pi/\sqrt{2\eta r}$ of the loop. Rod flow traces the same spiral as the iterates, reaching the onset slightly ahead of them and overshooting slightly less.

Why are the $x$--$s$ loops not closed? Unlike in an idealized Lotka--Volterra system, $r$ is not constant at the edge of stability. The oscillation pushes the center with velocity $-\tfrac{\eta}{2}s^2\nabla\lamone$, and the center's own descent relaxes this push along each eigenvector $\hat{u}_i$ at the rate $\eta\lambda_i$, which raises $r$. In the steady-state edge-of-stability regime, the $r$ of the iterates is about ten times that of the stable flow.

With this response, $c$ in the linearized loop is replaced by a frequency-dependent coefficient $c_{\mathrm{eff}}(z)$, and the loop's complex frequencies $z$ solve
\begin{equation*}
	z^2 + \eta^2\langle s^2\rangle\,c_{\mathrm{eff}}(z) = 0,
	\qquad
	c_{\mathrm{eff}}(z) = \sum_i b_i^2\,\frac{z}{z + \eta\lambda_i},
\end{equation*}
where $b_i = \inner{\nabla\lamone}{\hat{u}_i}$. Directions much flatter than the ring leave their contribution to $c$ intact, as the loop assumes. Much stiffer directions relax the push before it acts, softening the restoring force and lengthening the period. Directions in between respond with a lag, which damps the ring. For the kicked ring, the roots of this equation match the measured period to within $5\%$ and the decay rate to within $10\%$ (\cref{fig:ring-kick}).

